\PassOptionsToPackage{table}{xcolor} %
\documentclass{article}
\usepackage{iclr2026_conference,times}

\usepackage{amsmath,amsfonts,bm}

\def\eqref#1{equation~\ref{#1}}

\def\1{\bm{1}}

\DeclareMathAlphabet{\mathsfit}{\encodingdefault}{\sfdefault}{m}{sl}
\SetMathAlphabet{\mathsfit}{bold}{\encodingdefault}{\sfdefault}{bx}{n}

\usepackage{hyperref}
\usepackage{etoc}
\makeatletter
\AtBeginDocument{\let\orig@footnotetext\@footnotetext
  \long\def\@footnotetext#1{\orig@footnotetext{\fontsize{8}{9}\selectfont #1}}}
\makeatother
\hypersetup{hidelinks}
\usepackage{url}
\usepackage{graphicx}
\usepackage{booktabs}
\usepackage{array}
\usepackage{algorithm}
\usepackage{algpseudocode}
\usepackage{wrapfig}
\usepackage{subcaption}
\usepackage{tikz}
\usepackage[normalem]{ulem} %

\providecommand{\Tiny}{\fontsize{4.5}{5.4}\selectfont}
\DeclareRobustCommand{\unitchip}[1]{\tikz[baseline=(u.base)]{\node[draw=green!45!black, fill=green!12, rounded corners=1.5pt, inner sep=1.6pt, font=\scriptsize] (u) {#1};}}
\usetikzlibrary{positioning,arrows.meta,fit,calc,decorations.pathreplacing,decorations.pathmorphing,patterns}
\tikzset{
	blk/.style={draw, rounded corners=2pt, align=center, font=\small, inner sep=4pt, minimum height=2.1em},
	unit/.style={draw, fill=black!8, rounded corners=1pt, font=\scriptsize, inner sep=2pt},
	ghost/.style={draw, dashed, fill=none, rounded corners=1pt, font=\scriptsize, inner sep=2pt, text=black!55},
	lab/.style={font=\scriptsize\itshape, align=center},
	arr/.style={-{Stealth[length=2mm]}, thick},
	darr/.style={-{Stealth[length=2mm]}, thick, dashed, black!55},
}

\title{Articulation Upper Bounds for Language Models}

\author{Alexander Spangher \\
	Stanford University \\
	\texttt{alexspan@stanford.edu}
	\And
	Daniel E. Ho \\
	Stanford University
	\And
	Diyi Yang \\
	Stanford University
	\And
	Noah Goodman \\
	Stanford University
	\And
	Sanmi Koyejo \\
	Stanford University}

\iclrfinalcopy %

\begin{document}
\etocdepthtag.toc{mtbody}
	\maketitle
	\lhead{Preprint}  %

	\begin{abstract}
        Language captures only a subset of the perceptions, intuitions and intents that form meaning between speakers. What, then, are the limits of language? We take steps towards measuring this in LLMs as follows: for a set of concepts, we measure how well each concept can be transmitted between two LLMs as their capabilities scale. This requires novel measurement tools: (1) \textit{optimal messages}: we introduce a prompt optimization algorithm that beats existing algorithms by up to $14\%$ on benchmarks and accurately estimates \textit{prompt upper bounds}. (2) A \textit{transmission measure}: the sender is asked to label text; the receiver then tries to recover the sender's prompt, and recovery is measured; we find this correlates strongly with human label error. With these measures, we use \textit{observational scaling laws} to study how transmissibility of a concept, conveyed with optimal messages, increases as transmitter and receiver capacity grows: does it asymptote below or above a threshold? Measuring $1.2$k concepts in eight domains, we find concepts that (a) can be fully articulated; these concepts usually depend on \textit{local patterns}, e.g. ``\textit{joke ends with surprise}''; (b) might be articulated by future, stronger models; these rely on \textit{whole-text structural patterns}, e.g. ``story's worldbuilding coherence''; (c) concepts that face articulability limits; these concepts' meaning, we find, is tied to \textit{identity and social-group awareness}, e.g. \textit{``comic voice''}. Interesting, these same concepts are not all challenging for humans: humans have moderate-to-high levels of agreement in $76\%$ of concepts that LLMs struggle with; articulation limits, we find, are \textit{executor relative}. Finally, we characterize the nature of improvement in each case: while recoverable concepts can be articulated through definitions and explanations, irrecoverable concepts cannot be transmitted, even with examples, and might require novel learning mechanisms.
\end{abstract}

\section{Introduction}
\label{sct:introduction}

\textit{``Describe the aroma of coffee --- why can't it be done? Do we lack the words?''}: in \textit{Philosophical Investigations}, \citet{wittgenstein1953pi} invites the sleep-deprived reader to imagine the limits of \textit{articulation} in conveying a concept. %
Large language models perceive, act, and judge entirely through language; how well a concept can be articulated and understood, e.g. in rubrics \citep{gunjal2025rubrics} or constitutions \citep{bai2022constitutional}, determines how a well-meaning agent behaves. 
Some concepts (e.g. the ``aroma of coffee'') might be fully inarticulable, for others we might simply not have found the \textit{right} words or the \textit{right} speakers \& listeners.
We measure each component as follows. 

First, we start with optimizing the message: for a given transmitter (i.e. language model) we find the \textit{optimal message} for a given concept. We introduce the $\epsilon$-estimation algorithm, shown in Figure \ref{fig:mining-bound}, a prompt optimization approach and an upper bound estimator on prompt performance%
\footnote{We distinguish \textit{performance} from \textit{articulability}. Performance is a model's score against a labeled dataset; it depends on the prompt, the model and its background knowledge together. Articulability is how much of a concept can be moved from one party to another in words; it depends on what the two already share and on how identifiable the concept is in language.
}.
$\epsilon$-estimation works by: constructing a search space over \textit{units} of meaning\footnote{
A \textit{unit} (see Section \ref{sec:unit-mining}), is a string in a prompt that minimally changes LLM behavior.
}, exploring this space by optimally combining units and, finally, estimating missing search steps to calculate asymptotic prompt performance. We find $\epsilon$-estimation matches or beats existing prompt optimization algorithms \citep{agrawal2025gepa,opsahl-ong2024optimizing} by up to 14\%, at equal budgets, across six benchmarks
. Further, we find its estimates accurately predict improvement across iterations with ${\approx}5\%$ error mid-run.

Next, we introduce a novel instrument to measure concept transmission between language models. We introduce a reconstruction-bottleneck, shown in Figure \ref{fig:reconstruction}: one LLM is prompted to identify a concept in text (e.g. ``does this joke end with a surprise?''); a second LLM is shown these measurements with the text and asked to generate a prompt that \textit{would have} generated these labels. We measure how well the second LLM recovered the first LLM's prompt. This bottleneck is a robust \textit{unsupervised} transmission metric and, we find, reliably predicts \textit{supervised} performance in cases where ground truth labels exist.%

With these tools in hand, we measure scaling laws \citep{ruan2024observational} to study how transmission improves as model size increases: how well can an \textit{infinite capacity transmitter} convey \textit{a perfect message} to an \textit{infinite capacity receiver}? %
Across 1.2k concepts across eight qualitative domains \citep{spangher}, we identify (a) concepts that can already be articulated between language models or where transmission already reaches a threshold; we find these concepts to be determined primarily by local, self-contained patterns (e.g. setup$\to$punchline joke mechanics); (b) concepts that \textit{may} be articulable between future, more powerful language models or where scaling laws predict eventual transmission; we find these concepts to be primarily related to whole-text structural patterns (e.g. worldbuilding coherence) and (c) concepts that will likely remain tacit or where scaling laws plateau; we find these concepts are driven by \textit{``identity and social-group awareness''} (Section~\ref{sec:scaling}). Interestingly, humans do not face low levels of agreement on the same set of bounded concepts that LLMs to; in fact, we find that humans have moderate-to-high levels of agreement in $76$\% of LLM-bounded concepts. This shows that our claims on articulation limits are \textit{executor} relative, an executor being the model that runs a prompt (i.e. here, the transmitter).

Finally, we explore mechanisms that increase transmission in all three cases. In the first two sets of concepts, where transmission is possible, we find that \textit{explicit}, direct articulations like explanations, mechanics and descriptions increase transmission the most in optimal messages. In the third case, where transmission appears less feasible, we had hypothesized that, while explicit articulation might fail, \textit{tacit} knowledge could be transmitted through examples or demonstrations; while we find some evidence of this, it is limited and we are unable to conclude this form of tacit knowledge can increase transmission. Taken together, these boundaries illuminate possible shortcomings of language technology and directions of improvement.

\begin{figure}[t]
\centering
\resizebox{\textwidth}{!}{%
\begin{tikzpicture}[
  card/.style={draw=black!60, fill=white, rounded corners=1.5pt, font=\tiny, inner sep=2pt, minimum width=12.5mm, minimum height=6.5mm, align=left},
  bitR/.style={draw=red!60!black, fill=red!30, font=\tiny\bfseries, inner sep=0pt, minimum width=4.2mm, minimum height=4.2mm},
  bitG/.style={draw=green!50!black, fill=green!25, font=\tiny\bfseries, inner sep=0pt, minimum width=4.2mm, minimum height=4.2mm},
  llm/.style={draw=blue!60!black, fill=blue!10, rounded corners=4pt, font=\small, align=center, inner sep=4pt},
  pcard/.style={draw=orange!80!black, fill=yellow!25, rounded corners=1.5pt, font=\tiny, inner sep=2.5pt, align=center, minimum width=13mm},
  ocard/.style={draw=orange!80!black, fill=orange!25, rounded corners=1.5pt, font=\tiny, inner sep=2.5pt, align=center, minimum width=13mm},
  hdr/.style={font=\scriptsize\bfseries, anchor=north west},
  lab/.style={font=\scriptsize\itshape, align=center},
  tlab/.style={font=\tiny\itshape, align=center},
  arr/.style={-{Stealth[length=1.8mm]}, thick},
]
\draw[black!40, fill=black!4, rounded corners=3pt] (0,-2.45) rectangle (4.55,2.0);
\draw[black!40, fill=black!4, rounded corners=3pt] (4.95,-2.45) rectangle (9.35,2.0);
\draw[black!40, fill=black!4, rounded corners=3pt] (9.75,-2.45) rectangle (14.3,2.0);
\node[hdr] at (0.1,1.9)  {(1) encode examples with the prompt $p$};
\node[hdr] at (5.05,1.9) {(2) re-derive the prompt $p'$};
\node[hdr] at (9.85,1.9) {(3) calculate MI};
\node[card] (x1) at (0.95,1.03)  {``the service\\was slow...''};
\node[card] (x2) at (0.95,0.23) {``arrived hot\\and fresh...''};
\node[card] (x3) at (0.95,-0.57) {``would not\\go back...''};
\node[lab] at (0.95,-1.44) {prompt $p$};
\node[pcard] (pp) at (0.95,-1.8) {``mark complaints''};
\node[llm] (enc) at (2.6,0.23) {transmitter\\LLM $T$};
\draw[arr] (x1.east) to[out=0,in=140] (enc.140);
\draw[arr] (x2.east) -- (enc.west);
\draw[arr] (x3.east) to[out=0,in=220] (enc.220);
\draw[arr] (pp.east) to[out=0,in=-110] (enc.south);
\node[bitR] (b1) at (3.85,1.03)  {1};
\node[bitG] (b2) at (3.85,0.23) {0};
\node[bitR] (b3) at (3.85,-0.57) {1};

\draw[arr] (enc.east) to[out=30,in=180] (b1.west);
\draw[arr] (enc.east) -- (b2.west);
\draw[arr] (enc.east) to[out=-30,in=180] (b3.west);
\node[tlab, align=left, anchor=west] at (2.7,-1.75) {
	\colorbox{red!30}{1} = complaint\\[0.4mm] \colorbox{green!25}{0} = ok
};
\node[card, minimum width=11.5mm] (y1) at (5.85,1.03)  {``the service\\was slow...''};
\node[bitR, right=0.8mm of y1] (c1) {1};
\node[card, minimum width=11.5mm] (y2) at (5.85,0.23) {``arrived hot\\and fresh...''};
\node[bitG, right=0.8mm of y2] (c2) {0};
\node[card, minimum width=11.5mm] (y3) at (5.85,-0.57) {``would not\\go back...''};
\node[bitR, right=0.8mm of y3] (c3) {1};
\node[llm] (dec) at (8.2,0.23) {receiver\\LLM $R$};
\draw[arr] (c1.east) to[out=0,in=140] (dec.140);
\draw[arr] (c2.east) -- (dec.west);
\draw[arr] (c3.east) to[out=0,in=220] (dec.220);
\node[lab] at (8.28,-1.3) {recovered $p'$};
\node[ocard] (pr) at (8.2,-1.8) {``flag negative\\experiences''};
\draw[arr] (dec.south) to[out=-140,in=170] (pr.west);
\node[pcard, minimum width=7mm] (q1) at (11.15,1.24) {$p$};
\node[tlab] (mw) at (11.15,0.64) {$m$};
\node[bitR] at (11.15,0.14) {1};
\node[bitG] at (11.15,-0.31) {0};
\node[bitR] (r1e) at (11.15,-0.76) {1};
\draw[arr] (q1.south) -- (mw.north);
\draw[arr] (mw.south) -- (11.15,0.35);
\node[ocard, minimum width=7mm] (q2) at (12.9,1.24) {$p'$};
\node[tlab] (mp) at (12.9,0.64) {$m'$};
\node[bitR] at (12.9,0.14) {1};
\node[bitG] at (12.9,-0.31) {0};
\node[bitG] (r2e) at (12.9,-0.76) {0};
\draw[arr] (q2.south) -- (mp.north);
\draw[arr] (mp.south) -- (12.9,0.35);
\node[draw=black!60, fill=white, rounded corners=2pt, font=\scriptsize, inner sep=3pt] (mi) at (12.05,-1.82) {$I(m;\, m')$};
\draw[arr] (r1e.south) to[out=-90,in=120] ($(mi.north)+(-0.25,0)$);
\draw[arr] (r2e.south) to[out=-90,in=60] ($(mi.north)+(0.25,0)$);
\end{tikzpicture}}%
\caption{\textbf{Reconstruction bottleneck.} We measure how well a message (i.e. the prompt $p$) transmits from a transmitter LLM $T$ to a receiver LLM $R$. $T$ acts as an encoder and the receiver $R$ as a decoder; (1) $T$ annotates each
example under $p$; its output is shown as colored bits
(\colorbox{red!30}{\tiny\textbf{1}}~= complaint, \colorbox{green!25}{\tiny\textbf{0}}~= ok).
(2) $R$ sees only the annotated examples and reconstructs the prompt $p'$. (3) Each prompt induces a labeling of the same examples ($p \to m_\omega$, $p' \to m'$);
their dependence yields the mutual-information $I(m_\omega; m')$. We report mutual information and agreement (Section~\ref{sec:reconstruction-alg}).}
\label{fig:reconstruction}
\end{figure}
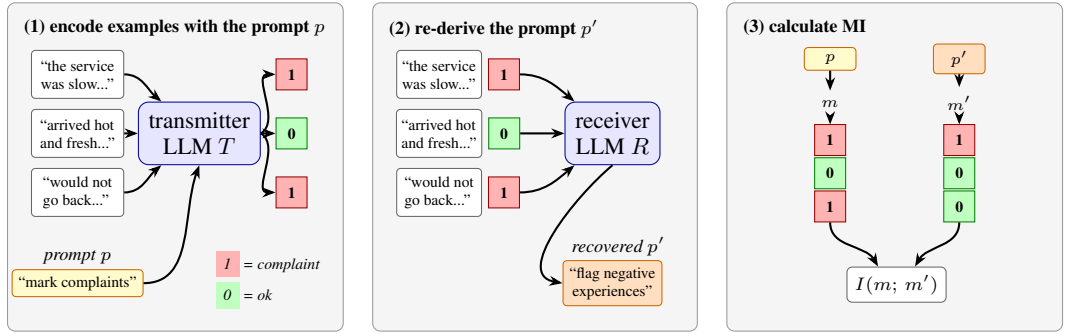

\begin{figure}[t]
	\centering
		\resizebox{\textwidth}{!}{%
		\begin{tikzpicture}[
			chip/.style={draw=blue!60!black, fill=blue!12, rounded corners=1.5pt, font=\fontsize{9}{10.5}\selectfont, inner sep=1.8pt, minimum width=5mm},
			sbox/.style={draw=black!45, fill=black!5, rounded corners=2pt, minimum width=34mm, minimum height=13mm},
			pt/.style={circle, draw=black!70, inner sep=1.5pt},
			lab/.style={font=\fontsize{10}{12}\selectfont\itshape, align=center},
			hdr/.style={font=\fontsize{10}{12}\selectfont\itshape, align=center, anchor=south, text=black, text width=50mm},  %
			tlab/.style={font=\fontsize{8}{9.5}\selectfont\itshape, align=center},
			blab/.style={font=\fontsize{7}{8.5}\selectfont\itshape, align=center},   %
			arr/.style={-{Stealth[length=1.8mm]}, thick},
			darr/.style={-{Stealth[length=1.6mm]}, thin, dashed, black!55},
			]
			\def\blob{(0,0) .. controls (0.15,1.05) and (1.1,1.65) .. (2.0,1.5)
				.. controls (2.9,1.35) and (3.35,0.5) .. (3.05,-0.35)
				.. controls (2.8,-1.15) and (1.7,-1.65) .. (0.85,-1.35)
				.. controls (0.15,-1.1) and (-0.12,-0.6) .. cycle}
			\begin{scope}
				\clip \blob;
				\fill[black!12] \blob;
				\fill[pattern=north east lines, pattern color=red!60] \blob;
				\fill[blue!25]   (1.05,0.45) circle (0.60);
				\fill[green!25]  (1.85,0.25) circle (0.62);
				\fill[violet!25] (1.35,-0.55) circle (0.55);
				\draw[blue!60!black]   (1.05,0.45) circle (0.60);
				\draw[green!50!black]  (1.85,0.25) circle (0.62);
				\draw[violet!60!black] (1.35,-0.55) circle (0.55);
				\foreach \x/\y in {0.35/-0.9, 2.6/0.9, 2.75/-0.5, 0.45/1.05, 2.45/-1.0}
				\fill[red!70!black] (\x,\y) circle (0.85pt);
			\end{scope}
			\draw[black!60, thick] \blob;
			\node[tlab, blue!60!black]   at (0.82,0.62)  {$s_1$};
			\node[tlab, green!50!black]  at (2.12,0.32)  {$s_2$};
			\node[tlab, violet!60!black] at (1.35,-0.62) {$s_3$};
			\node[hdr] at (1.55,2.25) {Step 1: Sampling Units \\(i.e. minimal strings that impact prompts)};
			\node[tlab, red!70!black, align=left, anchor=west] at (2.6,-1.35) {never-sampled\\mass};
			\draw[darr, red!70!black] (2.85,-1.15) -- (2.7,-0.7);
            \node[tlab, black, align=left, anchor=north, font=\fontsize{8.5}{9}\selectfont\itshape] at (1.6,-1.65){Sampler streams:};
			\node[tlab, black!60, align=left, anchor=north] at (1.6,-1.95)
			{$s_1$: Mining explicit metrics online\\ $s_2$: Mining from GEPA trajectories\\ $s_3$: Proposing from LLM};
			\draw[arr] (3.45,0.1) -- node[lab, above] {mine} (4.25,0.1);
			\node[chip] (o1) at (4.8,0.9)  {$u_1$};
			\node[chip] (o2) at (5.5,0.9)  {$u_2$};
			\node[chip] (o3) at (4.8,0.4)  {$u_3$};
			\node[chip] (o4) at (5.5,0.4)  {$u_4$};
			\node[chip] (o5) at (4.8,-0.1) {$u_5$};
			\node[chip] (o6) at (5.5,-0.1) {$u_6$};
			\node[draw=black!60, rounded corners=2pt, fit=(o1)(o6), inner sep=2.5pt] (pool) {};
			\node[lab, above=0.5mm of pool] {pool $\Omega$};
			\node[chip, draw=red!70!black, dashed, fill=red!8, text=red!70!black] (ux) at (4.8,-0.78) {$u_?$};
			\node[tlab, red!70!black, right=1mm of ux] {unseen unit};
			\draw[black!45, fill=black!5, rounded corners=2pt] (7.05,0.25) rectangle (10.15,2.05);   %
			\draw[black!45, fill=black!5, rounded corners=2pt] (7.05,-1.75) rectangle (10.15,-0.15);    %
			\draw[black!45, fill=black!5, rounded corners=2pt] (7.05,-3.8) rectangle (10.15,-2.2); %
			\node[hdr] at (8.6,2.25) {Step 2: Searching Over Prompts:\\(i.e. finding optimal combinations of units)};
			\draw[fill=white, draw=black!70, rounded corners=1pt] (7.25,1.2) rectangle (7.85,2.0);
			\foreach \yy in {1.77,1.6,1.43} \draw[blue!60!black, line width=1.1pt] (7.33,\yy) -- (7.77,\yy);
			\draw[arr] (7.95,1.6) -- (8.6,1.6);
			\foreach \x/\y/\c in {8.85/1.75/green!60, 9.2/1.75/red!40, 9.55/1.75/red!40, 8.85/1.4/red!40, 9.2/1.4/green!60, 9.55/1.4/red!40}
			\node[pt, fill=\c] at (\x,\y) {};
			\node[blab] at (7.6,0.72) {$p^{1}_{A_1}$\\(prompt)};
			\node[blab] at (9.2,0.72) {$m^{1}_{A_1}$\\(measurements)};
			\draw[fill=white, draw=black!70, rounded corners=1pt] (7.25,-1.1) rectangle (7.85,-0.3);
			\foreach \yy in {-0.53,-0.7,-0.87} \draw[blue!60!black, line width=1.1pt] (7.33,\yy) -- (7.77,\yy);
			\draw[arr] (7.95,-0.7) -- (8.6,-0.7);
			\foreach \x/\y/\c in {8.85/-0.55/green!60, 9.2/-0.55/green!60, 9.55/-0.55/red!40, 8.85/-0.9/red!40, 9.2/-0.9/green!60, 9.55/-0.9/red!40}
			\node[pt, fill=\c] at (\x,\y) {};
			\node[blab] at (7.55,-1.45) {$p^{2}_{A_2}$};
			\node[blab] at (9.2,-1.45) {$m^{2}_{A_2}$};
			
			\draw[fill=white, draw=orange!80!black, thick, rounded corners=1pt] (7.25,-3.15) rectangle (7.85,-2.35);
			\foreach \yy in {-2.58,-2.75,-2.92} \draw[blue!60!black, line width=1.1pt] (7.33,\yy) -- (7.77,\yy);
			\draw[arr] (7.95,-2.75) -- (8.6,-2.75);
			\foreach \x/\y/\c in {8.85/-2.6/green!60, 9.2/-2.6/green!60, 9.55/-2.6/green!60, 8.85/-2.95/green!60, 9.55/-2.95/green!60}
			\node[pt, fill=\c] at (\x,\y) {};
			\node[pt, fill=red!40, draw=orange!80!black, dashed, thick] (flip) at (9.2,-2.95) {};
			\node[blab] at (7.55,-3.5) {$p^{*}_{S}$};
			\node[blab] at (9.2,-3.5) {$m^{*}_{S}$};
			\draw[arr, blue!60!black] (7.55,0.25) -- (7.55,-0.15);
			\node[tlab, blue!60!black, anchor=west] at (7.75,0.05) {add / drop / swap units};
			\draw[blue!60!black, thick] (7.55,-1.75) -- (7.55,-1.88);
			\node[font=\fontsize{11}{13}\selectfont, blue!60!black] at (7.55,-1.99) {$\vdots$};
			\draw[arr, blue!60!black] (7.55,-2.08) -- (7.55,-2.2);
			\draw[arr] (pool.east) -- node[tlab, above, sloped, pos=0.55, inner sep=1pt] {assemble} (7.05,1.7);
			\draw[-{Stealth[length=1.6mm]}, thin, dashed, red!70!black,
			decorate, decoration={snake, amplitude=0.8mm, segment length=5mm, post length=3mm}]
			(ux.south) .. controls (4.7,-4.2) and (12.5,-5.2) .. (flip.east);
			\node[tlab, red!70!black, align=center] at (4.8,-3.5) {could flip\\ a label?};
			\begin{scope}[shift={(11.8,-1.9)}]
				\draw[arr] (0,0) -- (0,3.5);  \draw[arr] (0,0) -- (3.45,0);
				\node[tlab, rotate=90] at (-0.32,1.7) {task performance};
				\node[tlab] at (1.72,-0.62) {\# units in the prompt};
				\draw[thick, blue!60!black] plot[smooth] coordinates {(0.05,0.1) (0.5,1.1) (1.0,1.6) (1.6,1.9) (2.0,2.0)};
				\draw[thick, blue!60!black, dashed] plot[smooth] coordinates {(2.0,2.0) (2.4,2.16) (2.75,2.26)};
				\draw[dashed, black!60] (0,2.0) -- (2.0,2.0);
				\node[tlab, black!60, anchor=west] at (0.08,1.86) {$V(S)$};  %
				\draw[dashed, orange!70!black] (0,2.26) -- (2.75,2.26);
				\draw[red!70!black, decorate, decoration={brace, amplitude=1.2mm}] (2.85,2.26) -- (2.85,2.0);
				\node[tlab, red!70!black, anchor=west, align=left] at (2.98,2.09) {$\widehat\epsilon = $};
				\node[tlab, red!70!black, anchor=west, align=left] at (1.9,1.77) {$\widehat V(\widehat N) - V(S)$};
				\draw (2.0,-0.06) -- (2.0,0.06);  \node[tlab] at (1.78,-0.24) {$S$ units seen};
				\draw[blue!60!black, decorate, decoration={brace, mirror, amplitude=1mm}] (2.73,0.08) -- (2.02,0.08);
				\node[tlab, blue!60!black, align=center] at (2,0.58)
				{unseen units\\$P(\mathrm{next\ unit\ is\ new})=\frac{f_1}{n}$};
				\draw (2.75,-0.06) -- (2.75,0.06);
				\node[tlab, orange!60!black, anchor=west] at (2.82,-0.24) {$\widehat N$ total};
				\node[hdr] at (2.05,4.15) {Step 3: Prompt Upper Bounds:\\(i.e. estimating the value of unseen units)};  %
			\end{scope}
	\end{tikzpicture}
    }
    \vspace{-8mm}
	\caption{\textbf{\textit{Optimizing the message}: The $\epsilon$-estimation algorithm.} \underline{Left:} \emph{Step 1, Sampling Units:} We discover \textit{semantic units} with samplers $s_1$, $s_2$, $s_3$ sampling from the space of all possible units. Units are minimal prompt fragments that causally alter behavior, e.g. for the concept ``Concise Wording'', units might include: \smash{\unitchip{short, exact entity phrase}}, \smash{\unitchip{no filler words}}. Each probes a region of this space, yielding the pool $\Omega$, and the red is space no sampler has reached. \underline{Center:} \emph{Step 2, Searching Over Prompts:} we develop an iterative search algorithm to assemble units into prompts $p^1_{A_1}, p^2_{A_2}, \ldots, p^{*}_{S}$; each labels the same datapoints $m^{i}_{A_i}$; unseen units $u_?$ may have flipped a label in $m^{*}_{S}$. \underline{Right}: \emph{Step 3, Estimating Prompt Upper Bounds:} We estimate the total \# of units $\smash{\widehat{N}}$ with capture-recapture \citep{chao1984nonparametric,chiu2014improved} and the value $V$ of the optimal prompt by extrapolating  $V(\smash{\widehat N}) = \smash{\widehat\epsilon} + V(S)$.}
	\label{fig:mining-bound}
	\label{fig:method-overview}
\end{figure}
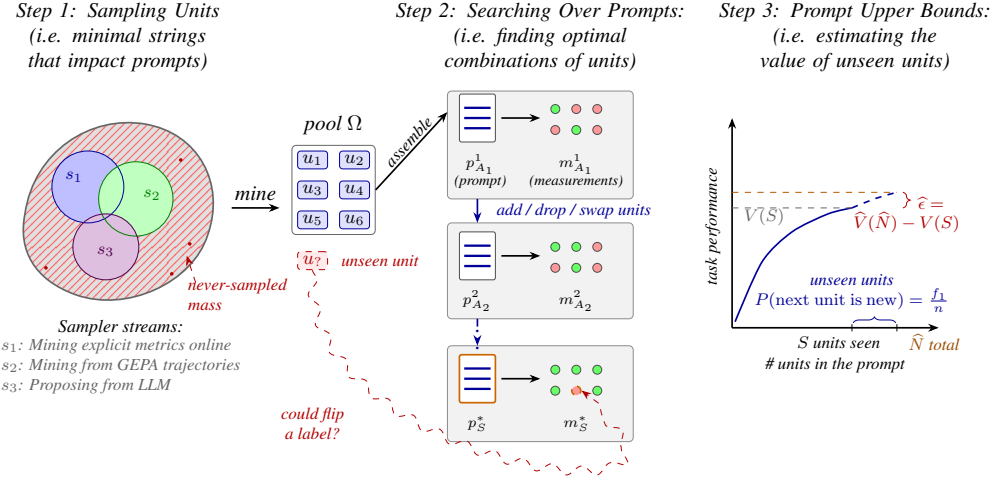

\section{The \textit{Transmitter}-\textit{Message}-\textit{Receiver} Pipeline}

In our setting, a \textit{transmitter} $T$ attempts to transmit concepts through a channel to a \textit{receiver} $R$, with a message, or a prompt $p$. We call the degree to which $R$ recovers the concept the \textit{articulability} of that concept, with respect to $T$, $R$ and $p$. An \textit{articulability upper bound} for a concept is the limit of its transmissibility over optimal messages and $T$ and $R$ of infinite capacity: concepts that are \textit{more} articulable can be transmitted between $T$ and $R$ with lower capacity and with messages that are worse; concepts that are \textit{less} articulable (e.g. ``the aroma of coffee''
) are the opposite.
We start by introducing methods for finding \textit{optimal messages}: holding the transmitter and receiver  fixed, we seek to \textit{estimate} the performance of the \textit{optimal} prompt for a concept (Section~\ref{sec:e_certifiability}). We then introduce methods for measuring the channel between $T$ and $R$ (Section~\ref{sec:reconstruction}). Finally, we analyze articulation upper bounds of concepts by scaling $p$, $T$ and $R$ (Section~\ref{sec:scaling}). %

\subsection{The \textit{Message}: Finding Optimal Messages with $\epsilon$-estimation}
\label{sec:e_certifiability}

Our goal in this section is to develop a search algorithm that finds prompts optimal for a given LLM up to a tolerance $\epsilon$ at a confidence level $\delta$. Existing optimizers such as GEPA improve prompts but do not estimate missing performance. Given a task (e.g. ``\textit{Measure: `Concise Wording' in text}'') and a measurement function $f$ 
, let $S = \{s_1, s_2, \ldots\}$ be different \emph{sampler streams} finding \textit{units} from $\mathcal{U}$, a space of all units. %
A \textit{unit}, defined in the next section, is a string that minimally impacts the behavior of a prompt. The pool of recoverable units, from $\bigcup_{i} s_i$ is $\Omega \subseteq \mathcal{U}$. $p_A$ is the prompt composed of $A$, a finite subset of units $A \subseteq \Omega$.\footnote{The total population size $|\mathcal{U}|$ of discoverable steps is estimated in Section~\ref{sec:missing-value-est}.} Prompts $p_{A}$ are assembled from the pool $\Omega \subseteq \mathcal{U}$ such that:
\begin{equation}
f(p^*_S) \;=\; \max_{A \subseteq \Omega} f(p_A); \quad \Pr\!\big[\, f(p^*_S) - f(p_{A}) \le \epsilon \,\big] \;\ge\; 1 - \delta .
\label{eq:certificate}
\end{equation}
In other words, prompt $p_A$ is within $\epsilon$ of optimal prompt $p^*_S$ for a given LLM. Eq.~\ref{eq:certificate} is supremum over our sampling draws. Two properties make this tractable. $\mathcal{U}$ is a countable population whose members can be re-identified across sampling runs\footnote{A candidate \textit{unit} enters $\mathcal{U}$ only if it repeatably and significantly changes measured behavior (Appendix \ref{app:unit-gate}). Each unit is a \textit{species} in \textit{capture--recapture} \citep{chao1984nonparametric} and units are strings matched across runs by word overlap (Appendix~\ref{app:mass}); \textit{capture--recapture} allows us to estimate the total number of missing units.} and $f(p_A)$ is well-defined and reproducible (Appendix~\ref{app:submodular}). Second, because $\mathcal{U}$ is a population, its unseen portion is estimable: the overlap between independent sampling runs yields a capture--recapture estimate of how many units remain undrawn. We fit a value curve to estimate the value of these unseen units, which we use to calculate $f(p^*_S)-f(p_{A})$, the \textit{within-$\Omega$ search gap}\footnote{Assumptions are: (A1) our samplers cover the unit universe of interest; (A2) a unit's discovery probability is informative about its value (Appendix~\ref{app:mass}); (A3) units are exchangeable and replicate (even though they are found conditionally by our sampler); (A4) the sampled submodularity ratio stands in for the population ratio, which it can only overstate; (A5) the value curve saturates, which the fitted forms impose rather than test. Assumptions addressed in more detail in Appendix~\ref{app:eps-assumptions}.}. This formalization has clear limits (e.g. prompts are not compositional, units might interact) which we discuss; in practice we find these do not impact our conclusions. See Appendix~\ref{app:submodular} and ~\ref{app:ecert-results} for details, proofs and limits.

\paragraph{Unit mining: Creating the Search Space.}
\label{sec:unit-mining}

To construct a search space over prompt variations, we define a \textit{semantic unit} as a minimal string that, when appended to existing prompts, causally improves the executor's performance (e.g. for the concept \textit{Concise Wording}, units could be \unitchip{no filler words} or \unitchip{specific formats}). Candidates sampled from $S$ are added to $\Omega$; they can come from multiple sampling sources; here: (1) $s_1$: retrieve units online, (2) $s_2$: ask an LLM directly for suggestions, and (3) $s_3$: extract from GEPA-optimized trajectories. For each, we measure $\Delta_i = E_{p\in prompts}[\Delta | p \oplus i]$, or their effect averaged across prompt variations. Units with $\Delta_i \leq 0$ are dropped from $\Omega$.
A unit that \textit{does not} repeatedly impact the prompt is discarded; one that \textit{does} is broken down recursively until it stops impacting the prompt, giving \textit{minimal} units.
This atomizes our space to make it searchable. We note that atomizing our prompts as such discards many kinds of possibly useful prompt perturbations, like the prompt's tone; thus, $\epsilon-$estimation estimates prompt optima only with respect to our search space and sampling procedures; this, we show, is already enough to beat SOTA optimizers.
\footnote{Many prompt changes are deeply interdependent and nonlinear (i.e. a prompt change might only work if other components of the prompt already exist). We note that unit interdependence is \textit{partially} recovered inside the search, described in Section \ref{sct:exploration}: units are re-tested conditionally when added to the current prompt during the prefix sweep and drop-one passes, so pairwise and prefix-level interactions are accounted for; only perturbations that never decompose into units (e.g. tone, persona) fall outside the space.}

\paragraph{Unit-Space Exploration: Optimizing the Prompt.}
\label{sct:exploration}

To find the optimal prompts from $\Omega$, we sample sets of units and use a simple template to format them into a prompt (i.e. \texttt{``\textless Task Name\textgreater: \textless unit $i_1$ \textgreater, \textless unit $i_2$ \textgreater, \textless unit $i_3$\textgreater ...}) in random order. We run each prompt $T$ times and average the performance across all variations. We then add units as follows: we measure each unit's marginal effect on the initial prompt, add units with high marginal value, and keep each addition only if whole prompt's score improves. We then try deleting each kept unit, removing any whose deletion costs nothing, and, in a second pass, try re-adding any skipped units that now help. We perform a check on held-out data to measure whether the prompt beats our initial prompt; if it does not, we keep the initial prompt. This is a double greedy sweep followed by a pruning-and-recovery pass, inspired by \citet{nemhauser1978analysis, feige2011maximizing}.
Classical guarantees on the optimality of our search result\footnote{For example, forward greedy is guaranteed to attain at least a $(1-1/e)\approx 63\%$ fraction of the optimal subset's value \citep{nemhauser1978analysis}, and double greedy a $\tfrac{1}{2}$ fraction \citep{buchbinder2015tight}.} 
would apply if our score function were submodular and monotone. Ours is neither: adding a unit could decrease performance \textit{and} a unit's value can depend on units are already present. We estimate how far our objective departs from submodularity by estimating the worst-case relationship between adding units individually vs. together (i.e. the submodularity ratio $\gamma$) and find $\hat\gamma\approx1$, i.e. \textit{combining} units rarely beats adding each separately; thus the observed gaps are nearly as tight as for submodular objectives \citep{das2011submodular,bian2017guarantees}\footnote{Submodularity ratio $\gamma$ is a sampler-relative minimum; it might miss rarer, worse pairs in the full population and \textit{overestimate} the true population $\gamma$. It tightens with more pairs. (Appendix~\ref{app:submodular}; \citealp{minoux1978accelerated}).}.

\paragraph{Missing-value estimation: Projecting an Upper Bound.}
\label{sec:missing-value-est}

Finally, we estimate the impact of units our sampling streams have not yet reached. We estimate: \textit{missing mass}, or how many units we have not yet found; and \textit{missing values}, or how much performance units give. For missing mass, we treat new samples as independent draws from the sampling process\footnote{In practice, each sampler samples units as one batch, so draws are not fully independent; (Appendix~\ref{app:mass})}, if new samples are units we have already found, the space is nearly exhausted; otherwise, it is not. A capture--recapture estimate $\widehat N=f_1^2 / 2f_2$ gives the total unit count, where $f_1$ is the total number of units seen once and $f_2$ is the total number of units seen twice\footnote{In practice, we use $\widehat N = S + f_1(f_1-1)/(2(f_2+1))$ as an unbiased lower-bound estimator (Appendix~\ref{app:mass}).} \citep{chao1984nonparametric}. $\widehat N$ estimates the \textit{minimum} number of distinct units consistent with that sampling and thus errs low: this gives a \textit{lower} bound on richness \citep{chiu2014improved}.
Next, to estimate missing value from missing mass, we fit a curve $\widehat V(k)$ to the value of units found so far and extrapolate it over unseen units, from the $S$ seen units to the estimated total $\widehat N$; this is relative to our search. $\epsilon$-estimation's two parameters, $\delta$ and $\epsilon$, trade off: $\delta$ is declared a-priori (by default, $.05$), while $\widehat\epsilon$ is the fitted value $\widehat V(\widehat N)-V(S)$. Sampling shrinks the unseen part of the search space, $\widehat N - S$ and sampling more trajectories reduces the noise in each $V(k)$ point. In the limit, $\widehat\epsilon \to 0$ at any $\delta$ only if the unit space $\Omega \to \mathcal{U}$ (Appendix~\ref{app:mass}). The risk exists that an unseen unit has unpredictably high effect\footnote{Definitively ruling this out requires assumptions on the tail of our unit-value distribution; our attempt to fit that tail directly degenerated, with a Pareto model of the largest observed unit values; Appendix~\ref{app:backtest}.}; in practice, we backtest on all trajectories and find negligible prediction error up to eight units ahead (Section \ref{sct:results:e-cert}, Appendix~\ref{app:backtest}). Finally, we note that these estimates are conditional on our declared samplers: they estimate the value of what \emph{this} sampling process has not yet drawn, not what no sampling process could ever find. We show examples of missing value measurement in Figure~\ref{fig:missing-value}.

\subsection{The \textit{Transmitter-Receiver}: Reconstruction Bottleneck}
\label{sec:reconstruction}

In the previous section, we introduced $\epsilon$-estimation, which allowed us to optimize prompts for a given language model, in terms of \textit{performance} on benchmarks. However, this approach collapses the \textit{transmitter-message-receiver} system by ignoring the receiver, letting it hide behind benchmark labels. Whether concepts can be \textit{successfully articulated} in a way that is \textit{understood} is still open. %
Consider an LLM-judge using a rubric to assess an agents' output and give rewards to the agent: just because the \textit{judge} can be shown to comprehend concepts in the rubric (say, on a relevant benchmark) does not mean that the \textit{agent} will understand the judge's output. The \textit{articulation} of the concept requires both the transmitter (the judge) and receiver (the agent) to find agreement.

\paragraph{\textit{Transmitter-Receiever} Agreement}
\label{sec:reconstruction-alg}

We now introduce an algorithm that can test whether the information provided by the transmitter can be comprehended by the receiver. Consider the following: imagine a language game \citep{wittgenstein1953pi} where Person A points to 10 objects and calls out ``Red!'', ``Blue!'', ``Blue!'', ``Red!''. Person B, who does not speak the same language, upon hearing A, tries to deduce A's decision function. After a while, B realizes that the \textit{concept} must be ``color''\footnote{Philosophically, this game, i.e. \textit{ostensive definition} is underdetermined; finite examples can support infinitely many candidate rules \citep{wittgenstein1953pi, quine1960word}. In our framework, this ambiguity is resolved because both agents share a restricted hypothesis space over possible concepts.}. How can we measure whether B understands the concept\footnote{If we resort to language, i.e. directly asking B what concept A was transmitting, this can capture surface-form similarities without capturing real meaning: the labeled examples carry lexical traces of how A's rule was phrased, so B can produce words that \emph{sound} like the rule, and thus seem more similar, without being able to apply the rule to new objects (Appendix~\ref{app:recon-alternatives})}? If both B and A can now point to another 10 objects and jointly call out ``Blue!'', ``Red!'', thereby finding agreement, B must understand A's decision rule to the degree B can operationalize it. 

More formally, as shown in Figure \ref{fig:reconstruction}, our algorithm measures:
\begin{equation}
\mathrm{I}\big(m;\, m'\big)
\quad \text{where} \quad
m_i = T(x_i;\, p), \qquad
p' = R\big(\{(x_i, m_i)\}_{i \in D}\big), \qquad
m'_j = T(x_j;\, p'), \;\; j \notin D .
\label{eq:reconstruction}
\end{equation}
Given a label stream $m_1, m_2, \ldots \in \{0,1\}$ that $T$ (i.e.\ person A) produces under $p$ (e.g. ``color''), we ask $R$ (i.e.\ person B) to observe $D$ labeled examples and reconstruct the prompt $p'$ that would have generated them. We then measure \textit{transmission recovery} by scoring held-out datapoints with $p'$, producing $m'$, and measuring the mutual information between the two labelings.\footnote{Since a receiver whose reverses the transmitter's labels would score $I(m;m')=H(m)$, we report balanced accuracy alongside MI: we find MI and balanced accuracy correlate at Spearman $.83$ and reversed reconstructions carry $1.3\%$ of the MI. (Appendix~\ref{app:recon-instrument}). In addition, the model that scores using the reconstructed prompt $p'$ does not impact much. We use: receiver $R$ writes prompt $p'$ from the labeled examples, transmitter $T$ runs $p$ \textit{and} $p'$ on the held-out items to give $m$ \textit{and} $m'$. When the $R$ runs $p'$, recovery is lower ($.040$ vs. $.052$ bits over $55$ criteria) (Appendix~\ref{app:recon-alternatives}).} Reconstruction improves on alternatives like consistency, where a labeling is scored by its stability across samples or judges\footnote{Consistency has two problems: (1) it is satisfied by rules that are degenerate, e.g. by labeling everything $1$ (reconstruction does not share this: the labeling must survive transmission through language and re-execution on unseen items; a degenerate stream has no information to transmit) and (2) consistency is satisfied by rules that are consistently wrong (in reconstruction, an inconsistent labeling will not generalize but a consistently wrong rule does, hence the planted controls). This protection is partial: a simple rule such as ``longer than the median'' generalizes to held-out items, so recovery shows that the transmitter applied \textit{some} statable rule; we use the planted rules of Section~\ref{sec:recon_results} as positive controls}; but also 
depends on assumptions: (i) the rule generating $m$ lies in the class of rules the transmitter and receiver can express; (ii) distinct rules agree on an item with likelihood $< 1$, so the chances that $n$ examples support a false interpretation decays exponentially; and (iii) the concept the transmitter labels is the one its prompt expresses. Reconstruction can combine with other unsupervised measures, e.g. consistency or semantic similarity\footnote{We tested simply comparing prompt $p'$ with $p$ directly, or asking the receiver to identify $p$ among distractors, as the reconstruction objective first; we found the relabeling channel the most robust. Appendix \ref{app:recon-alternatives}.} between $p$ and $p'$.%

\paragraph{Setting ceilings and floors for the reconstruction objective.} How should recovery values be interpreted? First, there exists an \textit{informational} upper bound, $I(m;m') \le H(m)$: prompts with low label entropy have little information to transmit. We report $H(m)$ alongside every recovery $I(m;m')$. Secondly, we estimate the capacity of our instruments by including \textit{planted} rubrics, or code-checkable rules whose ground truth we compute directly (e.g.\ ``this text contains at least one question mark''). We compute a planted rule's true labels ourselves and score reconstruction $m'$ against ground truth:  every real criterion's recovery is read against this ceiling. We establish a lower capacity bound from a permutation null (i.e. shuffling labels on examples to evaluate leakage), and a no-demonstration null (i.e. presenting the receiver with just the prompt template). This gives a floor; we only report reconstruction above the larger of these two controls.

\subsection{Performance Results}
\label{sec:results}

We measure both $\epsilon$-estimation and the reconstruction bottleneck in two ways: (1) Labeled benchmarks: we consider 6 benchmarks 
that were used to evaluate GEPA \citep{agrawal2025gepa}. (2) Unlabeled metrics: we use 1.2k annotation tasks collected and released by \citet{spangher} for understanding human preferences in creative tasks. Each task represents one metric a rubric might be composed of (e.g. in Creative Writing: \textit{``Concise Wording: Precise wording for drafts.''})\footnote{Terminology: we call each of the nine datasets (humor, creative writing, press releases, math, peer review, news homepages, notice-and-comment, patents, code review) a \textit{domain}; the validation of Section~\ref{sec:recon_results} the six with human labels (humor, creative writing, press releases, math, peer review, code review); we call each of the ${\sim}5{,}000$ labeling criteria within them an annotation task or a \textit{preference metric} (\textit{metric} for short, and \textit{criterion} interchangeably in Section~\ref{sec:scaling}). Where ``task'' appears unqualified in the unsupervised setting it means annotation task; the six supervised datasets are always \textit{benchmarks}.} %
Each task consists of a name, a short description and labeled examples. %

\subsubsection{$\epsilon$-estimation Results}
\label{sct:results:e-cert}
We compare $\epsilon$-estimation with existing prompt optimization algorithms. %
Table~\ref{tab:main-results} compares $\epsilon$-estimated prompts against existing prompt optimization methods \citep{agrawal2025gepa, opsahl-ong2024optimizing}. $\epsilon$-estimation outperforms existing methods by up to 14\% at an equal call budget on HotpotQA, HoVer, PUPA and AIME; and shows no significant difference on IFBench, LiveBench. See Appendix~\ref{app:ecert-supervised} for more comparisons. We ran ablations with certain samplers, $s_i$ removed: the result scores dropped insignificantly and still beats existing methods at matched budget, so either samplers converge on a region of the space or we are recovering units to saturation. On \textit{no} benchmark is our prompt significantly worse. We also backtest our estimates for upper-bound performance and find a median prediction error ${\approx}.02$ $k=8$ units ahead; at $k_{\max}{+}50$ and $k{=}\infty$,  four of the six benchmarks change by ${\le}.01$ (Appendix~\ref{app:ecert-results}). We also observe that length is not what makes a unit valuable: over $360$ units in five pools, a unit's length and its marginal gain have no correlation (Appendix~\ref{app:message-length}).%

\subsubsection{Reconstruction Bottleneck Results}
\label{sec:recon_results}

We run the reconstruction bottleneck on $1,270$ annotation tasks released by \citet{spangher}, Figure~\ref{fig:missing-value} (right) reports results for $272$ of these.
First, we ask whether reconstruction is a \textit{useful} objective: does optimizing prompts for reconstruction MI beat optimizing them using other optimizers? We compare $m_{rec}$ (labels made by prompts optimized via the reconstruction objective) against: $m_{LLM}$ (labels made by prompts selected by an LLM-critic objective\footnote{A judge model gives feedback to guide transmission improvement, similar to \citet{agrawal2025gepa}.}), and $m_{desc}$ (labels made by a simple prompt using the criterion's description text, given by \citet{spangher}). Optimization for sender and receiver is done with $\epsilon$-estimation. Reconstruction beats an LLM critic: matched comparison gives a margin of $+.033$ AUC ($99$ wins to $54$ losses). Against $m_{desc}$, reconstruction matches but does not beat it; by \citet{spangher}'s construction, labels were annotated in part \textit{using} description text, so the description has an advantage.%

Next, we ask whether reconstruction MI has \textit{criterion validity}: as MI \textit{improves}, does independently-measured error also decrease? For each metric, we gather prompt forms across the optimization trajectory (roughly $10$--$30$ per metric) and, for each prompt, calculate MI and AUC. We then correlate MI with AUC \textit{within} each metric. Correlations run $.37$--$.68$ (Appendix~\ref{app:mi-silver}), with permutation $p<.006$. Code review, for example, correlates at $\rho=.62$. Overall, our prior two results indicate that reconstruction is a valid unsupervised metric and a more useful optimization objective than other unsupervised objectives. Transfer across model families, (prompt author, transmitter) pairs and planted-rule recovery are reported in Appendix~\ref{app:recon-instrument}.

\section{Articulation Upper Bounds: Observational Scaling Laws for Transmitter-Receivers}
\label{sec:scaling}

Now we study \textit{articulation scaling behaviors}%
. Does a concept fail to transmit, even with optimal messages, simply because the receiver does not \textit{know} enough? For example, does ``the aroma of coffee'' \textit{become} articulable to a receiver who has background knowledge (e.g. ``acidic chemistry'') or the ability to compare (e.g. ``nuttiness'' vs. ``fruitiness''; ``Kona'' vs. ``Costa Rican'')? Conversely, between \textit{infinitely capable} interlocutors, are there concepts that we cannot fully transmit?

To address this problem, we operationalize observation scaling laws in the \textit{transmitter-message-receiver} pipeline as follows. For each annotation task, we run comparisons of sender and receiver models across all pairs of $14$ models from five families (Llama, Qwen2.5, Gemma-2, Mistral, Phi) from $1.2$B to $73$B parameters. Then, for each sender-receiver pair, we optimize messages they send by running $\epsilon$-estimation until we have stable estimates of $k\to\infty$ units. We report this estimate as the transmission score. 
To compare model families across scaling sizes, we measure each model's \textit{latent} capability index, inspired by \citet{ruan2024observational}. Our measure of latent capability, $z$, is model performance on $240$ code-checkable judgments (e.g. negation, presence, threshold, paraphrase and composite checks); this, we find agrees with benchmark reports\footnote{From the the OpenCompass leaderboard, url: \url{https://rank.opencompass.org.cn/}.} at $r=.64$--$.87$. 
For each of our $1{,}270$ tasks, we fit recovery against $z$\footnote{Each metric's curve is the set of points $(z_e, y_e)$: an executor's capability index and its recovery. Two lines are fit to these points by least squares, a straight line $y=a+bz$ and a logistic $y=L\,\sigma(k(z-z_0))$ with ceiling $L\le 1.2$, steepness $k$, and midpoint $z_0$. The logistic is preferred when its Akaike information criterion (AICc) is lower \citep{hurvich1989regression}. The ceiling $L$ has a $95\%$ interval $[L_{lo}, L_{hi}]$; a curve counts as bounded only if the upper end, $L_{hi}$, is below the threshold.}; we label the curve \textit{saturating} when the fit is logistic and then sort each curve into a regime. A saturating curve is \textit{bounded} when the upper confidence limit of its ceiling is below a threshold\footnote{The threshold is the \textit{transmission of planted rules} (e.g. negation, presence, etc.) by our three best models.}
and \textit{reaches} if it is above the threshold; an unsaturated curve is \textit{rising} when it's curve is can be fit linearly
\footnote{
See Appendix~\ref{app:osl-regimes} for more formal definitions for how we divide scaling curves into each regime. The choices behind regimes include: the curve form, the executor set, the exclusions and type contrasts. Regime sizes, domain composition and three of the five type contrasts survive measurement noise in recovery and capability index.
} 
. See Appendix~\ref{app:osl-regimes}, \ref{app:z-validation} and \ref{app:regime-robustness} for more details.

\paragraph{Finding \#1: Articulation scales in three regimes: concepts that are still-climbing (\emph{rising}), attained (\emph{reaches}), and capped (\emph{bounded}); bounded concepts are socially anchored.}

Our goal here is to explore how concept transmission expands as transmitter capability, $z$, grows\footnote{$z$ is a composite of a $240$-item battery of code-checkable LLM-judgments (e.g. negation, presence, threshold, paraphrase and composite checks) with prompts optimized using reconstruction. \citet{ruan2024observational} builds its capability index from public benchmark scores. We compare this approach to similar capability measurements: our battery-$z$ agrees with the external OpenCompass leaderboard \citep{opencompass2023} at $r=.64$--$.67$ on the co-listed executors, rising to $r=.87$ ($\rho=.90$) once one diagnosed instrument-incompatible executor (Gemma-2-27B) is excluded (see Appendix~\ref{app:z-validation})}. We find, as in \citet{ruan2024observational}, that scaling curves are regular enough to predict performance at larger models: refitting each curve without its most capable executor predicts that executor's value with median error $.062$, or $9\%$ of the held-out value\footnote{Compared with baselines: our median error $.062$ compares against $.075$ for a straight line, $.051$ for a logistic fitted regardless of AICc, and $.149$ for the mean of the remaining executors (Table~\ref{tab:regime-heldout}).}, Appendix~\ref{app:osl-regimes}. Next, we examine \textit{which} annotation tasks fit into which regimes: Figure~\ref{fig:osl-types6} shows every metric annotated with attributes of how we expect it to behave\footnote{We developed a task-agnostic codebook, with $\kappa=.75$ labeling agreement to humans.}. We find that \emph{rising} criteria skew toward \emph{whole-text structural patterns} and \textit{verifiable features} ($28\%$ and\ $29\%$ of \textit{rising} metrics fit these patterns): these are long compositional constructions (e.g. ``worldbuilding coherence: rules and internal logic'', ``scene craft: objective, conflict, change, escalation'') and more complex but objective features (e.g. ``claim--evidence alignment and causal caution''). \emph{Reaches} criteria, on the other hand, tend to be \emph{local patterns within the document} ($34\%$ of \textit{reaching} metrics): these are self-contained devices a mid-scale model already recognizes completely (e.g. setup$\to$punchline mechanics, endings). Finally, \emph{bounded} criteria are distinguished by content that lives outside the artifact: identity \& social-group awareness ($23\%$ of bounded tasks) and beyond-text features ($15\%$)\footnote{Three of these five contrasts survive a permutation test: \{reaches, local\} patterns $q=.001$, \{rising, whole-text\} structure $q=.009$, \{bounded, identity and social group\} $q=.03$; Appendix~\ref{app:osl-regimes}).}. These are criteria dependent on audiences, identities, and communities, and accord with \citet{collins2010tacit} notion of collective tacit knowledge. %
For each domain, we also analyze which \textit{regime} its metrics fall into. Figure~\ref{fig:osl-domains} shows the regime$\times$domain composition directly: bounded metrics exist in humor, news homepages, peer review, and creative writing. Appendix~\ref{app:taxonomy} defines each category and gives example criteria.

\begin{figure}[t]
	\centering
\resizebox{\textwidth}{!}{%
\begin{tikzpicture}
\input{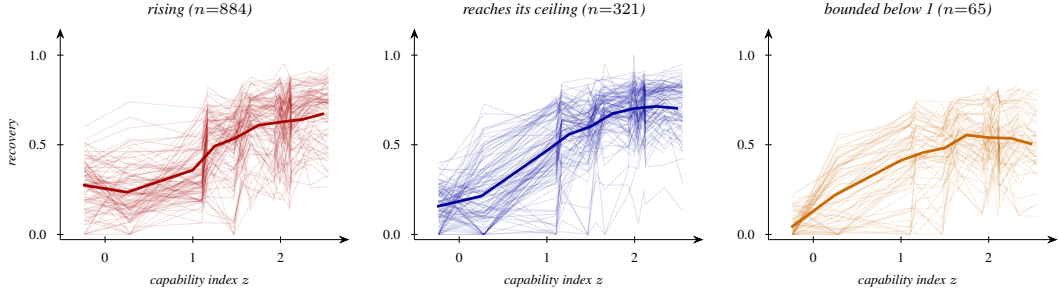}
\end{tikzpicture}}%
	\caption{\textbf{Unsupervised articulation scaling, split by fitted regime.} Each faint trace is one task's recovery against the capability index $z$ (14 executors, five model families);
		the heavy line is the within-regime binned mean. \emph{Rising} ($n{=}884$): still climbing at the largest observed $z$.
		\emph{Reaches} ($n{=}321$): the executor's estimated ceiling is attained within observed $z$. 
        \emph{Bounded} ($n{=}65$): recovery sits below the executor's ceiling's upper confidence limit, or the planted rules' mean agreement. These concepts stay partly inarticulable relative to the instrument (the $40$ planted rules are not fit).}
	\label{fig:osl-regimes}
\end{figure}

\begin{figure}[t]
\centering
\resizebox{.9\textwidth}{!}{%
\begin{tikzpicture}[tlab/.style={font=\tiny\itshape, align=center}]
\input{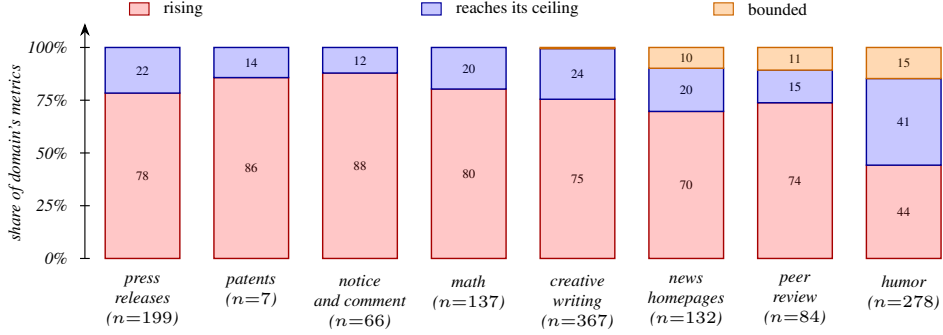}
\end{tikzpicture}}%
\caption{\textbf{Regime composition by domain} (each bar sums to $100\%$ of that domain's fitted tasks, ordered by bounded share). Bounded mass is confined to humor, peer review, news homepages, and marginally creative writing; the four remaining domains are bounded-free.}
\label{fig:osl-domains}
\end{figure}

\begin{figure}[t]
\centering
\resizebox{\textwidth}{!}{%
\begin{tikzpicture}[tlab/.style={font=\scriptsize\itshape, align=center}]
\IfFileExists{figs/gen_osl_types6.tex}{\input{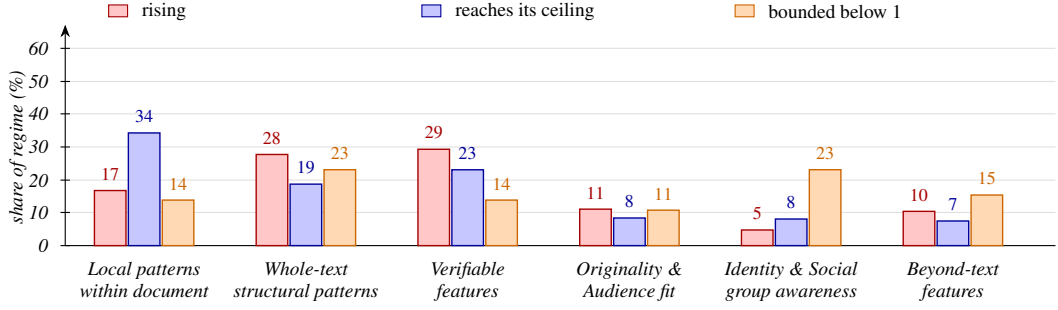}}{}
\end{tikzpicture}}%
\caption{\textbf{What kind of criterion lands in each scaling regime} ($1{,}270$ unsupervised tasks, the full $12$-category version is Appendix Fig.~\ref{fig:osl-granularity}; category definitions and example criteria are Appendix~\ref{app:taxonomy}). \textit{Rising} skews to whole-text structural patterns and verifiable features; \textit{reaches} to local patterns; \textit{bounded} tasks involve identity \& social-group awareness and beyond-text features.}
\label{fig:osl-types6}
\end{figure}

\paragraph{Finding \#2: Articulation strategies that emerge across tasks: definitions and explanations help \textit{rising} tasks; examples (i.e. \textit{showing}, not \textit{telling}) help \textit{bounded} concepts in limited cases.}

Across models, we categorize optimal prompts in terms of components: concept \textsc{name}, \textsc{definition}, mechanistic \textsc{explanation}, and a \textsc{full packet} (i.e. name, definition, explanation and few-shot examples). Doing this tells us which message form each receiver needs. First, as found in prior work \citep{wei2022emergent}, strategies emerge with scaling. At small model scales, or $\leq3$B parameters, no articulation strategy emerges and even planted mechanical rules execute at chance; at mid scale, definitions and explanations emerge as effective articulations (explanations perform best for $47\%$ of concepts at 7B); the full packet performs best at the largest receivers ($66\%$ at 70B).

Providing examples behave differently from prose. We had hypothesized that tacit meaning might transmit with in-context learning: colloquially, humans can \textit{show} even if we cannot \textit{tell}. To test this, we added examples to transmitter prompts for $449$ \textit{bounded} metrics\footnote{Examples for each prompt were selected randomly. We tried nine other ways of choosing or writing examples; Appendix~\ref{app:examples} describes each. 
}: two that the transmitter labeled YES and two it labeled NO. Examples, we found, barely outperformed the definition-only message: at Qwen2.5-14B and Llama-70B, recovery rises by $+.006$ bits, and at the three smaller transmitters it falls slightly. $43$ metrics with Llama-70B and $42$ with Qwen2.5-14B significantly improved\footnote{We calculate significance by measuring two standard deviations of the change that reconstruction noise alone produces between two runs of the same message ($.052$ bits at Qwen2.5-14B, $.055$ at Llama-70B).}, and $4$ exceed it at both models. These are thematically coherent: for example, ``delivery and timing craft'' ($+.17$ bits), ``incongruity and surprise'' ($+.03$) and ``conflict/tension as news value'' ($+.04$); they are novelty and craft metrics. Humor metrics overall have the most positive impact from in-context learning at $+.008$ and peer review least  at $+.005$. Interestingly, while examples the transmitter \textit{writes itself} raise recovery for \textit{bounded} criteria more than any other regime, the effects are model-dependent: $+.011$ for Qwen2.5-14B and $+.007$ bits for Llama-70B vs. $-.023$ for Phi-4 and $-.018$ for Qwen2.5-7B (Appendix~\ref{app:examples}). This contradicts prior in-context learning research which typically points to the benefits of providing examples \citep{brown2020language}; it's possible that the concepts we study had already been internalized by the model or sit at the edge of resolution. 

\begin{figure}[t]
\centering
\resizebox{\textwidth}{!}{%
\begin{tikzpicture}[lab/.style={font=\scriptsize\itshape, align=center},
  tlab/.style={font=\tiny\itshape, align=center}]
\begin{scope}[xscale=0.96]  %
\draw[-{Stealth[length=1.6mm]}] (-0.25,0) -- (-0.25,3.15);
\node[tlab, rotate=90] at (-0.85,1.45) {\% of concepts};
\draw (-0.31,0.00) -- (-0.19,0.00); \node[tlab, anchor=east] at (-0.34,0.00) {0};
\draw (-0.31,0.72) -- (-0.19,0.72); \node[tlab, anchor=east] at (-0.34,0.72) {25};
\draw (-0.31,1.45) -- (-0.19,1.45); \node[tlab, anchor=east] at (-0.34,1.45) {50};
\draw (-0.31,2.17) -- (-0.19,2.17); \node[tlab, anchor=east] at (-0.34,2.17) {75};
\draw (-0.31,2.90) -- (-0.19,2.90); \node[tlab, anchor=east] at (-0.34,2.90) {100};
\fill[black!15, draw=black!40, line width=0.4pt] (0.00,0.000) rectangle (0.52,1.631);
\fill[black!35, draw=black!60, line width=0.4pt] (0.00,1.631) rectangle (0.52,1.994);
\fill[blue!45, draw=blue!60!black, line width=0.4pt] (0.00,1.994) rectangle (0.52,2.809);
\fill[violet!45, draw=violet!60!black, line width=0.4pt] (0.00,2.809) rectangle (0.52,2.900);
\node[tlab, anchor=north] at (0.26,-0.06) {L1b\\ $z{=}{-}0.2$};
\fill[black!15, draw=black!40, line width=0.4pt] (0.86,0.000) rectangle (1.38,0.181);
\fill[black!35, draw=black!60, line width=0.4pt] (0.86,0.181) rectangle (1.38,0.997);
\fill[blue!45, draw=blue!60!black, line width=0.4pt] (0.86,0.997) rectangle (1.38,1.722);
\fill[violet!45, draw=violet!60!black, line width=0.4pt] (0.86,1.722) rectangle (1.38,2.538);
\fill[orange!70, draw=orange!80!black, line width=0.4pt] (0.86,2.538) rectangle (1.38,2.900);
\node[tlab, anchor=north] at (1.12,-0.06) {L3b\\ $z{=}1.2$};
\fill[black!15, draw=black!40, line width=0.4pt] (1.72,0.000) rectangle (2.24,1.903);
\fill[black!35, draw=black!60, line width=0.4pt] (1.72,1.903) rectangle (2.24,2.538);
\fill[blue!45, draw=blue!60!black, line width=0.4pt] (1.72,2.538) rectangle (2.24,2.900);
\node[tlab, anchor=north] at (1.98,-0.06) {q3b\\ $z{=}1.5$};
\fill[black!35, draw=black!60, line width=0.4pt] (2.58,0.000) rectangle (3.10,0.453);
\fill[blue!45, draw=blue!60!black, line width=0.4pt] (2.58,0.453) rectangle (3.10,0.906);
\fill[violet!45, draw=violet!60!black, line width=0.4pt] (2.58,0.906) rectangle (3.10,2.266);
\fill[orange!70, draw=orange!80!black, line width=0.4pt] (2.58,2.266) rectangle (3.10,2.900);
\node[tlab, anchor=north] at (2.84,-0.06) {q7b\\ $z{=}1.7$};
\fill[black!35, draw=black!60, line width=0.4pt] (3.44,0.000) rectangle (3.96,0.725);
\fill[blue!45, draw=blue!60!black, line width=0.4pt] (3.44,0.725) rectangle (3.96,1.450);
\fill[violet!45, draw=violet!60!black, line width=0.4pt] (3.44,1.450) rectangle (3.96,2.447);
\fill[orange!70, draw=orange!80!black, line width=0.4pt] (3.44,2.447) rectangle (3.96,2.900);
\node[tlab, anchor=north] at (3.70,-0.06) {L8b\\ $z{=}1.6$};
\fill[black!35, draw=black!60, line width=0.4pt] (4.30,0.000) rectangle (4.82,0.272);
\fill[blue!45, draw=blue!60!black, line width=0.4pt] (4.30,0.272) rectangle (4.82,1.087);
\fill[violet!45, draw=violet!60!black, line width=0.4pt] (4.30,1.087) rectangle (4.82,1.994);
\fill[orange!70, draw=orange!80!black, line width=0.4pt] (4.30,1.994) rectangle (4.82,2.900);
\node[tlab, anchor=north] at (4.56,-0.06) {q14b\\ $z{=}2.0$};
\fill[black!35, draw=black!60, line width=0.4pt] (5.16,0.000) rectangle (5.68,0.181);
\fill[blue!45, draw=blue!60!black, line width=0.4pt] (5.16,0.181) rectangle (5.68,0.544);
\fill[violet!45, draw=violet!60!black, line width=0.4pt] (5.16,0.544) rectangle (5.68,1.541);
\fill[orange!70, draw=orange!80!black, line width=0.4pt] (5.16,1.541) rectangle (5.68,2.900);
\node[tlab, anchor=north] at (5.42,-0.06) {q32b\\ $z{=}2.5$};
\fill[black!35, draw=black!60, line width=0.4pt] (6.02,0.000) rectangle (6.54,0.091);
\fill[blue!45, draw=blue!60!black, line width=0.4pt] (6.02,0.091) rectangle (6.54,0.362);
\fill[violet!45, draw=violet!60!black, line width=0.4pt] (6.02,0.362) rectangle (6.54,0.997);
\fill[orange!70, draw=orange!80!black, line width=0.4pt] (6.02,0.997) rectangle (6.54,2.900);
\node[tlab, anchor=north] at (6.28,-0.06) {L70b\\ $z{=}2.1$};
\fill[black!35, draw=black!60, line width=0.4pt] (6.88,0.000) rectangle (7.40,0.166);
\fill[blue!45, draw=blue!60!black, line width=0.4pt] (6.88,0.166) rectangle (7.40,0.829);
\fill[violet!45, draw=violet!60!black, line width=0.4pt] (6.88,0.829) rectangle (7.40,1.823);
\fill[orange!70, draw=orange!80!black, line width=0.4pt] (6.88,1.823) rectangle (7.40,2.900);
\node[tlab, anchor=north] at (7.14,-0.06) {q72b\\ $z{=}2.3$};
\end{scope}
\fill[black!15, draw=black!40] (0.00,3.28) rectangle (0.22,3.42);
\node[font=\tiny, anchor=west] at (0.27,3.35) {none above chance};
\fill[black!35, draw=black!60] (2.15,3.28) rectangle (2.37,3.42);
\node[font=\tiny, anchor=west] at (2.42,3.35) {name};
\fill[blue!45, draw=blue!60!black] (3.15,3.28) rectangle (3.37,3.42);
\node[font=\tiny, anchor=west] at (3.42,3.35) {definition};
\fill[violet!45, draw=violet!60!black] (4.45,3.28) rectangle (4.67,3.42);
\node[font=\tiny, anchor=west] at (4.72,3.35) {explanation};
\fill[orange!70, draw=orange!80!black] (6.00,3.28) rectangle (6.22,3.42);
\node[font=\tiny, anchor=west] at (6.27,3.35) {full prompt};
\node[lab] at (3.67,3.75) {Concept Recovery by Message Type (Reaches and Rising Concepts)};
\begin{scope}[cm={1.06,0,0,1,(-1.22,0)}]  %
\node[lab, align=center] at (10.89,3.60) {Criteria Where Examples Help\\ (real labeled items, $449$ criteria)};
\IfFileExists{figs/gen_examples_bank.tex}{\input{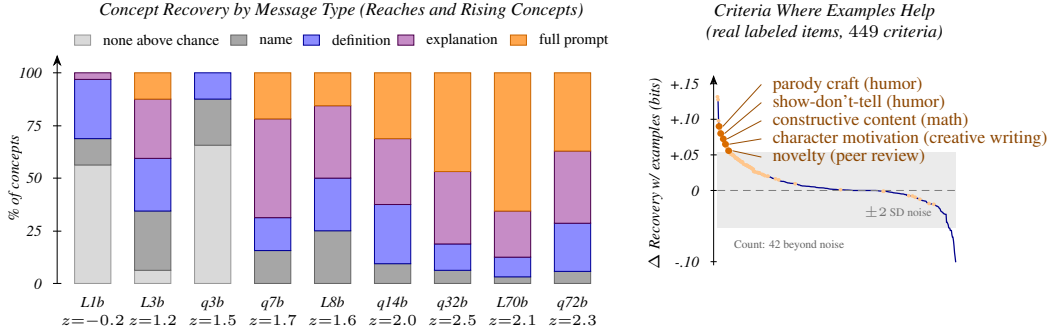}}{%
  \draw[black!35, densely dashed, rounded corners=3pt] (9.09,-0.1) rectangle (12.69,3.05);
  \node[tlab, red!60!black, align=center] at (10.89,1.45) {[pending: EXP-EXAMPLES-BANK-1\\ reconstruction-objective $\Delta I(m;m')$,\\ 236 metrics $\times$ 8 domains]};
}
\end{scope}
\end{tikzpicture}}%
\caption{\textbf{Which articulation strategy recovers each concept}. \emph{Left:} the message form that recovers \textit{Reaches} and \textit{Rising} concepts, by receiver. The grey share is concepts for which no form reaches $.55$ balanced accuracy against the held-out reference. Definitions and explanations carry the middle of the ladder, and the
full packet is needed only at the top. \emph{Right:} examples under the best method, four real items carrying the transmitter's own labels (Appendix~\ref{app:examples}): each of the $449$ criteria's gain in recovery over the definition-only message, $\Delta I(m;m')$ averaged over Qwen2.5-14B and Llama-70B, ranked. The grey band is two standard deviations of reconstruction noise (pooled $.054$ bits). $42$ criteria clear it at Qwen2.5-14B and $43$ at Llama-70B; four clear it at both (about four are expected by chance). Named points are examples from five tasks.}
\label{fig:message-forms}
\end{figure}

\paragraph{Finding \#3: Articulability is not based on pretraining, concept diffuseness or post-training dynamics.}
\label{sec:tacit-iso}

Finally, we attempt to explain whether learning mechanisms impact transmission of concepts. We first study pretraining dynamics. Exposure alone does not explain the ordering, measured as name-term frequency in training data: across all $1{,}270$ fitted metrics, reaching behavior, or acquisition onset, is uncorrelated with the corpus frequency of the concept's name terms: %
some criteria are acquired far earlier than frequency predicts; on the other hand, more common vocabulary criteria (e.g. ``active voice'', ``provide context'') are acquired late. We then tested whether acquisition tracks the \textit{form} of the concept: do compact concepts, defined in \cite{spangher} as concepts with few synonyms, internalize better while diffuse abstractions late? Overall, we find, it does not; compact and diffuse concepts are indistinguishable once exposure is accounted for (Appendix~\ref{app:exposure}). We observe that internalization is family-relative: a name that suffices in one model family still needs its definition in another, so in these cases is internalized is a property of the receiver, not of the concept. Finally, we observe that for metrics that are bounded (i.e. their scaling curve plateaus below the planted-rule ceiling) no acquisition exists in any model family. We examine post-training effects: running base and post-trained models through the same prompt scaffold, the post-trained model recovers $+.036$ more on reaches constructs and $+.034$ on rising ones, but only $+.008$ on bounded ones (Appendix~\ref{app:posttrain}). Taken together, neither exposure nor behavioral reinforcement seem to impact transmission of bounded concepts.

\section{Discussion}
\label{sec:discussion}

We started by defining the \textit{language limits} or \textit{articulation upper bounds} for a class of executors (e.g. LLMs) with the question: how well can an infinite-capacity transmitter convey a perfect message to an infinite-capacity receiver? We introduced two techniques to measure this: $\epsilon$-estimation to optimize messages and reconstruction-bottleneck to measure transmitter-receiver transmission; we used these tools to observe how transmission grows as LLMs scale. Most of the $1{,}270$ concepts we measured, we found, were either transmissible now ($321$ concepts), or could be with future scaled models ($884$ concepts). Only $65$ plateau, or were \textit{bounded}. Larger, more powerful versions of these same models, it seems, with better messages and richer inputs do not increase transmission for these concepts. What are these concepts, and why do they resist articulation even in the limits?

 Our first piece of evidence is gained from labeling them: these $65$ concepts primarily capture the identities and social groups of who wrote the text and who it is for (e.g. a persona held across a body of work, a community's in-jokes). This description aligns with \citet{collins2010tacit}'s view of  \textit{collective tacit knowledge}: concepts that name social norms might resist articulation because these norms capture personal or social existence; the fact we observe that transmission is \textit{bounded} indicates that language models do not possess the tacit knowledge required to jointly interpret these concepts (see Appendix \ref{app:human-agreement}). To more fully probe this claim, we ruled out certain null hypotheses: maybe these concepts \textit{were} articulable and not categorically different than saturating concepts, but did not transmit simply because they happened to be collinear with other concepts over some regions of the space, or were rarer labels. We test these (see Appendix \ref{app:plateau-checks}) by using \textit{more} transmission labels, $m$; we find that transmission plateaus at $|m|=30$ examples and bounded concepts remain bounded. Giving the receiver $90$ labeled items instead of $30$ does not raise recovery on bounded concepts; neither does extending scaling past its strongest member: GPT-5.6, far above every panel executor on the capability index, leaves all $65$ bounded concepts below the planted reference (Appendix \ref{app:more-labels} and \ref{app:frontier}). We also asked whether bounded concepts were simply the ones our models labeled inconsistently, or the ones where the largest models agreed with each other but not with the rest of the panel. Neither was true: bounded concepts are labeled as reliably as the others, and the largest models agree with each other \textit{least} on them (Appendix~\ref{app:plateau-checks}). Second, we hypothesized that maybe language models \textit{had} already internalized this information but simply could not express it in stated rules. With demonstrations and examples, maybe, a model would be able to \textit{show} but not \textit{tell} \citep{polanyi1966tacit}; however, our hypothesis was incorrect, we found that showing transmits no more than telling  (Finding~2)\footnote{One reason, we suspected, might be the design of our reconstruction bottleneck: in reconstruction, the receiver is asked to generate a rule to explain the labels $m$, so if an example in $p$ carries information that is not articulable in a rule, it will be lost at this step. To test this, we ran a test where the receiver generated its own prompt $p'$ using examples (Table~\ref{tab:examples-shown}). Across models, this also failed. Qwen2.5-14B recovered $.022$ bits with examples vs. $.036$ without on all concepts, $.011$ vs. $.034$ on the bounded ones. Llama-70B is similar, $.030$ vs. $.043$ bits over all concepts and $.047$ vs. $.076$ on the bounded ones.}. %

The above results indicate that such knowledge may not exist, already, in models. To gain more insight into how and when these concepts are learned we probed learning mechanisms by which knowledge \textit{does} enter models. First, we hypothesized that frequency and exposure in training datasets would have a strong impact on acquisition of \textit{all} concepts \citep{reber1967implicit,reber1989tacit,reber1993implicit} via the distributional hypothesis \citep{harris1954distributional}. However, nothing we measured in pretraining predicts which concepts a model internalizes (Finding~3). The frequency with which a concept's name appears in pretraining data does not predict when it is acquired, nor does the type of concept; whether the concept is diffuse, with many of synonyms or whether it is compact with few representations. Next, we tested whether \textit{behavioral} learning, or post-training, can impact transmission rates \cite{bf_skinner}. Post-training, we find, \textit{does} help transmission in \textit{rising} and \textit{reaches} concepts, but not \textit{bounded}. Base models and their post-trained siblings are bounded on the same constructs, and post-training adds recovery only where recovery already moves, so the plateau is not an artifact of post-training. Neither the distributional hypothesis nor behavioral mechanisms seem to be able to help models this class of bounded concepts.

Finally, just because a concept has asymptotically bounded transmission under \textit{one} class of executors (e.g. LLMs trained under roughly similar training recipes) does not mean it is inarticulable under \textit{all} executors. Humans, for example, have been shown to have high rates of agreement, or transmission, for certain concepts that appear as bounded in our experiments: e.g. \textit{``satire target''}, where trained annotators agree on whom a dark-humor meme targets at $\kappa=.72$ \citep{kasu2025dhumor}; \textit{``parody''}, which trained coders mark in television scenes at $\alpha=.82$ \citep{vanderwal2020humor}; and \textit{``personal self-disclosure''}, which annotators mark in forum posts at $\kappa=.88$ \citep{valizadeh2021selfdisclosure}. For each of the $65$ bounded concepts, we searched for prior literature\footnote{For each bounded concept, we ran four web searches with a language-model agent searching for relevant literature, finding $80$ relevant papers. We then performed a second pass to extract and check every statistic from the paper's text. When multiple papers covered the same concept, we kept the paper who's operationalized definition matched ours the closest, judged by an LLM. See Appendix~\ref{app:human-agreement} for more details.} and found human agreement studies for $37$ bounded concepts; of these, humans agree moderately-to-highly on $76\%$ and poorly on $24\%$\footnote{Of the $37$ concepts with prior literature performing measurement, $8$ ($22\%$) show high agreement (a single-rater coefficient of $.80$ or above), $20$ ($54\%$) moderate ($.40$ to $.79$) and $9$ ($24\%$) low (below $.40$); the other $28$ of the $65$ have no study we could find.}. On the other hand, some concepts are bounded both for LLMs as well as humans. Human annotators agree on ``originality of premise'' at $\kappa=.19$ \citep{amir2026jokespace}; visual promotability, i.e. clickbait strength, at $\kappa=.21$ \citep{potthast2018clickbait}; and the exercise quality of educational materials at a single-rater intraclass correlation of $.13$ to $.55$ \citep{vargo2003learning}. We repeated this experiment with \textit{rising} and \textit{reaches} concepts and also found an imperfect overlap: $679$ reaches and rising concepts had prior work measuring human agreement and humans agree poorly on $53\%$ of them, e.g. laugh yield ($\alpha=.12$; \citealp{meaney2021hahackathon}), balance and freedom from bias ($\kappa=.21$; \citealp{spinde2021mbic}) and argumentation quality in public comments ($\kappa=.14-.20$; \citealp{falk2022discourse}), all concepts that LLMs transmit at or toward the planted-rule level (Appendix~\ref{app:human-agreement}). Thus, our \textit{articulability upper bound} findings \textit{only} have implications within the classes of models that we study; articulation limits are executor-family relative. 

In summary, we see more work to be done to characterize this class of bounded concepts. Many bounded concepts tie to identity and social group are often also ones we wish our models to understand in order to perform frontier tasks \citep{kripke_rulefollowing, bai2022constitutional,sap2022annotators,santy2023nlpositionality,sorensen2024pluralistic}. Our finding that they may \textit{not} be articulable by current model families, prompt strategies or learning strategies has important implications for language modeling. Looking forward, to keep making aligned progress on frontier tasks, we see the need for a deeper understanding of how such knowledge may be acquired and the learning mechanisms to help us continue to make progress.

\section{Conclusion}
\label{sec:conclusion}

We studied \textit{articulation upper bounds} for classes of executors, or how much of a concept can be transmitted from one language model to another with optimal messages. We built two tools to answer this: $\epsilon$-estimation finds the best message a transmitter can send and estimates how far that message is from the best possible one; the reconstruction bottleneck measures how much of a message a receiver recovers, without human labels. Finally, we use observational scaling laws to extrapolate both along model capability. Across $1{,}270$ concepts from eight domains, most concepts can be articulated between language models now or will be with scale. A small set tied to identity and social group stays bounded in every model family we tested; examples and larger models largely do not change this performance. We release the tools and every analysis (Appendix~\ref{app:reproducibility}) so that rubrics and constitutions can be measured, per model, before they are relied on. Our work represents a potentially novel way to benchmark frontier performance and a call for novel learning techniques.

\paragraph{Limitations.} Our work needs to be taken with the following limitations. The reconstruction measure, by design, was unsupervised and, although it correlated strongly on tasks where human labels exist, it used no human labels. Like other unsupervised measures (e.g. consistency \citep{wang2023selfconsistency}), it has it's own set of flaws: reconstruction can reward any rule the receiver can restate consistently with the sender, including a wrong one (Appendix~\ref{app:recon-alternatives}). Additionally, while our definition of \textit{rising}, \textit{reaches} and \textit{bounded} is consistent on a population level, which bands individual concepts falls in is less certain: nder resampling, $163$ of the $1{,}270$ concepts have less than a $.6$ chance of landing in their printed regime, though the sizes and makeup of the regimes hold (Appendix~\ref{app:regime-robustness}). In our testing of learning mechanisms, our measurement for the frequency of a concept's name terms, is coarse and done through Dolma, which is not necesssarily the same training dataset as major open-source models. Finally, we specifically studied eight creative and qualitative domains; we do not know whether factual or procedural concepts behave the same way.

\newpage
\bibliographystyle{iclr2026_conference}
\bibliography{refs}

\clearpage
\appendix
\etocdepthtag.toc{mtappendix}
\section*{Appendix Contents}
{\small
\etocstandardlines
\etocsettocstyle{}{}
\etocsettagdepth{mtbody}{none}
\etocsettagdepth{mtappendix}{subsection}
\tableofcontents
}

\vspace{0.5em}
\noindent Appendices~\ref{app:unit-gate} to~\ref{app:z-validation} document the two instruments and the capability index, Appendices~\ref{app:osl-regimes} to~\ref{app:exposure} the scaling analyses, Appendix~\ref{app:plateau-checks} the simpler explanations for the bounded regime and the tests that bear on them, and Appendices~\ref{app:prereg} and~\ref{app:reproducibility} the protocols and the data.

\section{Unit sampling and the causal screen in detail}
\label{app:unit-gate}

\begin{figure}[h]
	\centering
	\resizebox{\textwidth}{!}{%
\begin{tikzpicture}[lab/.style={font=\scriptsize\itshape, align=left},
  tlab/.style={font=\tiny\itshape, align=center}]
\IfFileExists{figs/gen_unit_decomp.tex}{\input{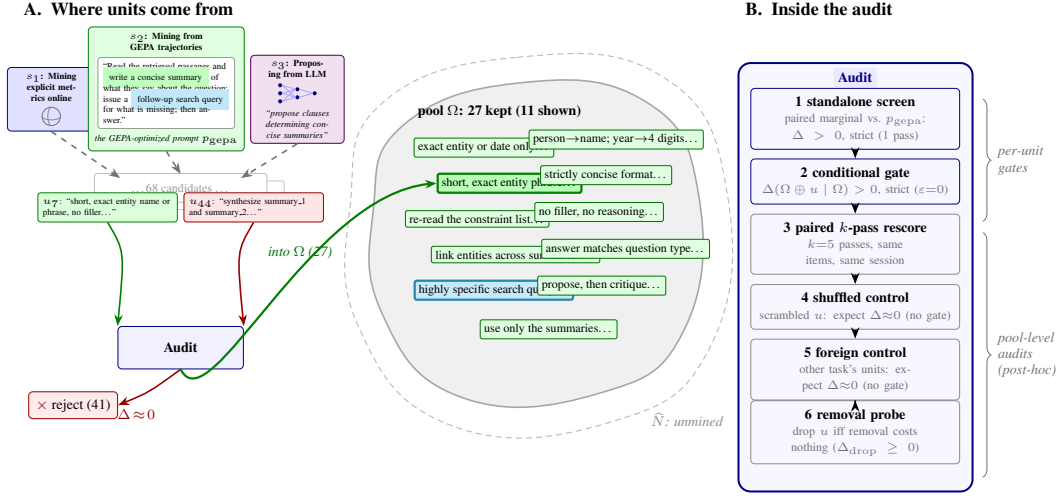}}{}
\end{tikzpicture}}%
	\caption{\textbf{Unit-mining process} (Left-Hand Side of Figure \ref{fig:mining-bound}) \emph{(A)} N channels supply unit candidates (shown: $s_1$: units found online; $s_2$: units proposed directly by an LLM; $s_3$: units mined from the GEPA-induced prompt $p_{\text{GEPA}}$). Candidates ($u_{7}$, green; and $u_{44}$, red) all enter the pool $\Omega$; what follows orders and selects them. \emph{(B)} As implemented: \#1 one single-pass marginal against the starting prompt ($p_{\text{GEPA}}$'s shipped candidate) on the $100$-item selection panel, with $\Delta>0$ making a unit eligible; \#2 units are re-tested in combination only during selection (prefix sweep, delete pass), never as a pool filter; \#3 multi-pass scoring is applied afterwards to shipped prompts and prefix curves, never to units; \#4--\#5 the shuffled- and foreign-unit controls are separate experiments on the frozen pool (Appendix~\ref{app:unit-gate}) and remove nothing; \#6 the delete pass runs during selection on the chosen set, on the same panel.}
	\label{fig:unit-decomp}
\end{figure}

Figure~\ref{fig:unit-decomp} expands the left-hand side of Figure~\ref{fig:mining-bound}: the sampling channels that supply unit candidates, and the screen that every candidate passes through before entering the pool $\Omega$.

\paragraph{Three samplers supply unit candidates: web retrieval, direct proposal and optimizer trajectories.} Three declared samplers feed the pool. (1) \emph{Web retrieval}: prompt-advice strings mined from task-adjacent web text, canonicalized into short imperative fragments. (2) \emph{Direct LLM proposal}: the proposer model is asked for candidate instructions for the task, across several framings, and its answers are split into minimal fragments. (3) \emph{Optimizer-trajectory extraction}: every textual change a reflective optimizer (GEPA) made to its prompt is diffed out and atomized. In the pools of record, LLM proposal contributes most of the units ($60/68$ on HotpotQA, $96/164$ on HoVer, $60/128$ on IFBench) and optimizer trajectories the rest ($8$, $68$ and $68$; on HoVer $20$ of the $68$ come from other task LMs' trajectories); web retrieval contributes no unit to these pools. Each channel is a declared sampler in the sense of Section~\ref{sec:e_certifiability}: every ceiling in the paper is relative to this generating process.

\paragraph{The screen is one single-pass marginal per fragment, and pairs that help only together are not lost.} We corrected this description against the shipped code. Every mined candidate enters $\Omega$; candidates are split once at sentence and bullet boundaries into fragments of $25$--$400$ characters. The only per-unit measurement is one single-pass marginal against the starting prompt (GEPA's shipped candidate) on the fixed $100$-item selection panel; units with a positive marginal can be selected, the rest stay in the pool unused. No unit is re-measured before selection, and no correction is made for the $68$--$164$ comparisons a pool involves; the prefix sweep, the delete pass and the held-out confirmation are the only protection against noise. The shuffled-, foreign- and keyword-unit controls are separate experiments on random draws from the frozen pool (HotpotQA at 0.6B and 8B, HoVer at 8B); they show that assembled prompts gain from unit content rather than length or vocabulary, and they remove nothing from $\Omega$. A unit's identity is its canonical string; whether it has a stable effect is tested only during selection, not at admission. We tested whether this single-unit gate discards units that help only in pairs, on the HoVer pool of record (protocol HB\,227, artifact \texttt{outputs/analyses/pair\_probe\_v1/}): $120$ random pairs of rejected units, $40$ pairs of a rejected unit with a top-ten accepted unit, and $20$ pairs of word-shuffled units, each appended to the starting prompt and scored on the selection panel, with a three-pass re-score for any pair above $+.03$. No rejected pair cleared the shuffled-unit band after re-scoring ($0$ of $120$, against $0$ of $20$ null pairs; $15\%$ against $20\%$ stayed positive), and adding a rejected unit to a strong one cost $.021$ on average. On this pool the gate loses little at the pair level; we have not tested triples or other pools.

\section{The subset-selection algorithm behind $M_\Omega$: approximation guarantees, their
assumptions and the measured submodularity check}
\label{app:submodular}

This appendix defines the ``submodular exploration (over units)'' /
``add\,/\,drop\,/\,swap units'' step of Figure~\ref{fig:mining-bound}, states which approximation
guarantees we can invoke for it, and, because none of those guarantees holds unconditionally, states which
assumptions each one rests on and what we do when an assumption fails.

\paragraph{Notation: the unit pool, the recovery of a subset and the selection problem.} Let $\Omega=\{u_1,\dots,u_K\}$ be the mined unit pool and, for $S\subseteq
\Omega$, let $R(S)$ denote the measured recovery of the canonical prompt realizing exactly the
units in $S$. The selection problem is $\max_{|S|\le k} R(S)$, and $M_\Omega$ is the prompt
realizing the selected subset. Two regimes are dispatched on $|\Omega|$.

\paragraph{The complete procedure, from sampling to the ceiling estimate, in one algorithm.} Algorithm~\ref{alg:ecert} assembles the full pipeline in
one place: the unit-sampling and screening stage of Appendix~\ref{app:unit-gate}, the selection stage this
appendix analyzes, and the ceiling-estimation stage of Section~\ref{sec:missing-value-est} and
Appendix~\ref{app:mass}. Selection is written as the two-regime dispatch introduced above; the
guarantees and failure modes attached to each line are the subject of the rest of this appendix.

\begin{algorithm}[h]
\caption{$\epsilon$-estimation: mine, select, estimate}
\label{alg:ecert}
\begin{algorithmic}[1]
\Require measurement $f$, seed prompt $p_0$, declared samplers $\{s_1, s_2, \ldots\}$, subset cap $k$, confidence level $1-\delta$
\Statex \emph{Stage 1: mine the unit pool}
\State draw candidate strings from every sampler; canonicalize each into minimal fragments
\ForAll{candidates $u$}
\State split $u$ once at sentence/bullet boundaries; add every fragment to $\Omega$; measure its single-pass marginal $\Delta_u$ against the initialized prompt
\State mark $u$ eligible iff $\Delta_u > 0$
\EndFor
\State (post hoc, outside the pipeline: shuffled-, foreign- and keyword-unit controls on the frozen pool; they filter nothing)
\Statex \emph{Stage 2: select the prompt} (benchmark runs take the second branch; the first is used for reconstruction-side criterion sets)
\If{$|\Omega| \le 15$}
\State score every subset of $\Omega$; set $S_g$ to the exact optimum
\Else
\State render each candidate subset with the fixed template, score once
\State prefix sweep: order eligible units by $\Delta_u$, append them one at a time up to the prefix cap ($48$), retain the best-scoring prefix
\State prune and recover (one pass each): delete any kept unit whose removal costs nothing; re-add any skipped unit that now helps
\State (double greedy is not run in the benchmark experiments; it is analysed in this appendix for the small-set case)
\EndIf
\State confirm on held-out data that $p_{S_g}$ beats the initialized prompt (GEPA's shipped candidate); otherwise ship that prompt unchanged
\Statex \emph{Stage 3: estimate the ceiling}
\State coverage: compute Good--Turing novelty $f_1/n$ and the bias-corrected Chao1 point estimate $\widehat N$ from the overlap of independent sampling runs
\State value: fit $\widehat V(\cdot)$ to the greedy prefix trajectory; set $\widehat\epsilon = \widehat V(\widehat N) - V(S)$ as a point estimate; report its backtest error and one-sided coverage (Appendix~\ref{app:backtest})
\State structure: estimate $\hat\gamma$ over evaluated subset pairs
\If{non-monotonicity is detected ($\hat\gamma$ clamps to $0$)}
\State withhold $U_2$; report the selected prompt as measured, not certified
\Else
\State report $U_2 = R(S_g) + \frac{1}{\gamma} \sum_{j=1}^{k} \delta_{(j)}$ as a curve in $\gamma$ (Eq.~\ref{eq:u2})
\EndIf
\State \Return prompt $p_{S_g}$, ceiling $\widehat B = f(p_{S_g}) + \widehat\epsilon$, estimate $(\widehat\epsilon, \delta)$ labeled \textsc{certified} only in the exact branch, \textsc{measured} otherwise, with $U_2$ when not withheld
\end{algorithmic}
\end{algorithm}

\paragraph{Pools of at most $15$ units are solved exactly by enumeration.} All $2^{|\Omega|}-1$ subsets are scored and the
optimum is found by enumeration. The resulting certificate is exact for the evaluated empirical objective on the fixed probe set (with noisy scores it is not a statement about the population optimum): no
monotonicity and no submodularity of $R$ is used anywhere. This branch applies to the reconstruction-side criterion sets; all six supervised benchmarks (pools of $68$--$164$ units) run in the large regime below. It also
serves as the validation harness for the large-$\Omega$ methods below (on instances small
enough to enumerate, the approximate selectors and the upper bound of
Eq.~\eqref{eq:u2} are cross-checked against the true optimum).

\paragraph{Larger pools (more than $15$ units) run the greedy selection algorithms.} No enumeration is attempted;
subsets are scored lazily (only subsets that an algorithm visits are evaluated). Two
selectors are defined; the benchmark experiments run only the first (the production paragraph below), and the second was run only in the exact-regime analysis:
\begin{enumerate}
\item \emph{Forward greedy}: starting from $\emptyset$, repeatedly add the unit with the
largest marginal gain $\delta(e\mid S)=R(S\cup\{e\})-R(S)$ until the budget $k$ is reached. For monotone
submodular objectives under a cardinality constraint greedy has the classical $(1-1/e)$
guarantee \citep{nemhauser1978analysis}; as discussed below, our $R$ is not certified to satisfy
either hypothesis, so here greedy is a strong heuristic whose quality is bounded
\emph{per-instance} by Eq.~\eqref{eq:u2} rather than by the worst-case factor.
\item \emph{Double greedy} \citep{buchbinder2015tight}: a single pass over the elements
maintaining an included set $A$ (initially $\emptyset$) and a surviving set $B$ (initially
$\Omega$); for each element $i$ it computes the two marginals $d_A=R(A\cup\{i\})-R(A)$ (gain of
adding) and $d_B=R(B\setminus\{i\})-R(B)$ (gain of deleting), keeping $i$ with probability
$d_A^+/(d_A^++d_B^+)$ (and with probability one when both clamp to zero); on a keep, $B$ is unchanged; on a discard, $i$ leaves $B$. An earlier draft and the code used the deletion \textit{cost} with the same clamp, reversing the second marginal; both were corrected on 2026-09-11. No reported number depends on this error, because double greedy is not used in the benchmark runs. For \emph{unconstrained non-monotone} submodular maximization the
randomized version is a tight $\tfrac12$-approximation and the deterministic keep-iff-$d_A\ge
d_B$ variant is a $\tfrac13$-approximation \citep{buchbinder2015tight}; the add/delete local
search of \citet{feige2011maximizing} gives the same $\tfrac13$ family, and under a binding
cardinality constraint the applicable non-monotone guarantee weakens to $1/e$ (random greedy,
\citealp{buchbinder2014submodular}). Double greedy tolerates non-monotonicity, but its factor requires submodularity of $R$, which is not established for our objective; here it is a heuristic with a per-instance bound.
\end{enumerate}
The production $M_\Omega$ procedure used in the benchmark experiments runs one selector: single-pass marginals against GEPA's shipped candidate (not the seed); a prefix sweep that appends units in order of marginal gain, up to $48$ units, and keeps the best-scoring prefix (LiveBench ran the earlier cap of $32$ and shipped GEPA's prompt); one delete pass; one add-back pass; and a single-pass confirmation on a separate panel. Neither enumeration nor double greedy runs there; every selection-time score is one pass ($T=1$), and multi-pass scoring is applied afterwards to shipped prompts only. Table~\ref{tab:procedure-per-result} gives the selector, pool size and pass count behind every reported result.

\paragraph{The leftover marginals after greedy bound how much any feasible subset could still gain.} After greedy terminates with $S_g$, the leftover
marginals convert into an upper bound on how much better \emph{any} feasible subset could be.
With $\delta(e)=\delta(e\mid S_g)$ for $e\notin S_g$, $\delta_{(1)}\ge\delta_{(2)}\ge\cdots$ the
sorted gains, and $\gamma\in(0,1]$ the submodularity ratio of $R$ \citep{das2011submodular},
\begin{equation}
\label{eq:u2}
\mathrm{OPT}_{\le k} \;\le\; U_2 \;:=\; R(S_g)\;+\;\frac{1}{\gamma}\sum_{j=1}^{k}\delta_{(j)}.
\end{equation}
At $\gamma=1$ this is Minoux's classical online bound from the accelerated-greedy literature
\citep{minoux1978accelerated}; the $1/\gamma$ form follows in three lines from the definition of
the ratio (monotonicity gives $R(S^*)\le R(S^*\cup S_g)$; the ratio definition at
$S=S_g,\ \Omega'=S^*\setminus S_g$ gives $R(S_g\cup S^*)-R(S_g)\le
\frac{1}{\gamma}\sum_{e\in S^*\setminus S_g}\delta(e)$; and $|S^*\setminus S_g|\le k$ bounds the
sum by the top-$k$ gains). The multiplicative worst-case counterpart
$R(S_g)\ge\frac{1}{\alpha}(1-e^{-\alpha\gamma})\,\mathrm{OPT}$, with curvature $\alpha$
\citep{bian2017guarantees}, is reported only to characterize the objective in advance;
run-level claims always use the instance bound~\eqref{eq:u2}. See \citet{krause2014submodular}
for the surrounding theory.

\paragraph{Measured departure from submodularity.} We measured how far the pools depart from submodularity from the selection logs of the rows of record, with no new scoring (artifact \texttt{outputs/analyses/submodularity\_diagnostics\_v1/}). Submodularity is diminishing returns: a unit's marginal gain can only fall as other units are added, so the joint gain of a set never exceeds the sum of its members' individual gains. The opposite, synergy, is what the $1/\gamma$ factor in Eq.~\eqref{eq:u2} guards against. On the selection batch of each benchmark we compared the sum of the single-unit marginals against the seed with the joint gain of every prefix the sweep evaluated. On every benchmark with an assembled gain the sum exceeds the joint gain, by a factor of $2$ to $9$ on HotpotQA (median $6.3$ over $39$ prefixes), $2$ to $25$ on IFBench (median $11.6$, $47$ prefixes), $3$ to $61$ on AIME (median $20$, $31$ prefixes) and $3$ to $26$ on PUPA; no evaluated prefix on any benchmark has a joint gain above the sum of its parts. Along these trajectories the ratio $\hat\gamma$ is therefore $1$, and $U_2$ reduces to Minoux's bound. The property that does fail is monotonicity: $3\%$ of the units in the HotpotQA prompt and $4\%$ in the IFBench prompt raise the score when removed, and of the $25$ units the IFBench sweep skipped, none raises the score when added to the assembled prompt ($1$ of $12$ does on HotpotQA). The within-pool gap is then small where it can be measured from the add-back marginals: at the best prefix on the selection batch, $U_2$ is $.710$ against an achieved $.700$ on HotpotQA at $\gamma=1$ ($.740$ at $\gamma=.25$), and $.892$ against $.892$ on IFBench at every $\gamma$, since no leftover unit is positive. PUPA's leftovers give $.993$ against $.951$ on a metric capped at $1$. AIME, LiveBench and HoVer have no add-back phase in their logs, so only a seed-marginal proxy exists for them, which is not a bound. Two cautions: every selection score is one pass ($T=1$), so the ratios carry scoring noise; and the additivity test uses marginals against the seed rather than conditional marginals, so it is a necessary condition for submodularity on the evaluated sets, not a proof over the whole lattice. On the reconstruction side the picture is the reverse: on nine rubric decompositions scored by ideal recovery $I(M;X_S)$ ($K=4$--$14$ criteria each), the spectral submodularity ratio is $.13$--$.41$, so rubric criteria are complementary and a greedy bound there would be loose by $1/\gamma = 2.5$--$8$. The two readings use different objectives and different units and are not pooled.

\paragraph{Assumptions of the selection guarantees.} Equation~\eqref{eq:u2} and the
$\tfrac12$/$\tfrac13$/$1/e$ factors are conditional results, and four of their hypotheses are
at risk in our setting. We enumerate them here:
\begin{enumerate}
\item \emph{Submodularity of $R$ is not assumed, and $\gamma$ cannot be safely plugged in.}
$R$ is an empirical black box; $\gamma$ is estimated by sampling violation pairs
$(S,\Omega')$ on the evaluated lattice. A sampled minimum over pairs can only \emph{overstate}
the population ratio ($\hat\gamma\ge\gamma_{\mathrm{pop}}$), so $1/\hat\gamma$ makes
Eq.~\eqref{eq:u2} too small, and for an upper bound this error is in the invalid direction. We
therefore treat the plug-in $U_2$ at $\hat\gamma$ as a diagnostic labeled an \emph{estimate}, not a certificate; reporting $U_2$ as a curve in $\gamma$ is the intended presentation. $\hat\gamma$ is the minimum over $400$ random pairs $(S,\Omega')$ with $|\Omega'|\le 6$, clamped to $[0,1]$; the $\hat\gamma=1$ of the measured departure above comes from the additivity test on the selection logs, not from this sampler.
\item \emph{Monotonicity fails, and the failure is detectable.} The derivation of Eq.~\eqref{eq:u2} uses
$R(S^*)\le R(S^*\cup S_g)$. Raw $R$ is non-monotone (the drop-one pass finds units whose removal helps), so the bound
is valid only for the free-disposal monotonization $R^{\uparrow}(S)=\max_{T\subseteq S}R(T)$,
an explicit upper idealization that cannot be constructed exactly at large $|\Omega|$. The $U_2$ values above therefore bound $R^{\uparrow}$, not $R$. When the sampled ratio clamps to $0$, $U_2$ is withheld and the selected prompt is reported as measured, not certified.
\item \emph{The value oracle is noisy.} All guarantees above assume exact access to $R(S)$; we
estimate it from finitely many items and scoring passes, so greedy ranks noisy marginals and
accepted decisions can be sampling artifacts. Submodular maximization under noise is strictly
harder, and the exact-oracle ratios do not transfer \citep{hassidim2017submodular}. We reduce the effect of noise
with paired evaluation of marginals, confirmation re-evaluations of accepted units, and
multi-pass test scoring, but we do not claim a noise-adjusted approximation ratio.
\item \emph{Guarantee scope.} The $\tfrac12$ and $\tfrac13$ factors of double greedy (above) hold for the unconstrained problem. In our runs the cap is set
above the number of positive-marginal units, so the unconstrained reading applies, but this is
a property of the instances rather than of the method.
\end{enumerate}

Where every hypothesis fails, the
output is the measured optimum over evaluated subsets plus the missing-value ceiling of
Section~\ref{sec:missing-value-est}, with no claimed approximation ratio; the ceiling-estimation
procedure is designed to fall back to that statement rather than silently keep its guarantees.

\begin{table}[h]
\small
\begin{center}
\resizebox{\textwidth}{!}{%
\begin{tabular}{p{3.3cm} p{1.9cm} p{4.6cm} c c p{3.0cm}}
\toprule
result & pool size(s) & selector & enumeration? & double greedy? & passes at selection \\
\midrule
Table~\ref{tab:all-baselines}, Table~\ref{tab:main-results}, Table~\ref{tab:arm-accounting} (HotpotQA, HoVer, IFBench, PUPA, AIME) & $68$, $164$, $128$, $120$, $48$ & single-pass paired marginals against GEPA's shipped candidate, a cumulative prefix sweep (cap $48$), one drop-one pass, one add-back pass, single-pass confirmation on a disjoint panel & No & No & $T=1$; the shipped candidate is then scored with five test passes \\
Table~\ref{tab:all-baselines}, Table~\ref{tab:main-results}, Table~\ref{tab:arm-accounting} (LiveBench) & $48$ & same selector, prefix cap $32$ (the earlier default, not $48$); the confirmation guard rejected every candidate unit & No & No & $T=1$; no unit compiled, so the shipped prompt is GEPA's $600$-call candidate rescored \\
HotpotQA no-GEPA-unit ablation (\S\ref{app:budgets}) & $60$ & same selector, re-run on the pool with the $8$ trajectory-mined (GEPA-derived) units removed & No & No & $T=1$ at selection; the reported comparison used one scoring session at five passes \\
HoVer selection replicates (Table~\ref{tab:arm-accounting}, note $b$; \S\ref{app:budgets}) & $164$ (same frozen pool, run twice) & same selector & No & No & $T=1$; five-pass test $.539$ and $.541$ against the row of record's $.572$ (HB\,226) \\
Screened robustness run, HoVer only (HB\,225) & $164$ (same frozen pool) & lazy greedy on conditional marginals, restricted to units that clear a shuffled-unit null band after a $3$-pass rescore & No & No & $T=3$ screening rescore, then $T=1$ fresh evaluations (at most $8$ per step); five-pass test $.520$, $-.052$ $[-.077,-.027]$ against the row of record (HB\,226) \\
Value-curve ceilings (Figure~\ref{fig:missing-value} left, \S\ref{app:eps-assumptions}, \S\ref{app:backtest}) & same pools as above & a curve fit to the prefix-sweep trajectory selection already produced (HotpotQA: the same session's sweep extended to $k=68$ on the test split; the other five: the selection-panel sweep) & No & No & $T=1$ \\
Unsupervised bank results (Figure~\ref{fig:missing-value} right, \S\ref{sec:scaling}) & n/a & n/a; this is the reconstruction channel, not a selection result & n/a & n/a & n/a \\
\bottomrule
\end{tabular}}
\end{center}
\caption{Selection procedure behind each reported result. The paper describes several selection procedures (ranking marginals once against the seed, recomputing the best marginal at every step, pruning and re-adding, double greedy, exhaustive enumeration); this table states which one produced which result. Every reported number in the paper traces to exactly one row of this table.}
\label{tab:procedure-per-result}
\end{table}
\section{$\epsilon$-estimation results in full}
\label{app:ecert-results}

\subsection{Supervised benchmarks: comparison against prompt optimizers, call budgets and run-to-run spread}
\label{app:ecert-supervised}

Table~\ref{tab:main-results} is the comparison referenced in Section~\ref{sec:results}. It compares
$\epsilon$-estimated prompts against GEPA, GEPA+Merge and MIPROv2 on six benchmarks (Qwen3-8B,
official metrics and splits); every candidate is scored in one uniform session per benchmark with
five independent test passes. Absolute levels are session-specific: the same shipped prompts scored ${\sim}.05$--$.06$ lower on AIME in September than in July, and PUPA's judge shifted by ${\sim}.2$ between two August sessions. Only paired within-session differences are comparable; no cell combines scores from different sessions, and the fitted asymptotes of Appendix~\ref{app:backtest} (validation-side prefix means) are never compared with these test scores.

Table~\ref{tab:all-baselines} reports the same paired contrast against every baseline ($20{,}000$ paired item bootstraps; artifact \texttt{runs\_paperexact/all\_baseline\_deltas\_v1.json}). $M_\omega$ beats all three baselines on HotpotQA, HoVer and AIME; on LiveBench and IFBench it beats MIPROv2 only; on PUPA it beats GEPA and GEPA+Merge, and its $+.025$ over MIPROv2 is not significant. On HotpotQA the three GEPA seeds at $16{,}700$ calls scored $.567$, $.413$ and $.402$, so any single-run comparison is subject to that spread; against the seed mean the difference is $+.183$ $[+.147,+.219]$.

\begin{table}[h]
\begin{center}
{\small
\begin{tabular}{lccc}
\toprule
benchmark & $M_\omega-$GEPA & $M_\omega-$GEPA+Merge & $M_\omega-$MIPROv2 \\
\midrule
HotpotQA & $+.077$ $[+.045,+.109]$ & $+.222$ $[+.180,+.266]$ & $+.208$ $[+.165,+.251]$ \\
HoVer & $+.071$ $[+.043,+.099]$ & $+.052$ $[+.025,+.079]$ & $+.081$ $[+.052,+.110]$ \\
LiveBench & $+.016$ $[-.010,+.043]$ & $+.025$ $[-.002,+.054]$ & $+.049$ $[+.021,+.080]$ \\
IFBench & $+.025$ $[-.003,+.053]$ & $+.023$ $[-.004,+.049]$ & $+.060$ $[+.029,+.092]$ \\
AIME & $+.035$ $[+.011,+.061]$ & $+.027$ $[+.001,+.053]$ & $+.083$ $[+.052,+.115]$ \\
PUPA & $+.047$ $[+.018,+.076]$ & $+.069$ $[+.041,+.098]$ & $+.025$ $[-.006,+.057]$ \\
\bottomrule
\end{tabular}}
\end{center}
\caption{\textbf{Paired contrasts against every baseline}: mean item-level difference with $95\%$ bootstrap interval, same scoring sessions as Table~\ref{tab:main-results}. The HotpotQA row is the equal-call session (HB\,220; GEPA at $16{,}700$ calls).}
\label{tab:all-baselines}
\end{table}

\begin{table}[h]
\begin{center}
\begin{tabular}{lccccc}
\toprule
benchmark & GEPA & GEPA+Merge & MIPROv2 & $\epsilon$-est.\ ($M_\omega$) & $\Delta$ vs GEPA \\
\midrule
HotpotQA & $.567$ & $.421$ & $.435$ & $\mathbf{.643}$ & $+.077^{***}$ \\
HoVer & $.497$ & $.515$ & $.487$ & $\mathbf{.567}$ & $+.071^{***}$ \\
LiveBench & $.695$ & $.686$ & $.661$ & $.710$ & $+.016$ \\
IFBench & $.417$ & $.419$ & $.382$ & $.442$ & $+.025$ \\
AIME & $.323$ & $.331$ & $.275$ & $\mathbf{.357}$ & $+.035^{**}$ \\
PUPA & $.638$ & $.616$ & $.659$ & $\mathbf{.685}$ & $+.047^{**}$ \\
\bottomrule
\end{tabular}
\end{center}
\caption{\textbf{$\epsilon$-estimation: prompt optimization comparisons}. For every benchmark we score each candidate with five independent test passes (scores are the per-item means), using Qwen3-8B as the proposer and language model, to match \citet{agrawal2025gepa}. All benchmarks are run with the same number of steps as in \citet{agrawal2025gepa}; on every benchmark the GEPA column is additionally the \emph{equal-call} run, in which GEPA is re-run at the call budget that $\epsilon$-estimation spent (Appendix~\ref{app:budgets}). The comparison is therefore call-matched per benchmark, not budget-capped. On LiveBench the confirmation guard rejected every unit and shipped GEPA's $600$-call prompt unchanged, so that row compares GEPA's own prompts at two budgets and contains no unit-selection effect. $\Delta$ is the mean paired item-level difference, with a $20{,}000$-draw bootstrap interval over items (each item's score is its five-pass mean; the passes do not enlarge the sample); stars mark two-sided bootstrap $p<.01$ ($^{**}$) and $<.001$ ($^{***}$). Five passes measure scoring noise, not the variance between optimizer runs, which Appendix~\ref{app:budgets} reports. Contrasts against GEPA+Merge and MIPROv2 are in Table~\ref{tab:all-baselines}. Items per benchmark: HotpotQA 300, HoVer 300, IFBench 294, PUPA 221, AIME 150, LiveBench 126.}
\label{tab:main-results}
\end{table}

\paragraph{Optimization budgets and budget comparability.}\label{app:budgets}
All Table~\ref{tab:main-results} methods run at declared task-call budgets.
The GEPA column is GEPA re-run at the number of calls $M_\omega$ spent on that benchmark (HotpotQA $16{,}700$; HoVer $10{,}110$; LiveBench $18{,}708$; IFBench $23{,}300$; AIME $6{,}870$; PUPA ${\sim}29{,}526$, of which $98.8\%$ completed before a disk failure). Where $M_\omega$'s own spend is logged (Table~\ref{tab:arm-accounting}), it equals that budget on HotpotQA and IFBench and falls below it on LiveBench ($4{,}616$ selection-and-validation calls against $18{,}708$) and AIME ($2{,}490$ against $6{,}870$), so on those two benchmarks GEPA had the larger budget. GEPA+Merge and MIPROv2 (heavy mode) run at their own defaults. $M_\omega$'s spend is its realized selection cost, not the $2{,}400$-call default quoted in an earlier draft. GEPA gains most of its budget benefit early: on HotpotQA, $600$, $2{,}400$ and $16{,}700$ calls gave $.40$, $.53$ and $.57$, and at full parity $M_\omega$ still leads by $+.077$. At equal calls GEPA generates more candidate text than $M_\omega$ (${\sim}70$M against ${\sim}46$M characters on HotpotQA), so call parity does not give our method a token advantage. $M_\omega$ starts from GEPA's shipped $600$-call prompt (on HotpotQA that prompt is the seed) and mines GEPA's own runs for units. On HotpotQA the ablation without those units scores $-.013$ $[-.035,+.008]$ against the full pool and $+.062$ $[+.034,+.092]$ over GEPA at equal calls (HB\,226a), so the gain there does not depend on the GEPA-derived units. On HoVer, in one session at five passes, the ablation without GEPA-derived units scores $-.037$ $[-.058,-.016]$ against the full pool and $+.031$ $[+.004,+.059]$ against GEPA at the matched budget. Two fresh selection runs on the same pool score $.539$ and $.541$ against the row of record's $.572$ (its HB\,226-session value; the uniform scoring session of Table~\ref{tab:main-results} gives $.567$), differences of $-.033$ and $-.031$ ($-.002$ between them), and both beat GEPA ($+.035$ $[+.009,+.062]$, $+.037$ $[+.011,+.064]$), so the HoVer gain is $+.035$ to $+.068$ across three selection runs. GEPA itself, restarted from that same shipped prompt with $M_\omega$'s remaining budget ($16{,}100$ and $9{,}510$ further calls), scores $.569$ on HotpotQA and $.491$ on HoVer against $.569$ and $.507$ from seed ($+.001$ and $-.015$, neither significant). $M_\omega$ leads it by $+.071$ $[+.036,+.107]$ and $+.077$ $[+.050,+.105]$ (one session each, five passes). Table~\ref{tab:arm-accounting} gives the full per-method accounting behind this section: initial prompt and inherited artifacts, independent optimization runs, calls by stage, realized totals and characters where logged, and the certified test score against GEPA.

\begin{table}[t]
\begin{center}
{\footnotesize
\resizebox{\textwidth}{!}{%
\begin{tabular}{l p{3.3cm} l r r l}
\toprule
method & init / inherited & calls: prop/sel/val/test & total & chars & test score ($\Delta$ vs GEPA) \\
\midrule
\multicolumn{6}{l}{\textbf{HotpotQA} ($n_{\text{test}}{=}300$, HB220/HB224, fp 2026-09-12T00:19:48Z)} \\
GEPA (eq.-call)$^{a}$ & seed program & 16{,}821/n/a/n/a/3{,}000 & 19{,}821 & 7{,}016 &.567 \\
GEPA+Merge & seed program & 600/n/a/n/a/600 & 1{,}200 & 1{,}426 &.421 \\
MIPROv2 & seed program & 600/n/a/n/a/600 & 1{,}200 & 1{,}048 &.435 \\
$M_\omega$ & GEPA-600 init $+$ 68u (60L/8T) $\to$ 28 compiled & 0/16{,}000/700/600 & 17{,}300 & 4{,}619 &.643 (+.077*** [+.045,+.109]) \\
\addlinespace
\multicolumn{6}{l}{\textbf{HoVer} ($n_{\text{test}}{=}300$, HB212, fp 2026-08-20T07:01:05Z)} \\
GEPA (eq.-call) & seed program & 10{,}110/n/a/n/a/3{,}000 & 13{,}110 & 6{,}017 &.497 \\
GEPA+Merge & seed program & 24{,}000/n/a/n/a/600 & 24{,}600 & 7{,}757 &.515 \\
MIPROv2 & seed program & 2{,}400/n/a/n/a/n/l & n/l & 1{,}841 &.487 \\
$M_\omega^{b}$ & GEPA-600 init $+$ 164u (96L/48T/20X) $\to$ 41 compiled & n/l/n/l/n/l/n/l & n/l & 9{,}506 &.567 (+.071*** [+.043,+.099]) \\
\addlinespace
\multicolumn{6}{l}{\textbf{LiveBench} ($n_{\text{test}}{=}126$, HB210, fp 2026-08-20T01:32:58Z)} \\
GEPA (eq.-call) & seed program & 18{,}708/n/a/n/a/1{,}260 & 19{,}968 & 3{,}655 &.695 \\
GEPA+Merge & seed program & 600/n/a/n/a/252 & 852 & 2{,}154 &.686 \\
MIPROv2 & seed program & 600/n/a/n/a/252 & 852 & 436 &.661 \\
$M_\omega^{c}$ & GEPA-600 init; guard rejected all 48u ($\to$ ships GEPA-600 verbatim) & 0/4{,}536/80/252 & 4{,}868 & 5{,}239 &.710 (+.016 [$-$.010,+.043]) \\
\addlinespace
\multicolumn{6}{l}{\textbf{IFBench} ($n_{\text{test}}{=}294$, HB215, fp 2026-08-25T07:48:51Z)} \\
GEPA (eq.-call) & seed program & 23{,}300/n/a/n/a/n/l & n/l & 138 &.417 \\
GEPA+Merge & seed program & 600/n/a/n/a/588 & 1{,}188 & 138 &.419 \\
MIPROv2 & seed program & 600/n/a/n/a/588 & 1{,}188 & 1{,}471 &.382 \\
$M_\omega$ & GEPA-600 init $+$ 128u (60L/68T) $\to$ 23 compiled & 0/22{,}600/700/588 & 23{,}888 & 3{,}856 &.442 (+.025 [$-$.003,+.053]) \\
\addlinespace
\multicolumn{6}{l}{\textbf{PUPA} ($n_{\text{test}}{=}221$, HB217, fp 2026-08-27T13:48:44Z)} \\
GEPA (eq.-call)$^{d}$ & seed program & 29{,}166/n/a/n/a/n/l & n/l & 9{,}109 &.638 \\
GEPA+Merge & seed program & 600/n/a/n/a/n/l & n/l & 2{,}818 &.616 \\
MIPROv2 & seed program & 2{,}400/n/a/n/a/n/l & n/l & 877 &.659 \\
$M_\omega$ & GEPA-600 init $+$ 120u (60L/36T/24F) $\to$ 1 compiled & n/l/n/l/n/l/n/l & n/l & 3{,}512 &.685 (+.047** [+.018,+.076]) \\
\addlinespace
\multicolumn{6}{l}{\textbf{AIME} ($n_{\text{test}}{=}150$, HB219, fp 2026-09-11T09:54:14Z)} \\
GEPA (eq.-call) & seed program & 6{,}870/n/a/n/a/n/l & n/l & 1{,}958 &.323 \\
GEPA+Merge & seed program & 600/n/a/n/a/300 & 900 & 2{,}158 &.331 \\
MIPROv2 & seed program & 600/n/a/n/a/300 & 900 & 63 &.275 \\
$M_\omega$ & GEPA-600 init $+$ 48u (24L/24T) $\to$ 1 compiled & 0/2{,}460/30/300 & 2{,}790 & 1{,}462 &.357 (+.035** [+.011,+.061]) \\
\bottomrule
\end{tabular}}}
\end{center}
\caption{\textbf{Per-procedure accounting.} One block per benchmark; rows are the four Table~\ref{tab:main-results} methods in its column order. \emph{Init/inherited}: GEPA, GEPA+Merge and MIPROv2 all start from the benchmark's seed program; $M_\omega$ initializes from GEPA's shipped 600-call candidate and mines a unit pool from optimizer trajectories plus reflection-LM suggestions (L{=}LLM-suggested, T{=}own-trajectory-mined, X{=}cross-task-LM-trajectory-mined, F{=}failure-grounded-mined; ``u'' units, ``compiled'' the number kept in the shipped prompt after drop-one/add-back and the no-regret guard). \emph{Calls: prop/sel/val/test}: metric-evaluation calls by stage, summed from each run's own \texttt{proposals.jsonl} (prop{=}proposal/mining-stage metric calls, i.e. failure-grounded sampling only, since LLM-suggested units cost reflection-LM calls rather than metric calls, see below; sel{=}select-init/select-marginal/prefix-sweep/drop-one/add-back; val{=}confirm-init/confirm-compiled, the no-regret guard; test{=}this run's own final-test pass, own-session, never the certified score in the last column). GEPA/GEPA+Merge/MIPROv2 do not log a phase-separated selection/validation split, so their ``prop'' cell is their whole declared search budget (\texttt{result.json['budget\_calls']}, the \texttt{max\_metric\_calls} cap) except HotpotQA GEPA (16{,}821, realized, parsed from \texttt{gepa\_state.bin}, HB199 \S1) and PUPA GEPA (29{,}166 realized $=$ 98.8\% of 29{,}526 declared before a disk failure, HB216/217); MIPROv2's heavy mode has no cap, so its number is realized by construction. \texttt{n/a} marks a stage that does not apply to that optimizer; \texttt{n/l} marks a number not logged in this run's own artifacts (IFBench and AIME's GEPA column and HoVer's MIPROv2 column retain no \texttt{proposals.jsonl}, only \texttt{result.json}, so their own-run test-stage calls cannot be recomputed from artifacts here; PUPA's GEPA/Merge/MIPROv2/$M_\omega$ directories and HoVer's $M_\omega$ directory retain none either, so every stage of those rows is \texttt{n/l}). \emph{Total}: realized calls summed across logged stages (LiveBench and AIME's $M_\omega$ directories mix rows from several restarted launch attempts; the number shown isolates the attempt whose own final-test score matches \texttt{result.json}, not the full file, which sums to 16{,}046 and 3{,}855 respectively including the failed restarts). \emph{Chars}: shipped-candidate character count; the aggregate character mass proposed across an entire search is reconstructed only for HotpotQA (HB199 \S1: GEPA${\sim}$70M vs $M_\omega{\sim}$46.3M characters); elsewhere it is not logged. \emph{Test score}: the certified same-session score from the row-of-record ledger entry named per block (never this run's own-session number); $M_\omega$'s row adds the paired bootstrap $\Delta$ and 95\% CI against the GEPA column, stars as in Table~\ref{tab:main-results}. $^{a}$Three independent seeds (0,1,2); this row is seed 0 (declared before seeds 1-2 existed), which scored .567 here against .413/.402 for seeds 1/2 in the same certified session. $^{b}$Two $M_\omega$ selection replicates on the same frozen pool scored $.539$ and $.541$ in the HB226 session, against the row of record's $.572$ (HB221/HB226). $^{c}$$M_\omega$'s compiled candidate never cleared the no-regret guard on LiveBench: the shipped prompt is byte-identical to GEPA's shipped $600$-call candidate (the eq.-call row itself is the $18{,}708$-call run), so this row's test score reflects that candidate re-scored under a different label, not a unit-recombination contribution. $^{d}$GEPA's own run reached 29{,}166 of 29{,}526 declared calls (98.8\%) before an out-of-disk failure recovered by HB216; its own \texttt{result.json} left seed/best test null, so the certified score in this row comes only from the separate row-of-record session. Artifact: \texttt{runs/arm\_accounting\_v1.json}.}
\label{tab:arm-accounting}
\end{table}

\subsection{Estimating the missing value: missing-mass curves, backtest of the value curve, assumptions and calibration of the estimate}
\label{app:mass}

\paragraph{Missing mass estimates how much of the unit population the samplers have not drawn.} Missing mass is the coverage instrument behind every ceiling: it estimates how much of the unit
population $\mathcal{U}$ the samplers have not yet drawn. The Good--Turing statistic $f_1/n$
(the fraction of units seen exactly once) estimates the probability that the \emph{next} draw is
new. When it decays toward zero, sampling has closed its space, and the Chao1 estimator converts
the singleton and doubleton counts into an estimate of the total population size $\widehat N$.
Two presentation details apply to the figures below. First, the raw mined sequence
arrives grouped (by replicate and framing) and is therefore not exchangeable. Every accumulation
curve is averaged over $150$ random re-orderings of the \textit{pooled} draws; the permutation smooths the curve but neither preserves replicate identity nor makes the process independent. The singleton and doubleton counts behind $\widehat N$ (bias-corrected Chao1, $S+f_1(f_1-1)/(2(f_2+1))$) come from the pooled stream. Second, what counts as ``the same'' unit is
set by the clustering granularity: fine clustering treats near-paraphrases as distinct (novelty
stays near $1$), and coarse clustering merges them (the space closes, giving the bounded $\widehat N$
estimates quoted in the captions). We show both because the contrast is itself a finding:
whether a unit space saturates depends on the granularity at which units are individuated.

\begin{figure}[h]
\centering
\resizebox{\textwidth}{!}{%
\begin{tikzpicture}[
  lab/.style={font=\scriptsize\itshape, align=center},
  tlab/.style={font=\tiny\itshape, align=center},
  arr/.style={-{Stealth[length=1.8mm]}, thick},
  darr/.style={-{Stealth[length=1.5mm]}, thin, black!60},
]
\begin{scope}
  \draw[arr] (0,0) -- (0,3.2); \draw[arr] (0,0) -- (4.1,0);
  \node[tlab, rotate=90] at (-0.78,1.55) {$P(\mathrm{next\ unit\ unseen})$};
  \node[tlab] at (2.0,-0.42) {discovery pool size $N$};
  \IfFileExists{figs/gen_mass_supervised.tex}{\draw (-0.06,0.12) -- (0.05,0.12); \node[tlab, anchor=east] at (-0.09,0.12) {0.00};
\draw (-0.06,0.83) -- (0.05,0.83); \node[tlab, anchor=east] at (-0.09,0.83) {0.25};
\draw (-0.06,1.54) -- (0.05,1.54); \node[tlab, anchor=east] at (-0.09,1.54) {0.50};
\draw (-0.06,2.26) -- (0.05,2.26); \node[tlab, anchor=east] at (-0.09,2.26) {0.75};
\draw (-0.06,2.97) -- (0.05,2.97); \node[tlab, anchor=east] at (-0.09,2.97) {1.00};
\draw (0.67,-0.05) -- (0.67,0.05); \node[tlab] at (0.67,-0.2) {$10^{1}$};
\draw (3.03,-0.05) -- (3.03,0.05); \node[tlab] at (3.03,-0.2) {$10^{2}$};
\draw[red!70!black, densely dotted] (0.15,0.41) -- (3.95,0.41);
\draw[red!65!black, opacity=0.18, thick] plot[smooth] coordinates {(0.15,2.82) (0.86,2.65) (1.27,2.50) (1.57,2.43) (1.80,2.36) (1.98,2.29) (2.14,2.24) (2.28,2.20)};
\draw[red!65!black, opacity=0.18, thick] plot[smooth] coordinates {(0.15,2.89) (0.86,2.82) (1.27,2.69) (1.57,2.58) (1.80,2.49) (1.98,2.38) (2.14,2.28) (2.28,2.20)};
\draw[blue!60!black, very thick] plot[smooth] coordinates {(0.15,2.85) (0.86,2.71) (1.27,2.57) (1.57,2.41) (1.80,2.29) (1.98,2.16) (2.14,2.04) (2.28,1.96) (2.40,1.87) (2.51,1.80) (2.60,1.73) (2.69,1.64) (2.77,1.57) (2.85,1.52) (2.92,1.46) (2.99,1.41) (3.05,1.36) (3.11,1.30) (3.16,1.25) (3.22,1.22) (3.27,1.17) (3.31,1.13) (3.36,1.09) (3.40,1.06) (3.44,1.02) (3.48,0.99) (3.52,0.96) (3.56,0.93) (3.60,0.90) (3.63,0.88) (3.66,0.86) (3.70,0.83) (3.73,0.81) (3.76,0.79) (3.79,0.77) (3.82,0.76) (3.84,0.74) (3.87,0.73) (3.90,0.72) (3.92,0.71) (3.95,0.70)};
\draw[red!65!black, very thick] plot[smooth] coordinates {(0.15,2.88) (0.86,2.83) (1.27,2.77) (1.57,2.74) (1.80,2.66) (1.98,2.60) (2.14,2.56) (2.28,2.51) (2.40,2.46) (2.51,2.42) (2.60,2.38) (2.69,2.34) (2.77,2.29) (2.85,2.26) (2.92,2.21) (2.99,2.18) (3.05,2.13) (3.11,2.10) (3.16,2.07) (3.22,2.04) (3.27,2.01) (3.31,1.98) (3.36,1.95) (3.40,1.93) (3.44,1.91) (3.48,1.88) (3.52,1.87)};
\draw[blue!60!black, very thick] plot[smooth] coordinates {(0.15,2.84) (0.86,2.60) (1.27,2.45) (1.57,2.31) (1.80,2.13) (1.98,2.02) (2.14,1.89) (2.28,1.80) (2.40,1.69) (2.51,1.60) (2.60,1.50) (2.69,1.43) (2.77,1.36) (2.85,1.28) (2.92,1.21) (2.99,1.16) (3.05,1.10) (3.11,1.05) (3.16,1.00) (3.22,0.95) (3.27,0.91) (3.31,0.86) (3.36,0.82) (3.40,0.79) (3.44,0.75) (3.48,0.72) (3.52,0.69) (3.56,0.66) (3.60,0.63) (3.63,0.61) (3.66,0.58) (3.70,0.56) (3.73,0.54) (3.76,0.52) (3.79,0.50)};
\node[tlab, anchor=south west, align=left, draw=black!30, fill=white, inner sep=2pt] at (2.45,0.15) {\textcolor{blue!60!black}{\rule[0.5ex]{3mm}{1pt}} saturating (2) \\[0.2mm] \textcolor{red!65!black}{\rule[0.5ex]{3mm}{1pt}} climbing (3)};
\node[tlab, blue!60!black, anchor=west, fill=white, draw=black!20, rounded corners=1pt, inner sep=1.5pt] (a142) at (0.22,1.42) {hotpot};
\draw[darr, blue!60!black] (a142.east) -- (3.56,0.93);
\node[tlab, red!65!black, anchor=west, fill=white, draw=black!20, rounded corners=1pt, inner sep=1.5pt] (a109) at (0.22,1.10) {hover};
\draw[darr, red!65!black] (a109.east) -- (3.16,2.07);
\node[tlab, blue!60!black, anchor=west, fill=white, draw=black!20, rounded corners=1pt, inner sep=1.5pt] (a77) at (0.22,0.78) {ifbench};
\draw[darr, blue!60!black] (a77.east) -- (3.40,0.79);}{}
  \node[lab] at (2.0,3.45) {supervised benchmarks};
\end{scope}
\begin{scope}[shift={(5.6,0)}]
  \draw[arr] (0,0) -- (0,3.2); \draw[arr] (0,0) -- (4.1,0);
  \node[tlab, rotate=90] at (-0.78,1.55) {$P(\mathrm{next\ rubric\ unseen})$};
  \node[tlab] at (2.0,-0.42) {discovery pool size $N$};
  \IfFileExists{figs/gen_mass_unsupervised.tex}{\input{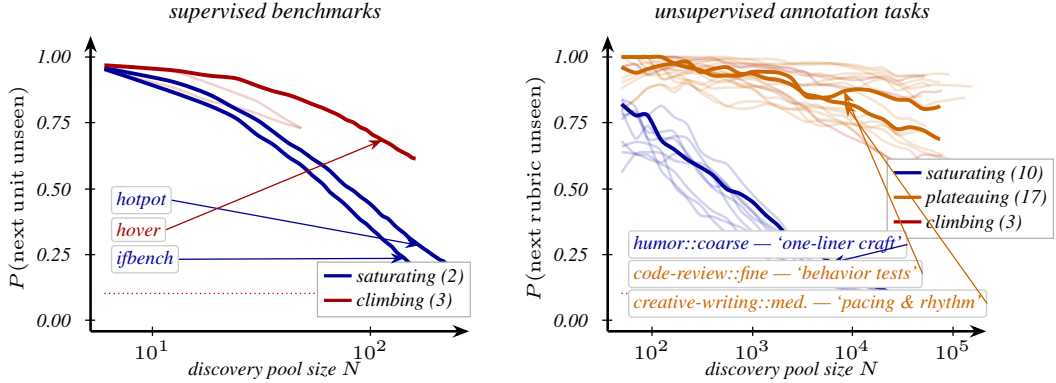}}{}
  \node[lab] at (2.0,3.45) {unsupervised annotation tasks};
\end{scope}
\end{tikzpicture}}%
\caption{\textbf{Missing mass.} Good--Turing novelty $p_{\mathrm{unseen}}(n)=f_1/n$ as mining
proceeds, per task; the curves show data only, with no extrapolation. Accumulation curves are averaged over $150$
random orderings. \emph{Left:} five of the six supervised GEPA-suite benchmarks (all but PUPA, added after this sampling pass); species are
semantic unit clusters. \emph{Right:} unsupervised annotation tasks; species are normalized
rubric items from each task's task bank, and three tasks are labeled by name. Curves fall into
\emph{saturating} (the articulable space closes), \emph{plateauing} (novelty persists at a steady
rate), and \emph{climbing} (open-ended drift) regimes.}
\label{fig:mass-value}
\end{figure}

\begin{figure}[h]
\centering
\includegraphics[width=\textwidth]{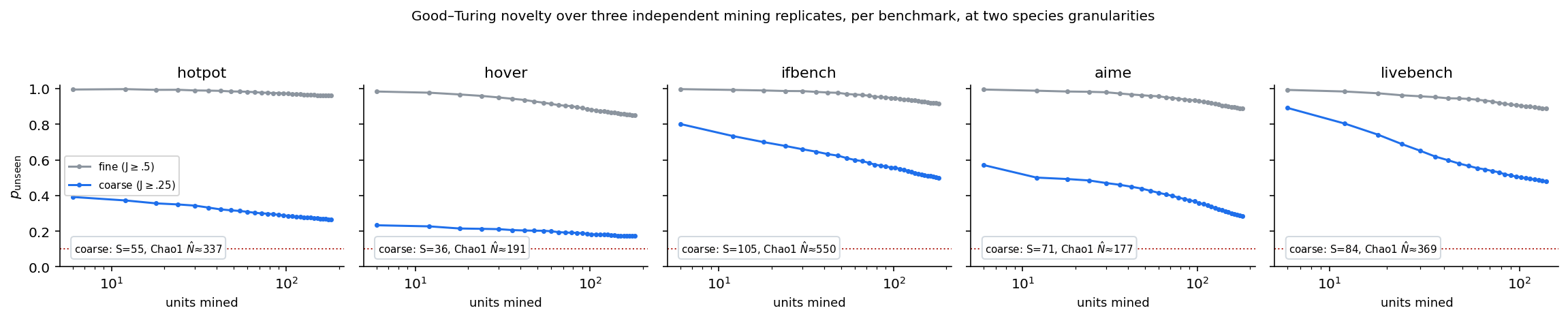}
\caption{\textbf{Missing mass for five of the six supervised benchmarks} (all but PUPA), from three independent mining
replicates per benchmark ($3{\times}60$ units, one declared sampler). Species are formed by
deterministic word-overlap clustering at two granularities. Curves are averaged over $150$ random orderings (the raw sequence arrives grouped by framing, so unaveraged $f_1/n$ is non-exchangeable and can even rise). At the \emph{fine} granularity (Jaccard $\geq.5$, grey) nearly every unit is its own species and novelty stays near $1$; the articulable space appears unbounded. At the \emph{coarse} granularity (Jaccard $\geq.25$, blue)
the same data yield bounded estimates: HoVer's space is smallest and closes fastest (Chao1
$\widehat N{\approx}191$), and IFBench's remains the most open ($\widehat N{\approx}550$).
Within every panel, the contrast between the two lines shows that \emph{whether a
prompt-unit space saturates is a property of the granularity at which units are individuated,
not of the task}. Syntactic clustering is conservative for missing-mass claims (paraphrases a
semantic clusterer would merge stay separate, inflating $f_1$ and hence $p_{\mathrm{unseen}}$);
these curves are upper bounds on novelty.}
\label{fig:appendix-mass}
\end{figure}

\paragraph{Backtesting the value projection.}\label{app:backtest}
\begin{figure}[h]
\centering
\def\FigThreeLhsTitleXY{(2.0,3.55)}    %
\def\FigThreeLhsSubXY{(2.0,3.38)}      %
\def\FigThreeLhsYlabXY{(-0.86,1.55)}   %
\def\FigThreeLhsXlabXY{(2.0,-0.42)}    %
\def\FigThreeRhsXshift{5.9cm}          %
\def\FigThreeRhsYshift{0cm}            %
\def\FigThreeRhsTitleXY{(2.0,3.55)}    %
\def\FigThreeRhsYlabXY{(-0.80,1.55)}   %
\def\FigThreeRhsXlabXY{(2.0,-0.42)}    %
\resizebox{.85\textwidth}{!}{%
\begin{tikzpicture}[lab/.style={font=\scriptsize\itshape, align=center},
  tlab/.style={font=\tiny\itshape, align=center},
  arr/.style={-{Stealth[length=1.8mm]}, thick},
  darr/.style={-{Stealth[length=1.5mm]}, thin, black!60}]
\begin{scope}
  \node[tlab, rotate=90] at \FigThreeLhsYlabXY {task metric};
  \node[tlab] at \FigThreeLhsXlabXY {units included $k$};
  \IfFileExists{figs/gen_value_supervised.tex}{\draw[-{Stealth[length=1.6mm]}] (0,0.12) -- (4.10,0.12);
\draw[-{Stealth[length=1.6mm]}] (0,0.12) -- (0,3.20);
\draw (-0.06,0.12) -- (0.05,0.12); \node[tlab, anchor=east] at (-0.09,0.12) {0.54};
\draw (-0.06,0.81) -- (0.05,0.81); \node[tlab, anchor=east] at (-0.09,0.81) {0.65};
\draw (-0.06,1.49) -- (0.05,1.49); \node[tlab, anchor=east] at (-0.09,1.49) {0.77};
\draw (-0.06,2.18) -- (0.05,2.18); \node[tlab, anchor=east] at (-0.09,2.18) {0.89};
\draw (-0.06,2.86) -- (0.05,2.86); \node[tlab, anchor=east] at (-0.09,2.86) {1.00};
\draw (0.15,0.06) -- (0.15,0.18); \node[tlab] at (0.15,-0.06) {1};
\draw (1.36,0.06) -- (1.36,0.18); \node[tlab] at (1.36,-0.06) {20};
\draw (2.63,0.06) -- (2.63,0.18); \node[tlab] at (2.63,-0.06) {40};
\draw (3.90,0.06) -- (3.90,0.18); \node[tlab] at (3.90,-0.06) {60};
\fill[orange!60, opacity=0.18] plot coordinates {(1.36,1.05) (1.61,1.18) (1.87,1.31) (2.12,1.43) (2.37,1.55) (2.63,1.67) (2.88,1.79) (3.14,1.90) (3.39,2.02) (3.65,2.13) (3.90,2.24)} -- plot coordinates {(3.90,0.51) (3.65,0.52) (3.39,0.53) (3.14,0.53) (2.88,0.54) (2.63,0.54) (2.37,0.54) (2.12,0.54) (1.87,0.53) (1.61,0.53) (1.36,0.52)} -- cycle;
\draw[orange!80!black, thick, densely dashed] plot coordinates {(1.36,0.78) (1.61,0.85) (1.87,0.92) (2.12,0.99) (2.37,1.05) (2.63,1.11) (2.88,1.16) (3.14,1.22) (3.39,1.27) (3.65,1.32) (3.90,1.37)};
\draw[blue!60!black, opacity=0.35, thick] plot coordinates {(1.42,0.95) (1.48,0.90) (1.55,0.90) (1.61,0.90) (1.68,0.90) (1.74,0.84) (1.80,1.01) (1.87,0.95) (1.93,0.72) (1.99,0.72) (2.06,0.54) (2.12,0.66) (2.18,0.78) (2.25,0.66) (2.31,0.78) (2.37,0.72) (2.44,0.78) (2.50,0.78) (2.57,1.07) (2.63,0.90)};
\fill[blue!60!black, opacity=0.35] (1.42,0.95) circle (0.8pt);
\fill[blue!60!black, opacity=0.35] (1.55,0.90) circle (0.8pt);
\fill[blue!60!black, opacity=0.35] (1.68,0.90) circle (0.8pt);
\fill[blue!60!black, opacity=0.35] (1.80,1.01) circle (0.8pt);
\fill[blue!60!black, opacity=0.35] (1.93,0.72) circle (0.8pt);
\fill[blue!60!black, opacity=0.35] (2.06,0.54) circle (0.8pt);
\fill[blue!60!black, opacity=0.35] (2.18,0.78) circle (0.8pt);
\fill[blue!60!black, opacity=0.35] (2.31,0.78) circle (0.8pt);
\fill[blue!60!black, opacity=0.35] (2.44,0.78) circle (0.8pt);
\fill[blue!60!black, opacity=0.35] (2.57,1.07) circle (0.8pt);
\draw[blue!60!black, very thick] plot coordinates {(0.15,0.48) (0.21,0.42) (0.28,0.36) (0.34,0.42) (0.40,0.60) (0.47,0.60) (0.53,0.36) (0.59,0.60) (0.66,0.66) (0.72,0.78) (0.79,0.72) (0.85,0.54) (0.91,0.42) (0.98,0.48) (1.04,0.54) (1.10,0.90) (1.17,0.95) (1.23,0.78) (1.29,0.78) (1.36,0.72)};
\fill[blue!60!black] (0.15,0.48) circle (1pt);
\fill[blue!60!black] (0.28,0.36) circle (1pt);
\fill[blue!60!black] (0.40,0.60) circle (1pt);
\fill[blue!60!black] (0.53,0.36) circle (1pt);
\fill[blue!60!black] (0.66,0.66) circle (1pt);
\fill[blue!60!black] (0.79,0.72) circle (1pt);
\fill[blue!60!black] (0.91,0.42) circle (1pt);
\fill[blue!60!black] (1.04,0.54) circle (1pt);
\fill[blue!60!black] (1.17,0.95) circle (1pt);
\fill[blue!60!black] (1.29,0.78) circle (1pt);
\node[tlab, blue!60!black, anchor=west, fill=white, draw=black!20, rounded corners=1pt, inner sep=1.6pt] at (0.21,1.25) {observed $k\le 20$};
\node[tlab, blue!60!black, anchor=west, fill=white, draw=black!20, rounded corners=1pt, inner sep=1.6pt] at (2.41,0.46) {actual $k>20$};
\node[tlab, orange!70!black, anchor=west, fill=white, draw=black!20, rounded corners=1pt, inner sep=1.6pt] at (1.85, 2.16) {projection ($k > 20$)};}{}
  \node[lab] at \FigThreeLhsTitleXY {Supervised task (shown here, Hotpot): \\ Prompt Performance Increases};
\end{scope}
\begin{scope}[xshift=\FigThreeRhsXshift, yshift=\FigThreeRhsYshift]
  \draw[arr] (0,0.12) -- (0,3.2); \draw[arr] (0,0.12) -- (4.1,0.12);
  \node[tlab, rotate=90] at \FigThreeRhsYlabXY {bits};
  \node[tlab] at \FigThreeRhsXlabXY {metrics, ordered by achieved bits};
  \IfFileExists{figs/gen_value_unsupervised.tex}{\input{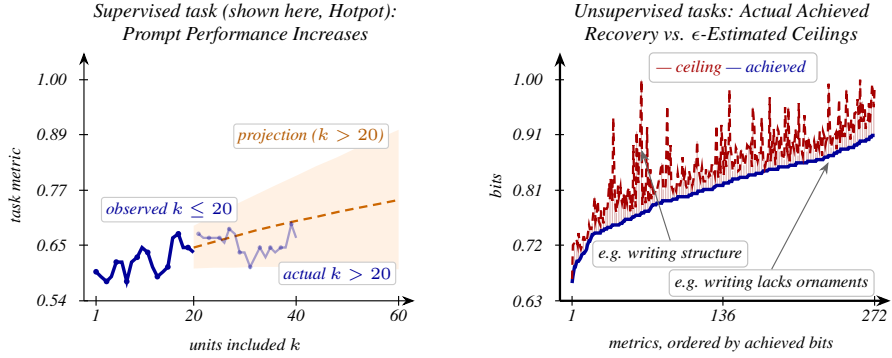}}{%
    \node[tlab, red!60!black] at (2.0,1.6) {[pending: unseen-value accumulation\\from the metric-bank value artifacts]};}
  \node[lab] at \FigThreeRhsTitleXY {Unsupervised tasks: Actual Achieved \\ Recovery vs. $\epsilon$-Estimated Ceilings};
\end{scope}
\end{tikzpicture}}%
\caption{\textbf{Missing value measurements.} \emph{Left, supervised tasks:} value at each \# units $k=1,2...$ (blue). A forward projection $\widehat V(k)$, $k>20$ is made using $\epsilon$-estimation (orange dashed, $95\%$ band widening with horizon), and the actual $k>20$ drawn underneath. HotpotQA (fit $k\le 20$, actual to $40$, projection to $60$) is the still-rising regime. Backtesting the projection procedure over
all six benchmarks' trajectories (Appendix~\ref{app:backtest}) gives pooled median prediction error $.018$ at horizons of $1$--$3$ units ahead, $.020$ at $4$--$8$, and $.029$ at $9{+}$. \emph{Right, unsupervised:} per-task value (blue) and its missing-value ceiling (red), for $272$ metrics across $8$ creative domains, estimated by asking what a \emph{fresh draw} from the same sampling process could still add.}
\label{fig:missing-value}
\end{figure}
We test how well the fitted value curves of Figure~\ref{fig:missing-value} predict \emph{future}
behavior. We backtest on the greedy prefix trajectories of all six benchmarks: for every fit-window
size $j \ge 5$, we fit the curve on the first $j$ points only, predict every later point, and record
the absolute error bucketed by horizon (units ahead of the window). We compare two functional
forms: saturating exponential $y = a - b\,r^{k}$ and power law $y = a - b\,k^{-c}$.

\begin{center}
\begin{tabular}{lcccccc}
\toprule
 & \multicolumn{3}{c}{power law, median $|$err$|$} & \multicolumn{3}{c}{sat.\ exp., median $|$err$|$} \\
benchmark & $1$--$3$ & $4$--$8$ & $9{+}$ & $1$--$3$ & $4$--$8$ & $9{+}$ \\
\midrule
HotpotQA ($k \le 68$) & $.019$ & $.022$ & $.034$ & $.017$ & $.021$ & $.062$ \\
AIME ($k \le 48$) & $.024$ & $.023$ & $.019$ & $.020$ & $.019$ & $.020$ \\
IFBench ($k \le 48$) & $.014$ & $.016$ & $.019$ & $.014$ & $.016$ & $.016$ \\
HoVer ($k \le 32$) & $.023$ & $.025$ & $.028$ & $.023$ & $.030$ & $.054$ \\
LiveBench ($k \le 32$) & $.023$ & $.030$ & $.052$ & $.025$ & $.035$ & $.063$ \\
PUPA ($k \le 23$) & $.012$ & $.011$ & $.012$ & $.011$ & $.010$ & $.011$ \\
\midrule
pooled & $.018$ & $.020$ & $.029$ & $.018$ & $.021$ & $.043$ \\
\bottomrule
\end{tabular}
\end{center}

Median error is ${\approx}.02$ at short horizons and remains under $.03$ (power law) at horizons of $9{+}$
units; at long horizons the power-law error grows more slowly than the saturating-exponential
error. The error distribution is
heavy-tailed for small fit windows (pooled max error $.43$ for the power law, $2.1$ for the
saturating exponential, both from $5$-point fits on LiveBench). The backtest therefore supports a calibrated median
error, not a worst-case guarantee: fitted on ${\ge}8$--$10$ prefix points, the curve predicts
$4$--$8$ units ahead with median error ${\approx}.02$--$.03$. We attempted a direct extreme-value estimate of the score
distribution's endpoint and dropped it, because the fitted tail pinned the endpoint to
the sample maximum on every benchmark. Artifact:
\texttt{runs/ceiling\_backtest\_all.json}.

\paragraph{The five assumptions behind $\widehat\epsilon$.}\label{app:eps-assumptions}
Section~\ref{sec:e_certifiability} lists five assumptions behind $\widehat\epsilon$. This section reports what each assumption costs, which ones we could correct from existing artifacts, and the corrected estimate (artifact \texttt{runs/epsilon\_calibration\_v1.json}; note \texttt{notes/2026-09-11\_\_epsilon-calibration.md}).

\textit{A1, sampler support.} The same benchmark pool has $\widehat N=337$ under the Qwen3-8B sampler and $1{,}061$ under a frontier sampler; the unit universe depends on the proposer, and the ceiling is relative to the declared samplers.

\textit{A2, discovery probability and value.} Units found once have more value than units found three or more times on four of five benchmarks (HotpotQA $+.027$ against $+.004$; AIME $+.044$ against $+.033$), so pricing unseen units by the curve's decaying tail undervalues them. Pricing them at the singleton mean instead gives a number in the hundreds, because $\widehat N-S$ is large, and is clipped to the asymptote gap on every benchmark; the unseen units cannot be priced one by one from these data.

\textit{A3, exchangeability.} Resampling whole sampling runs replicates pooled draws and moves the Chao1 estimate by an order of magnitude (HotpotQA $\widehat N=337$ pooled; leave-one-replicate-out range $159$--$498$; block-bootstrap lower limit $23$), and the Good--Turing novelty's lower limit reaches zero. The pooled $\widehat N$ is one point in a wide range, and $\widehat\epsilon$ inherits that width.

\textit{A4, submodularity ratio.} Measured in Appendix~\ref{app:submodular}: no evaluated prefix shows synergy, and the within-pool gap is at most $.01$ on HotpotQA and $0$ on IFBench.

\textit{A5, imposed saturation.} The fitters allowed $b<0$, so a descending curve could report its floor as the asymptote; this is why AIME's point estimate ($.480$) was below its realized maximum ($.633$) and $\widehat\epsilon$ was negative on two benchmarks. AIME's power-law fit has $b=-.098$, and its $a=.480$ is also below the trajectory mean ($.536$). AIME's apparent inverted-U is a two-to-three-item fluctuation on a score grid of $1/30$. We now constrain $b\ge 0$ and $a\ge\max y$. A non-saturating form ($a+b\log k$, evaluated at $k_{\max}{+}50$) is the low end of the range on every benchmark.

\paragraph{One-sided coverage of the ceiling.} Median error is two-sided, and a ceiling needs a one-sided check. For every fit window we therefore count how often a later prefix \textit{exceeded} the ceiling the window implied (artifact \texttt{runs/coverage\_backtest\_v1.json}; windows are nested prefixes, so binomial limits are too narrow and a few early windows drive the pooled rates). Over every fit window of at least $5$ points and pooled over benchmarks and horizons (power law), the point forecast is exceeded $43\%$ of the time, close to the $50\%$ of an unbiased fit. The bare fitted asymptote is exceeded $27\%$ of the time, against the $5\%$ a $\delta=.05$ ceiling allows (HotpotQA $.16$, IFBench $.18$, AIME $.31$, PUPA $.45$, LiveBench $.55$, HoVer $.56$). The asymptote plus a $95\%$ residual quantile is exceeded $9\%$ of the time, and a refit constrained to lie at or above the best observed value $8\%$. The bare asymptote is therefore not a $1-\delta$ bound. The two widened versions come close but still miss on some benchmark-horizon pairs (HoVer, saturating exponential, $.32$--$.44$ from $4$ units on; HotpotQA constrained, $.17$ at $9{+}$). With the constraints of A5 and fit windows of at least $8$ points, later prefixes exceed the fitted ceiling $2.4\%$ of the time pooled at horizons $1$--$8$, so a leave-one-benchmark-out calibration of an added margin returns zero width in every fold, because the constraints alone achieve this coverage. Held out, IFBench still exceeds its ceiling $8.9\%$ of the time at horizons $1$--$8$ and HotpotQA $8.4\%$ at $9{+}$; a margin sized on the worst benchmark widens $\widehat\epsilon$ by $.006$--$.023$ and does not correct IFBench. We report the constrained estimate $\widehat\epsilon_{\mathrm{con}}$ with its measured miss rate; it is calibrated on six trajectories, not proven.

\paragraph{Projected and constrained ceilings.} The same fitted curves can be extended beyond
the observed trajectory: $\hat y(k_{\max}{+}\Delta)$ for finite look-ahead, and the fitted
asymptote $a$ as the $k{\to}\infty$ ceiling (both forms are $y = a - \text{decaying term}$, so
$a$ \emph{is} the infinite-prefix limit). Table~\ref{tab:ceilings} gives both forms fitted on each benchmark's full canonical
trajectory, with the constrained asymptote and the resulting $\widehat\epsilon$.

\begin{table}[h]
\small
\begin{center}
\resizebox{\textwidth}{!}{%
\begin{tabular}{lrccccccccccc}
\toprule
 & & & \multicolumn{3}{c}{power law $\hat y(k_{\max}{+}\Delta)$} & \multicolumn{3}{c}{$k{=}\infty$ ceiling $a$} & & & & \\
\cmidrule(lr){4-6}\cmidrule(lr){7-9}
benchmark & $k_{\max}$ & $y(k_{\max})$ & $+10$ & $+20$ & $+50$ & power & sat.\ exp. & constrained & $\widehat\epsilon$ (point) & $\widehat\epsilon_{\mathrm{con}}$ & viol.\ $4$--$8$ / $9{+}$ & $\widehat N$ pooled [block] \\
\midrule
HotpotQA & 68 & $.593$ & $.602$ & $.602$ & $.604$ & $.611$ & $.600$ & $.644$ & $+.018$ & $+.051$ & $.03$ / $.08$ & $337$ $[23,337]$ \\
IFBench & 48 & $.855$ & $.856$ & $.856$ & $.856$ & $.856$ & $.856$ & $.892$ & $+.001$ & $+.037$ & $.10$ / $.05$ & $550$ $[42,550]$ \\
HoVer & 32 & $.560$ & $.592$ & $.592$ & $.592$ & $.592$ & $.570$ & $.660$ & $+.032$ & $+.100$ & $.00$ / $.00$ & $191$ $[16,191]$ \\
LiveBench & 32 & $.658$ & $.718$ & $.721$ & $.726$ & $.778$ & $.723$ & $.778$ & $+.121$ & $+.121$ & $.00$ / $.00$ & $369$ $[37,369]$ \\
AIME & 48 & $.533$ & $.524$ & $.523$ & $.520$ & $.480$ & $.445$ & $.633$ & $-.053$ & $+.100$ & $.00$ / $.00$ & $177$ $[33,177]$ \\
PUPA & 23 & $.891$ & $.889$ & $.889$ & $.889$ & $.889$ & $.888$ & $.952$ & $-.002$ & $+.061$ & $.00$ / $.00$ & --- \\
\bottomrule
\end{tabular}}
\end{center}
\caption{\textbf{Projected and constrained ceilings}, one row per benchmark. Values are in the value curves' own units
(validation-side greedy prefix means, as in Figure~\ref{fig:missing-value} and the
backtest above), not Table~\ref{tab:main-results} test scores; only the saturation verdict
transfers from one to the other. The constrained asymptote, both $\widehat\epsilon$ columns and the violation rates are power-law fits. $\widehat\epsilon_{\mathrm{con}}$ is the constrained asymptote minus $y(k_{\max})$; the calibrated margin is zero. Violation = share of later prefixes above the window's ceiling, held out. The block interval resamples whole sampling runs; its upper limit necessarily equals the pooled point (a duplicated run adds only doubletons), so the leave-one-run-out range above is the better measure of width. PUPA has a single sampling run. The constrained asymptotes replace the unconstrained ones wherever a ceiling is quoted. Artifacts: \texttt{runs/ceiling\_projection\_v1.json} and \texttt{runs/epsilon\_calibration\_v1.json}.}
\label{tab:ceilings}
\end{table}

(1) \emph{Saturation licenses the limit}: every fitted
trajectory's last-quintile slope is ${\le}.0004$/unit, so the asymptote is identified by the
data rather than extrapolated through a rise; had any curve still been rising, its
$k{=}\infty$ cell would be reported only as a censored lower bound (the same convention the
scaling analysis uses for \emph{rising} tasks). Projections move the curve by ${\le}.01$
at $+50$ on four of six benchmarks; the exceptions are HoVer (the fit smooths a noisy final
point, $+.03$) and LiveBench (the noisiest trajectory; its $.055$ power-vs-exponential
asymptote spread is the width of its ceiling). (2) \emph{Validated horizon}: the backtest
above supports median error only to ${\sim}9{+}$ units ahead; $+50$ and $\infty$ inherit no
error guarantee beyond the saturation argument. (3) \emph{Scope}: every ceiling is
conditional on the sampling pool. It is the limit of continued draws from \emph{this} process,
never a bound over all prompts (the all-prompt supremum is vacuously $1.0$ on
deterministic-label benchmarks; the extreme-value attempt is reported with the backtest above).

\textit{What the constrained ceiling says.} The constraint $a\ge\max y$ binds on five of the six benchmarks: there the constrained asymptote equals the highest point of the sweep, so $\widehat\epsilon_{\mathrm{con}}$ is the sweep's fall from its peak to its last point, not value the fit projects beyond what was seen. Only LiveBench's curve projects above its own maximum, by $.010$. The ceiling is $a_{\mathrm{con}}$ itself, not the best score plus $\widehat\epsilon_{\mathrm{con}}$; on five benchmarks the fitted curve finds no value beyond the best prefix, and the $\widehat N-S$ unseen units are priced at zero by construction.
\section{The reconstruction instrument: choice of measure and validity}
\label{app:recon-instrument}

\subsection{Choice of measure, and checks on the relabeling channel}
\label{app:recon-alternatives}
We considered three measures of the receiver's understanding. \textbf{(1) Prompt comparison.} Ask a judge whether $p'$ ``means the same thing'' as $p$. This is the measure that language leaks through: the labeled examples carry lexical echoes of the criterion's register, so a receiver can emit a $p'$ that sounds right without holding the decision rule, and a similarity judge rewards the sound. Judged similarity is also wording-sensitive in the way this paper studies: re-specifying the same concept (name alone vs.\ name with definition) flipped $46\%$ of arbiter verdicts in our label checks (Appendix~\ref{app:mi-silver}), so similar wording is neither necessary nor sufficient for the same rule. We use prompt similarity only as a per-run sanity check (Section~\ref{sec:reconstruction}), never as the score. \textbf{(2) Identification.} Show the receiver the labeled examples and ask it to pick $p$ from a set of hard distractors. This is cheap and we keep it in the harness as a screen, but it measures discrimination, not operationalization: a receiver can eliminate distractors on surface grounds while being unable to apply the rule to fresh items. In our pilots the identification measure also ranked metrics differently from the behavioral one (agreement with an independent certificate: $\rho=.65$ behavioral vs.\ $.18$ identification), so no number in this paper uses it. A decoy-label control shows why (protocol HB\,222, artifact \texttt{outputs/decoy\_control/}). For $60$ bank metrics ($20$ per regime, seven annotation tasks) we replaced the demonstrations' labels with those of a balanced planted rule (length, question mark, digit, quotation) and kept the same four options, so the intended criterion was still offered although the transmitter had not applied it. Llama-3.1-8B still picked the intended criterion $45\%$ of the time (chance $25\%$; $n=56$) against $47\%$ with the real labels, and Qwen2.5-14B $64\%$ against $73\%$ ($n=30$; on the other $30$ metrics the executor's labels did not vary across items). When the decoy rule's own description was offered as a fifth option, the receivers chose it $5\%$ and $17\%$ of the time, below chance. Identification picks the plausible option almost regardless of the labels. This control does not test the relabeling channel. We test that channel separately, and it is not fooled the same way (protocol HB\,228, artifact \texttt{outputs/analyses/behavioral\_decoy\_v1/}). Sending the planted rule's labels through the channel of Section~\ref{sec:reconstruction} instead of the criterion's, on the same $60$ metrics and items, recovers $.013$ bits against $.044$ for the criterion at Llama-3.1-8B (decoy minus criterion $-.031$ $[-.043,-.022]$ over the $55$ metrics with defined mutual information) and $.009$ against $.024$ at Qwen2.5-14B ($-.015$ $[-.029,-.004]$, $31$ metrics; the executor's labels were constant on the rest). The decoy labels are unrelated to the criterion's labels (AUC $.51$ and $.48$), and the rules the receiver writes from decoy-labeled demonstrations are about the task, not the planted surface feature. A balanced surface rule therefore does not produce high recovery for these receivers. This result does not show that a high recovery reflects the intended criterion. \textbf{(3) Behavioral relabeling}, the channel used throughout: re-execute $p'$ on unseen items and measure $I(m;m')$. Understanding is credited only to the degree it can be operationalized, which is the criterion the language game of Section~\ref{sec:reconstruction-alg} motivates.

\paragraph{Reversed reconstructions are $8\%$ of rows and carry $1.3\%$ of the mutual information.} Because $I(m;m')$ ignores orientation, we recomputed every examples-bank reconstruction ($236$ metrics $\times$ four message forms $\times$ two judges, $1{,}574$ rows, labels thresholded at $.5$) with oriented statistics beside the MI (artifact \texttt{outputs/analyses/recon\_polarity\_audit\_v1/}). Reversed reconstructions (negative Matthews $\phi$) are $128$ of $1{,}574$ rows and account for $1.3\%$ of the total MI ($0.7\%$ at $\phi<-.2$); MI and balanced accuracy correlate at Spearman $.83$ across metrics, $.82$ without the reversed rows. Most reversals are not sign errors: only $8$ of $128$ have both base rates inside $[.25,.75]$; the typical case is a transmitter that rarely says YES (median rate $.11$) met by a receiver that says YES to most items (median $.73$). A further $456$ rows have $\phi$ undefined because one side is constant after thresholding; they account for $0.5\%$ of the MI. Mean balanced accuracy of the definition form is $.61$ (Llama-70B) and $.66$ (Qwen-72B).

\paragraph{The receiver executes its own rule.} In our score the receiver writes the rule $p'$ and the transmitter runs it: the receiver's rule counts only through the labels $m'$ the transmitter produces when it runs that rule, scored against the transmitter's own labels $m$. To check that this last step is not what caps recovery, we also let the receiver execute its own rule on the held-out items and scored $I(m; m')$ against the same transmitter labels ($60$ criteria across seven domains, $18$ of them bounded; Llama-3.1-8B transmitter, Qwen2.5-14B receiver, $R=8$ reconstructions; artifact \texttt{outputs/decoy\_control/receiver\_executes\_v1.json}). The receiver recovers less of the transmitter's labeling by running its rule itself: $.040$ bits against $.052$ when the transmitter runs it, a difference of $-.013$ $[-.023,-.003]$ over the $55$ criteria with defined values, with the receiver ahead on $42\%$; the two scores rank criteria alike (Spearman $.61$). By regime, the transmitter's execution is ahead for reaches ($.073$ against $.037$) and rising ($.053$ against $.041$) criteria and behind for bounded ones ($.030$ against $.041$). The transmitter's re-execution is not what bounds recovery.

\paragraph{Transfer across model families, prompt-author families and planted rules.} Per-metric recovery levels correlate across model families, with Llama-3.1-70B and Qwen2.5-72B as transmitter and receiver, at Spearman $\rho=.61$--$.69$. We test whether sharing a model family between transmitter and receiver inflates recovery. The message-form ladder already contained this test (artifact \texttt{outputs/analyses/fam\_v1\_author\_executor\_v1/}): for $52$ concepts in four annotation tasks, the explanation and full packet messages were written by both Llama-3.1-70B and Qwen2.5-72B, executed by nine receivers (Llama 1B--70B, Qwen2.5 3B--72B), and scored as in Figure~\ref{fig:message-forms} (balanced agreement with a leave-one-out consensus whose target was written by a third family, Claude Sonnet). $35$ of the $52$ concepts pass the class-balance filter (humor $25/36$, creative writing $5/6$, math $4/5$, peer review $1/5$).

\begin{center}
{\small
\begin{tabular}{lccc}
\toprule
executor family & Llama-authored & Qwen-authored & cells \\
\midrule
Llama & $.648$ & $.647$ & $256$ \\
Qwen2.5 & $.695$ & $.700$ & $326$ \\
\bottomrule
\end{tabular}}
\end{center}

Same-family minus cross-family agreement, paired over concepts ($2{,}000$ bootstrap draws): $+.003$ $[-.002,+.008]$; author-by-executor interaction $+.006$ $[-.004,+.016]$; author main effect $-.003$. On the same cells a full packet beats a name by $+.026$ $[+.004,+.049]$ ($+.059$ $[+.030,+.087]$ at $7$B and above), so a family effect a fraction of that size would have been detected. The three smallest receivers prefer Llama-written text whatever their own family ($+.010$ $[+.004,+.016]$), a readability effect rather than an interaction. In this design the receiver does not run its own reconstruction; the variant in which it does is reported above (HB\,231). We find that planted rules are recovered at $AUC=.51$ by a 1B reciever (chance), $.61$ at 3B, $.73$ at 8B, and ${\sim}.80$ from 14B, demonstrating a smooth scaling curve. Planted recovery rates from the most capable executors , as AUC against the code-checked truth, run from $.57$--$.84$ across the eight annotation tasks. This is a different quantity from the planted reference of Appendix~\ref{app:osl-regimes}, which is the planted rules' agreement $y$ with the leave-one-executor-out consensus at the three highest-$z$ executors ($.36$ to $.79$) and is the level a saturating curve must fall below to be called bounded.

\subsection{Message length: definitions, the receiver's rules and the search units, against regime and recovery}
\label{app:message-length}
\begin{figure}[h]
\centering
\includegraphics[width=\textwidth]{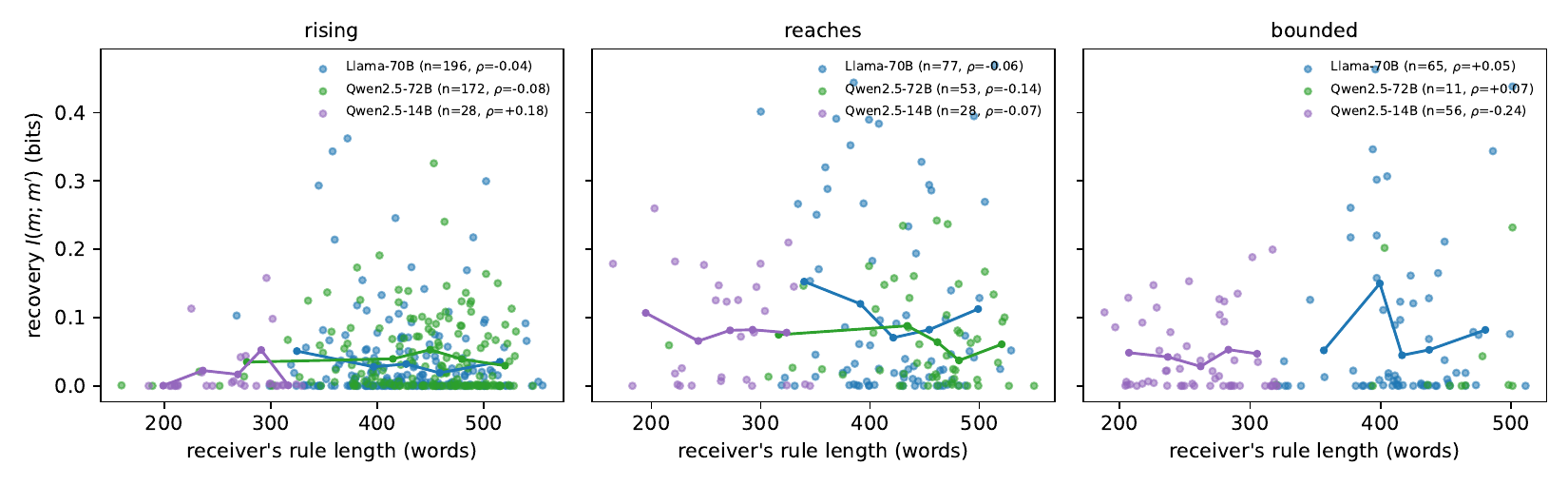}
\caption{\textbf{The receiver's rule length against recovery, by regime.} One point per criterion, three receivers overlaid: Llama-70B on the $130$-criterion set and the bank set ($338$ distinct criteria), Qwen2.5-72B on the bank set, and Qwen2.5-14B as the mean over its four $30$-item rules per criterion. Lines are the mean recovery in each length quintile; $\rho$ is the rank correlation of length with recovery.}
\label{fig:message-length}
\end{figure}

Neither measure in this paper uses message length, and Section~\ref{sec:e_certifiability} improves messages by what they contain rather than by making them longer. This subsection records what length does and does not track, from the message texts the artifacts store (Figure~\ref{fig:message-length}, Table~\ref{tab:message-length}). Four quantities are available: the transmitter's definition of each criterion, the message under which $T$ labels, for all $1{,}270$ fitted metrics; the receiver's reconstructed rule $p'$ with its recovery $I(m;m')$, for the $130$-criterion set and the $236$-criterion bank set at Llama-70B ($338$ distinct criteria; Appendix~\ref{app:examples}), for the bank set at Qwen2.5-72B, and for the $30$-, $60$- and $90$-item run at Qwen2.5-14B (Appendix~\ref{app:more-labels}); and the $360$ units of the five frozen pools of Section~\ref{sec:e_certifiability}, each with its marginal gain at Qwen3-8B. Lengths are in words.

Figure~\ref{fig:message-length} plots the receiver's rule length against recovery, one panel per regime, with the three receivers overlaid. Rule length carries no information about recovery in any regime at any receiver: the rank correlations run from $-.14$ to $+.18$, none with $p<.08$, and the quintile means are flat. The rules are long, $410$ to $460$ words at the two 70B-class receivers and about $265$ at Qwen2.5-14B, and bounded criteria do not draw longer rules than the other regimes. The same holds for the other messages. Definitions are the same length in every regime, a median of $19$ words for rising and reaches criteria and $17$ $[16,19]$ for bounded ones; a unit's length and its marginal gain correlate at $-.04$ over the $360$ units and between $-.08$ and $+.10$ within any one benchmark; and the receiver's rules do not grow with more labeled items, the median falling from $263$ to $256$ words on bounded criteria and from $269$ to $239$ on their controls as the receiver sees $30$, $60$ and $90$ items. What the receiver writes is as long at $30$ items as at $90$, and as long for a concept it recovers as for one it does not.

\begin{table}[h]
\footnotesize
\begin{center}
\resizebox{\textwidth}{!}{%
\begin{tabular}{l l c c c}
\toprule
message & model & $n$ & median length (words) & Spearman with recovery or gain \\
\midrule
transmitter's definition, by regime & -- & $1{,}270$ & bounded $17$, reaches $19$, rising $19$ & -- \\
receiver's rule, rising & Llama-70B / Qwen2.5-72B / Qwen2.5-14B & $196$ / $172$ / $28$ & $424$ / $454$ / $272$ & $-.04$ / $-.08$ / $+.18$ \\
receiver's rule, reaches & Llama-70B / Qwen2.5-72B / Qwen2.5-14B & $77$ / $53$ / $28$ & $430$ / $461$ / $266$ & $-.06$ / $-.14$ / $-.07$ \\
receiver's rule, bounded & Llama-70B / Qwen2.5-72B / Qwen2.5-14B & $65$ / $11$ / $56$ & $413$ / $462$ / $260$ & $+.05$ / $+.07$ / $-.24$ \\
receiver's rule, $30$ / $60$ / $90$ items & Qwen2.5-14B & $112$ & bounded $263$ / $260$ / $256$; controls $269$ / $256$ / $239$ & $-.04$ \\
search unit, five pools & Qwen3-8B & $360$ & $23$ & $-.04$ \\
\bottomrule
\end{tabular}}
\end{center}
\caption{\textbf{Median message length in words and its rank correlation with recovery (rules) or marginal gain (units).} The $30$, $60$ and $90$-item row counts four rules per criterion and size. Artifact \texttt{outputs/analyses/rule\_length\_v1/}.}
\label{tab:message-length}
\end{table}

\subsection{Reconstruction-validity correlations, the arbiter check, and what is validated at which level}
\label{app:mi-silver}

Table~\ref{tab:mi-silver} is the validity table referenced in Section~\ref{sec:recon_results}:
the attenuation-corrected Spearman correlation between a task's reconstruction MI and the AUC of
its judge against silver positive-assertion labels, computed \emph{within} each metric across its prompt forms and aggregated over the annotation task's metrics. Correction for attenuation:
$\rho_{\text{corrected}} = \rho_{\text{observed}} / \sqrt{r_{\mathrm{MI}}\, r_{\mathrm{AUC}}}$,
where $r_{\mathrm{MI}}$ and $r_{\mathrm{AUC}}$ are split-half reliabilities of the two measured
quantities (Spearman--Brown adjusted). The correction estimates how strongly
MI and AUC would correlate if both were measured without noise. Both quantities are estimated
from finite panels, so the observed correlation understates the association; dividing by the
square root of the two reliabilities removes that bias under the classical measurement-error model (independent errors; exact for Pearson, approximate for the rank correlations and selected trajectories used here), at the cost of wider uncertainty
when reliabilities are low. We therefore report the observed values, the reliabilities, and the
panel sizes alongside the corrected values below. Provenance (check of 2026-09-11, artifact \texttt{outputs/analyses/mi\_silver\_reconstruction\_v1/}): every printed row reproduces from the July 18--26 analysis files on the compute box (\texttt{mention\_auc/}), none from a committed file; the findings are incorporated into the text and Table~\ref{tab:mi-silver-v2}.

\begin{table}[h]
\begin{center}
{\scriptsize
\resizebox{\textwidth}{!}{%
\begin{tabular}{lcccccccccc}
\toprule
task & $n$ (pool) & obs.\ $\rho$ & perm.\ $p$ & pool $\rho$ & $r_{\mathrm{MI}}$ & $r_{\mathrm{AUC}}$ & corr.\ unstepped & corr.\ stepped & 95\% CI (stepped) & printed \\
\midrule
peer-review & $21$ ($42$) & $.498$ & $.0005$ & $+.381$ & $.660$ & $.510$ & $.86$ & $.68$ & $[.58,.95]$ & $.68$ (s) \\
code-review & $31$ ($31$) & $.328$ & $.0002$ & --- & $.708$ & $.391$ & $.62$ & $.48$ & $[.18,.71]$ & $.62$ (u) \\
creative-writing & $15$ & $.419$ & $.0002$ & --- & $.809$ & $.360$ & $.78$ & $.61$ & $[.28,1.00]$ & $.61$ (s) \\
math & $14$ ($46$) & $.352$ & $.0005$ & $+.029$ & $.738$ & $.310$ & $.74$ & $.56$ & $[.20,.76]$ & $.56$ (s) \\
press-releases & $23$ ($23$) & $.203$ & $.0052$ & --- & $.789$ & $.212$ & $.50$ & $.37$ & $[-.04,.88]$ & $.50$ (u) \\
humor & $21$ ($106$) & $.344$ & $.0002$ & $+.130$ & $.816$ & $.727$ & $.45$ & $.40$ & $[.26,.54]$ & $.45$ (u) \\
\bottomrule
\end{tabular}}}
\end{center}
\caption{\textbf{Reconstruction validity under both reliability conventions.} $n$ = metrics in the printed row; in parentheses, the full pool size where the row was restricted after the fact to metrics whose judge reaches mean AUC $\ge.55$ against the mention labels; ``pool $\rho$'' is the correlation on the full pool (math: $+.029$, $p=.49$). Peer review and math are also the best configuration of a sweep over label variants ($1$ of $14$ and $1$ of $8$). ``printed'' is the value an earlier version of this table printed, with its convention: (s) stepped, (u) unstepped. Permutation $p$-values shuffle within metric and stay valid after selection; the restricted effect sizes are not comparable to the pools. The 95\% CI resamples metrics $2{,}000$ times (seed $20260912$) and redraws a document or probe split for each reliability on every resample (artifact \texttt{outputs/analyses/mi\_silver\_reliability\_uncertainty\_v1/}); every row had the per-form data needed for $r_{\mathrm{MI}}$, so no sensitivity band was needed.}
\label{tab:mi-silver}
\label{tab:mi-silver-v2}
\end{table}

The bootstrap interval for the stepped correction (Table~\ref{tab:mi-silver-v2}) excludes zero for every row except press-releases, and the unstepped interval for peer-review exceeds $1$ in $42\%$ of resamples, which indicates that the unstepped convention is unstable at peer review's low $r_{\mathrm{MI}}$.

The reliabilities are the $r_{\mathrm{MI}}$ and $r_{\mathrm{AUC}}$ columns of Table~\ref{tab:mi-silver}; peer review's recorded split-half $r_{\mathrm{MI}}$ is $.664$, and its correction used $.66$. These are not estimated uniformly: every $r_{\mathrm{AUC}}$ is a \textit{document} split-half (hash-parity for peer review, math and code review; the mean of $12$ random splits for creative writing), and for peer review, math and code review $r_{\mathrm{MI}}$ is a constant taken from an earlier run. The
reconstruction probe documents and the AUC-scored documents are disjoint by construction
($0/300$ overlap on the checked tasks), so disattenuation is not inflated by shared-document
correlated errors. The rows did not share one convention, and an earlier description of the conventions was incorrect: peer review, math and creative writing divide by the Spearman--Brown \textit{stepped} reliabilities ($.498/\sqrt{.795\times.676}=.68$), code review, press releases and humor by the unstepped ones ($.328/\sqrt{.708\times.391}=.62$). The ``unstepped $.74$'' previously quoted for creative writing came from a different document split ($r_{\mathrm{AUC}}=.391$); the unstepped counterpart of its $.61$ is $.78$. Table~\ref{tab:mi-silver-v2} gives every row under both conventions.

\paragraph{What is validated at which level.} Our claims compare recovery across prompt forms, criteria, executors and scaling regimes, and the validation evidence does not cover these four levels equally. \textit{Across prompt forms, within a criterion}: the level of the table above, and the only one with a direct MI-versus-error correlation; an earlier run with $10$ forms per metric gave $+.39$ on creative writing ($20/22$ metrics positive, permutation $p=2\times10^{-4}$), $+.11$ on math, and no correlation on humor and press releases, where the judges' AUC against the mention labels was at chance ($.50$ and $.48$). \textit{Across criteria}: correlations of MI with mention-derived AUC between metrics are within noise ($+.09$ to $+.17$, $n\approx 50$ per task), so the criterion-level comparisons of Section~\ref{sec:scaling} rest on the planted controls and the arbiter check. \textit{Across executors}: the judges' AUC against the silver labels rises with capability (Spearman $+.60$ over the $12$ executors of the validation panel) and the planted-rule AUC ladders rise within both families (Llama $.52/.77/.95/.97$, Qwen2.5 $.82/.89/.93/.96$); this validates the capability axis, not any criterion's recovery at a given executor. \textit{Across regimes}: no validation evidence; membership is a fitted verdict (Appendix~\ref{app:osl-regimes}). The choice of measure matters little: a pairwise forced-choice measure and the $P(\mathrm{yes})$ measure agree at $r=.82$ over $40$ metrics ($.65$ over $152$).

\paragraph{How the correlations are computed, and the one preregistered window.} Correlations are Spearman, pooled by Fisher-$z$ over each annotation task's metrics. Readouts are
stratified rather than pooled; pooled readouts are composition-diluted and are not reported.
Code review is the one row with a preregistered window, and its corrected value ($.62$) was \textit{above} the window ($[.37,.51]$); only the preregistered contrast between high- and low-positive-count strata held ($+.44$ against $+.20$, full set $+.33$). The table describes metrics the judge can measure at all, not the bank at large. The press-releases observed interval includes zero.

\paragraph{Direct validation against arbiter-filtered labels.} The mention-derived silver
labels used above are partial-observation proxies (a label records that a commenter
\emph{invoked} a criterion, not that it applies, so an unmentioned criterion is not a negative), and a disagreement review found the judge
correct in the majority of judge-vs-label conflicts. We call the relabeled subset \textit{arbiter-filtered}: the arbiter is a stronger judge with more context, not ground truth, and restricting to its confident verdicts and stratifying on judge-score terciles changes the evaluated population. The table validates the \textit{judges} against the arbiter; we test below whether reconstruction MI predicts judge correctness on these items, and that test is null and underpowered. To measure judge validity against a
less noisy criterion, a stratified sample of $(\text{document}, \text{metric})$ pairs per task
(${\sim}270$--$770$ after two waves) was re-labeled by an independent frontier arbiter
(different model family from every scoring judge; full-document context vs.\ the judges'
truncated views; blind to the silver labels and to all judge scores), certified on sealed
mechanically-verifiable anchor items ($93\%$/$94\%$ per wave; wave voided below $80\%$).
Confident arbiter labels (extreme-decile verdicts only) give:

\begin{center}
\resizebox{\textwidth}{!}{%
\begin{tabular}{lccc}
\toprule
task & $n$ & AUC vs.\ arbiter-filtered labels & (vs.\ mention-label floor) \\
\midrule
peer-review & $614$ & $.933$ & $.59$ \\
creative-writing & $163$ & $.815$ & $.58$ \\
press-releases & $226$ & $.686$ & $.55$ \\
code-review & $163$ & $.668$ & $.59$ \\
humor & $272$ & $.631$ & $.52$ \\
\midrule
mean (unweighted / $n$-weighted) & $1{,}438$ & $.747$ / $.794$ & \\
\bottomrule
\end{tabular}}
\end{center}

Three qualifications apply to this table. (1) The sample is stratified (balanced across label
value and judge-score terciles), so these are balanced-sample AUCs, not corpus-weighted
ones. (2) Arbiter-filtered labels are \emph{specification-relative}: re-arbitrating the same items
under a changed concept specification (criterion name alone vs.\ name with definition)
flipped $46\%$ of verdicts, so the labels inherit the concept's written form. This is an
instance of the wording-sensitivity this paper studies, here on the label side. Prompts that
share the specification text are therefore structurally advantaged on this criterion, and no
prompt-optimization comparison is scored against it without a specification-independence
control. (3) The mention-label floors in the last column confirm that label noise, not judge
capability, bound the silver-label correlations above: every task's judge exceeds its
mention floor by $.08$--$.34$ AUC once labels are arbiter-filtered.

\textit{Whether MI predicts correctness on these labels.} A direct test is null and underpowered (artifact \texttt{outputs/analyses/mi\_vs\_arbiter\_correctness\_v1/}; the rebuilt labels reproduce the counts and AUCs above). Across metrics, the correlation between a metric's reconstruction MI and its judge's AUC against the arbiter-filtered labels is $+.90$ on creative writing ($n=5$), $-.79$ on press releases ($n=8$) and $+.33$ on humor ($n=8$), pooled $+.25$ (permutation $p=.40$); within metric across prompt forms it runs $-.08$ to $+.13$, all $p>.12$; controlling for $H(m)$, the base rate or the mention-label AUC does not change these results. The same test on the mention labels gives $+.09$ over $47$ peer-review metrics, so the effect being sought is about $.10$, smaller than $5$--$11$ metrics per task can detect. Two facts qualify the table: in peer review $24$ of $30$ metrics have confident verdicts of one class only, so no per-metric AUC exists and the $.933$ is a between-metric contrast (centred within metric it is $.60$); and peer review's prompt forms were scored on different documents, so its across-form comparison is not paired. Code review and math contribute no data to this test (no form-level MI; no confident verdicts).

\section{External validation of the capability scalar $z$}
\label{app:z-validation}
The executor-capability axis used throughout the scaling analyses is a \emph{declared
instrument}: $z_E=\mathrm{logit}(\mathrm{AUC})$ on a frozen 240-item anchor battery of
code-checkable criterion judgments, scored for every executor under the identical serving stack
as the recovery measurements. Because a declared axis can fail idiosyncratically, we
cross-validated it against an external instrument that we did not build: the OpenCompass
(CompassRank) LLM leaderboard, which evaluates models on \emph{closed}, contamination-resistant
evaluation sets refreshed each edition and reports per-dimension scores (language, knowledge,
reasoning, math, code, agent).

Six of our fourteen local executors match OpenCompass entries at the exact checkpoint level
(Llama-3.2-3B, Llama-3.1-8B, Qwen2.5-7B, Qwen2.5-72B, Gemma-2-9B, Gemma-2-27B; all
instruction-tuned variants), and all six appear in three consecutive leaderboard editions. Within each
edition, our battery-$z$ correlates with the OpenCompass average at $r\approx.64$--$.67$
(Spearman $\rho\approx.54$), and the disagreement is due
\emph{entirely} to Gemma-2-27B. With that executor excluded, agreement rises to
$\mathbf{r=+.87}$, $\boldsymbol{\rho=+.90}$ ($n=5$).

The Gemma-2-27B case is a diagnosable instrument--executor incompatibility rather than a
scoring dispute. Our battery scores it \emph{below} Gemma-2-9B ($z$ $1.10$ vs.\ $2.04$), while
OpenCompass ranks it above 9B in every edition. The battery's per-family breakdown localizes
the cause: Gemma-2-27B performs at chance ($\mathrm{AUC}=.487$) on the
\emph{composite} (multi-clause) battery items, where Gemma-2-9B scores $.833$, while remaining
functional on negation ($1.0$), presence ($.867$), paraphrase and threshold items
($\approx.71$); its answer distribution is strongly bimodal (3\% of $P(\mathrm{YES})$ mass
in $[.4,.6]$), with zero parse failures and stable split-halves. That is, the model answers
decisively but at chance on multi-clause forced-choice judgments. This is a format-specific failure,
consistent with its form-fragility on other measures in our pipeline and not visible in
generative multi-format harnesses. For this reason the Gemma-2 ladder is not used as a same-family ladder for forecasting (a decision made on the inversion before this external check); its two executors remain in the pooled panel on which the regime curves are fitted, where family enters only through the consensus (Appendix~\ref{app:regime-robustness}). The leaderboard
comparison validates the battery on the remaining panel and independently confirms that
exclusion.

This check has three limitations: the exact-match panel is small ($n=6$); OpenCompass closed-set
scores are \emph{not comparable across editions} (the same frozen checkpoint's score drops as
the eval set is re-hardened, e.g.\ Qwen2.5-72B $70.3\to57.3\to49.9$ across three editions),
so all correlations are computed within-edition only; and the leaderboard's frontier entries
are predominantly reasoning-mode variants while our probes run with reasoning disabled, so the
check covers the local ladder, not the frontier points.
\section{Observational scaling by regime: definitions, fits, type structure and robustness}
\label{app:osl-regimes}
\begin{figure}[t]
	\centering
	\resizebox{\textwidth}{!}{%
		\begin{tikzpicture}[
			lab/.style={font=\scriptsize\itshape, align=center},
			tlab/.style={font=\tiny\itshape, align=center},
			arr/.style={-{Stealth[length=1.8mm]}, thick},
			]
			\begin{scope}
				\draw[arr] (0,0) -- (0,3.2); \draw[arr] (0,0) -- (4.3,0);
				\node[tlab, rotate=90] at (-0.82,1.55) {recovery};
				\node[tlab] at (2.1,-0.42) {model scale (Qwen3 family)};
				\IfFileExists{figs/gen_scaling_supervised.tex}{\draw (-0.06,0.12) -- (0.05,0.12); \node[tlab, anchor=east] at (-0.09,0.12) {0.21};
\draw (-0.06,0.81) -- (0.05,0.81); \node[tlab, anchor=east] at (-0.09,0.81) {0.32};
\draw (-0.06,1.49) -- (0.05,1.49); \node[tlab, anchor=east] at (-0.09,1.49) {0.43};
\draw (-0.06,2.18) -- (0.05,2.18); \node[tlab, anchor=east] at (-0.09,2.18) {0.54};
\draw (-0.06,2.86) -- (0.05,2.86); \node[tlab, anchor=east] at (-0.09,2.86) {0.66};
\draw (0.33,0.06) -- (0.33,0.18); \node[tlab] at (0.33,-0.22) {0.6B};
\draw (1.65,0.06) -- (1.65,0.18); \node[tlab] at (1.65,-0.22) {1.7B};
\draw (2.74,0.06) -- (2.74,0.18); \node[tlab] at (2.74,-0.22) {4B};
\draw (3.62,0.06) -- (3.62,0.18); \node[tlab] at (3.62,-0.22) {8B};
\draw[blue!60!black, very thick] plot coordinates {(0.33,0.66) (1.65,1.59) (2.74,2.39) (3.62,2.44)};
\fill[blue!60!black] (0.33,0.66) circle (1.1pt);
\fill[blue!60!black] (1.65,1.59) circle (1.1pt);
\fill[blue!60!black] (2.74,2.39) circle (1.1pt);
\fill[blue!60!black] (3.62,2.44) circle (1.1pt);
\draw[blue!60!black, densely dashed, thin] plot coordinates {(0.33,0.43) (1.65,1.04) (2.74,1.40) (3.62,1.30)};
\draw[orange!80!black, very thick] plot coordinates {(0.33,1.01) (1.65,1.66) (2.74,2.11) (3.62,2.20)};
\fill[orange!80!black] (0.33,1.01) circle (1.1pt);
\fill[orange!80!black] (1.65,1.66) circle (1.1pt);
\fill[orange!80!black] (2.74,2.11) circle (1.1pt);
\fill[orange!80!black] (3.62,2.20) circle (1.1pt);
\draw[orange!80!black, densely dashed, thin] plot coordinates {(0.33,0.81) (1.65,1.38) (2.74,1.71) (3.62,1.77)};
\node[tlab, anchor=north west, align=left, draw=black!25, fill=white, inner sep=2pt] at (0.22,2.79) {\textcolor{blue!60!black}{\rule[0.5ex]{3mm}{1pt}} hotpot \ \textcolor{orange!80!black}{\rule[0.5ex]{3mm}{1pt}} hover \\ solid: random-draw mean \ dashed: GEPA-shipped};}{%
					\node[tlab, red!60!black] at (2.1,1.6) {[OSL battery in flight]};}
				\node[lab] at (2.1,3.45) {supervised benchmarks};
			\end{scope}
			\begin{scope}[shift={(5.6,0)}]
				\draw[arr] (0,0) -- (0,3.2); \draw[arr] (0,0) -- (4.3,0);
				\node[tlab, rotate=90] at (-0.82,1.55) {recovery};
				\node[tlab] at (2.1,-0.42) {capability index $z$ (logit AUC, anchor battery)};
				\IfFileExists{figs/gen_scaling_unsupervised.tex}{\input{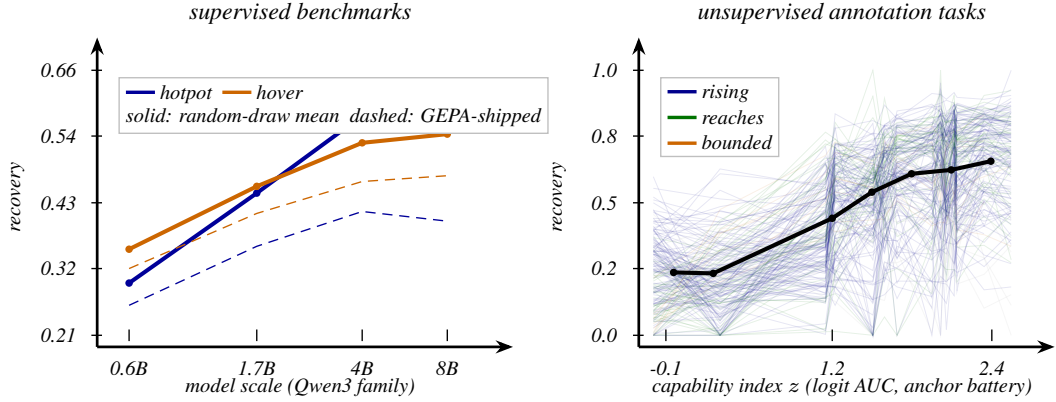}}{}
				\node[lab] at (2.1,3.45) {unsupervised annotation tasks};
			\end{scope}
	\end{tikzpicture}}%
	\caption{\textbf{Articulation scaling} in the observational-scaling presentation: existing models as
 scale points, plotted against a measured capability index rather than parameter count.
 \emph{Right, unsupervised (measured):} $1{,}347$ tasks evaluated by $14$ executors spanning
 $1.2$--$73$B across five model families; each light trace is one task's recovery curve,
 colored by its fitted verdict, with the binned mean and its $95\%$ interval in black.
 \emph{Left, supervised (measured):} the complete Qwen3 ladder ($4$ rungs $\times$ both benchmarks). Solid lines are the mean of $40$ random unit-pool draws per rung, dashed lines the GEPA-shipped prompt. At $8$B the draws beat the shipped prompt in $40/40$ cases on \emph{both} benchmarks (HotpotQA $+.186$, HoVer $+.070$), and the gain grows with scale.}
	\label{fig:scaling-envelopes}
\end{figure}

This appendix documents the fits behind the regime split shown in the main body
(Figure~\ref{fig:osl-regimes}) and gives the underlying curves in the observational-scaling
presentation (Figure~\ref{fig:scaling-envelopes}, this appendix): existing models as scale
points against the measured capability index $z$, with the supervised Qwen3 ladder alongside.
The regime definitions below are transcribed from the fitting code. \textit{Vertical axis.} For executor $e$ and metric $n$, the executor's labels under the metric's prompt (its definition, rendered in four forms) are $z$-scored over the $300$ probe items; the reference is the consensus of the other executors with the executor itself left out: each other executor's $z$-scored labeling is averaged within its family and the family means are then averaged, so no family dominates; $r$ is the Spearman correlation between the two; $\mathrm{rel}$ is the Spearman--Brown split-half reliability of the executor's form panel (even against odd forms; $\mathrm{rel}\le.2$ dropped); the plotted value is $y=\min(r/\sqrt{\mathrm{rel}},\,1.15)$, with a standard error from even- against odd-numbered probes. This quantity is agreement with the leave-one-executor-out consensus of the other executors, corrected for reliability; it is not mutual information. \textit{Fit.} A curve needs at least $8$ executors. A logistic $y=L\,\sigma(k(z-z_0))$ with $L\le 1.2$ and a line are fitted by least squares; the curve is \textit{saturating} when the logistic wins on AICc. $L$ has a profile-likelihood interval $[L_{lo},L_{hi}]$. \textit{Planted reference.} Per task, the planted reference is the mean over the five planted rules of the rule's mean $y$ at the three highest-$z$ executors (chosen by $z$, never by $y$). The five rules are the same in every task and are checked by code against the text: ``This text is longer than $k$ words'' and ``This text is shorter than $k$ words'', where $k$ is the task's median word count; ``This text contains at least one question mark''; ``This text contains at least one digit (0--9)''; and ``This text contains quoted speech (a quotation mark)''. The reference by task is creative writing $.77$, humor $.79$, math $.62$, news homepages $.65$, notice-and-comment $.58$, patents $.36$, peer review $.67$ and press releases $.63$. \textit{Decision order.} \emph{Bounded} if saturating and $L_{hi}$ is below the planted reference; else \emph{reaches} if saturating; else \emph{rising} if the Spearman slope of $y$ in $z$ is at least $.35$; else \emph{noisy}. The classes cannot overlap, and ``bounded'' means below the planted reference, not below $1$. Verdicts are fitted on the pooled $14$-executor panel, not family by family. $77$ curves with a
noisy or undetermined fit are excluded from the panels and from any regime-level claim (the $40$ planted rules were never fitted; Appendix~\ref{app:regime-robustness}). The
supervised Qwen3 ladder is shown for context and is not re-fit here. The uncertainty of the fits, the held-out test of the strongest executor, the exclusions and the type contrasts are examined in Appendix~\ref{app:regime-robustness} below.

\paragraph{Bounded ceilings are attained and tight; rising ceilings are extrapolations.} The fitted ceilings are ordered across regimes, but they
are not equally trustworthy: \emph{bounded} ceilings are attained within range with the tightest
intervals (median CI width $.105$); \emph{reaches} ceilings are attained ($.33$); \emph{rising}
ceilings are extrapolations (median CI width $.56$). Of the rising fits, $99\%$ place their
upper bound above $1$, i.e.\ the ceiling is unidentified and ``rising'' chiefly records
censoring by our strongest executor. Realized recovery at the top of the executor range gives
a clearer comparison: reaches $.75$ (finished), rising $.68$ (still climbing), bounded $.52$
(finished, low).

\begin{figure}[h]
\centering
\resizebox{.75\textwidth}{!}{%
\begin{tikzpicture}[tlab/.style={font=\tiny\itshape, align=center}]
\IfFileExists{figs/gen_osl_types.tex}{\input{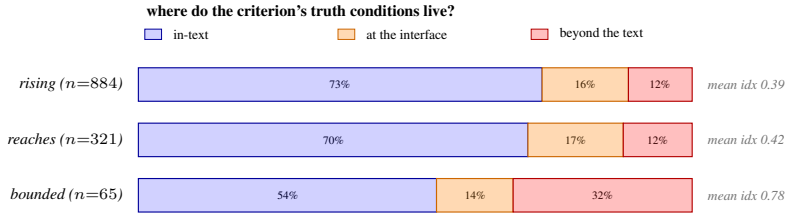}}{}
\end{tikzpicture}}%
\caption{\textbf{The three regimes on one axis}: where each criterion's truth
conditions live (shared task-agnostic taxonomy, coarsened; regime-blinded labeling; two
independent passes, Appendix~\ref{app:taxonomy}). Rising and
reaches are near-identical; bounded shifts sharply beyond the text; the numbers are in the text. The regime-by-task composition is
Figure~\ref{fig:osl-domains} in the main body.}
\label{fig:osl-types}
\end{figure}

\paragraph{A task-agnostic type structure.} To compare the regimes on a common basis, we labeled all
$1{,}270$ criteria with a shared, task-agnostic taxonomy of what a criterion \emph{asks about}
(regime held out of the labeling input): nine types, defined in Appendix~\ref{app:taxonomy} and coarsened onto one axis, \emph{where the criterion's truth conditions live}: in the text, at the interface or beyond the text. Figure~\ref{fig:osl-types} shows the result. Rising and reaches are nearly
identical on this axis (in-text $73\%$/$70\%$, beyond-text $12\%$ both; mean externality index
$.39$/$.42$); what separates \emph{them} is domain and device compactness rather than externality.
Bounded criteria differ on this axis: $32\%$ beyond-text and mean index $.78$, driven by identity \&
persona ($17\%$ vs.\ ${\sim}2.5\%$ elsewhere), community transformation (monotone
$0.8\%\to4.7\%\to9.2\%$ across regimes), and reception outcomes (monotone
$5.3\%\to7.5\%\to10.8\%$), while evidence \& rigor and audience fit both \emph{fall} toward
bounded. The gradient holds under domain control: within humor, news, and peer review separately,
bounded criteria are $31$--$34\%$ beyond-text against $4$--$18\%$ for the other regimes. It also replicates in an independent second labeling pass (mean externality index $.71$ against $.38$/$.37$).

\subsection{Robustness of the regime findings: curve form, executors, exclusions, domains, contrasts and capability axis}
\label{app:regime-robustness}
The regime split rests on four choices, the ones named in Section~\ref{sec:scaling}: the \emph{curve form}, the \emph{executor set}, the \emph{exclusions} and the \emph{type contrasts}. This section defines each, then checks each against the shipped panels (artifacts \texttt{outputs/analyses/regime\_uncertainty\_v1/}, \texttt{heldout\_baselines\_v1/}, \texttt{nuisance\_regression\_v1/} and \texttt{osl\_pca\_v1/}). Two kinds of measurement noise run through every check, and they are defined here as well.

\emph{Curve form.} The function fitted to a metric's points $(z_e, y_e)$, one per executor: a straight line $y=a+bz$ or a logistic $y=L\,\sigma(k(z-z_0))$ with ceiling $L\le 1.2$, chosen by AICc, the logistic's ceiling carrying a profile-likelihood interval $[L_{lo},L_{hi}]$. The form decides whether a curve is saturating (the logistic wins) and, through $L_{hi}$ against the planted reference, whether a saturating curve is bounded or reaches; a non-saturating curve is rising when the Spearman slope of $y$ in $z$ is at least $.35$ and noisy otherwise (Appendix~\ref{app:osl-regimes}).

\emph{Executor set.} The $14$ instruction-tuned open models of Table~\ref{tab:regime-executors}, $1.2$B to $73$B in five families, each scoring every metric under the same prompt and serving stack. Every curve is fitted on the pooled set, never family by family: a median of $14$ points, with $950$ of the $1{,}270$ metrics having all $14$. A fifteenth model, Qwen3.5-122B, enters only the leave-one-executor-out consensus on the vertical axis and is never a curve point.

\emph{Exclusions.} Of the $1{,}347$ bank curves, $77$ with neither a preferred logistic nor a slope of at least $.35$ are set aside as noisy, which leaves the $1{,}270$ fitted metrics; the $40$ planted rules (five per task) are never fitted and set the planted reference instead.

\emph{Type contrasts.} Five statements of the form ``criteria of type $T$ are over-represented in regime $R$'': reaches and local patterns, rising and whole-text structure, rising and verifiable features, bounded and identity and social group, bounded and beyond-text features. The types are the six-type codebook of Appendix~\ref{app:codebook-six}, assigned without sight of the regime. Each contrast is the share of type $T$ among the criteria of regime $R$ against its share among the other two regimes, tested against a null that permutes regime labels within domain ($10{,}000$ draws), with Benjamini--Hochberg control over the $18$ regime-by-type cells.

\emph{Measurement noise in recovery.} Each executor's recovery $y_e$ has a standard error from even- against odd-numbered probes ($.025$ at the median). The check redraws every point from its standard error, refits the curve and re-applies the decision rule ($2{,}000$ draws); contrasts are then weighted by each metric's share of draws in each regime.

\emph{Measurement noise in the capability index.} Each executor's $z_e$ has a split-half standard error from the battery ($.07$ for most executors, up to $.24$ for Llama-70B). The check redraws every $z_e$, recomputes the planted reference and refits ($100$ draws).

\emph{Regime sizes and domain composition.} The counts $884$, $321$ and $65$, and the number of bounded criteria in each domain (Figure~\ref{fig:osl-domains}). A quantity survives a check when its value under the check stays within the range the tables below report and the same contrasts remain significant.

\begin{table}[h]
\small
\begin{center}
\begin{tabular}{l l r c}
\toprule
executor & family & parameters & capability index $z$ \\
\midrule
Llama-3.2-1B & Llama & $1.2$B & $-.24$ \\
Mistral-7B & Mistral & $7.2$B & $.28$ \\
Gemma-2-27B & Gemma-2 & $27.2$B & $1.10$ \\
Llama-3.2-3B & Llama & $3.2$B & $1.17$ \\
Qwen2.5-3B & Qwen2.5 & $3.1$B & $1.47$ \\
Llama-3.1-8B & Llama & $8.0$B & $1.56$ \\
Qwen2.5-7B & Qwen2.5 & $7.6$B & $1.66$ \\
Phi-4 & Phi & $14.7$B & $1.93$ \\
Qwen2.5-14B & Qwen2.5 & $14.8$B & $2.00$ \\
Gemma-2-9B & Gemma-2 & $9.2$B & $2.04$ \\
Mistral-Small-24B & Mistral & $23.6$B & $2.10$ \\
Llama-3.3-70B & Llama & $70.6$B & $2.13$ \\
Qwen2.5-72B & Qwen2.5 & $72.7$B & $2.34$ \\
Qwen2.5-32B & Qwen2.5 & $32.8$B & $2.55$ \\
\midrule
Qwen3.5-122B (consensus only) & Qwen3.5 & $122$B & $2.23$ \\
\bottomrule
\end{tabular}
\end{center}
\caption{\textbf{The executor set.} The $14$ instruction-tuned open models on which every regime curve is fitted, ordered by the capability index $z$ (logit AUC on the $240$-item battery, Appendix~\ref{app:z-validation}). Qwen3.5-122B, a mixture of experts with $10$B active parameters, contributes only to the leave-one-executor-out consensus on the vertical axis and is not a curve point.}
\label{tab:regime-executors}
\end{table}

\paragraph{Curve form: saturation takes precedence, and rising is what is left when the line wins.} A metric is \emph{saturating} when the logistic beats the straight line on AICc; $386$ of the $1{,}270$ curves are. The two tests overlap: $375$ of the $386$ saturating curves also have a Spearman slope of at least $.35$, and the decision order gives saturation precedence, so \emph{rising} is the class left over when the line is preferred rather than a positive test that a curve keeps climbing. The ceilings behave as described under ``Interpreting the fits'' above: bounded ceilings are attained with the tightest intervals, rising ceilings are extrapolations. To test whether the chosen curve has predictive value, we refit every metric without its highest-$z$ executor and predicted that executor (Table~\ref{tab:regime-heldout}). The fitted form's median error is $9\%$ of the held-out value, and it over-predicts on $72\%$ of metrics. The AICc choice coincides with the logistic on $77\%$ of metrics, and on the $386$ saturating curves it beats the line by a wide margin ($.041$ against $.111$). Fitted within one family instead of on the pooled panel, the error is $.096$ (Llama, four executors) and $.067$ (Qwen2.5, five), the latter no better than that family's own train mean ($.070$); at five points or fewer the AICc comparison is undefined, so no within-family verdict exists. The rising cutoff on the rank correlation, $.35$, has no external criterion to validate it against; changing it only moves curves between rising and noisy. No fitted curve is below it, $23$ of the $77$ excluded curves would join rising at $.25$, and raising it to $.46$ (the value a curve with no trend exceeds one time in twenty at $14$ executors) reclassifies $44$ rising curves as noisy, $.55$ reclassifies $122$, $.65$ reclassifies $226$; the rising/whole-text contrast remains significant at every cutoff from $.25$ to $.65$ ($q\le.014$) and rising/verifiable at none.

\begin{table}[h]
\small
\begin{center}
\begin{tabular}{l c c c}
\toprule
predictor of the held-out largest executor & median $|$error$|$ & median error / SE & within $2$ SE \\
\midrule
fitted form (AICc choice, as in the paper) & $.062$ & $2.4$ & $42\%$ \\
logistic always & $.051$ & $2.1$ & $49\%$ \\
straight line always & $.075$ & $3.0$ & $35\%$ \\
mean of the remaining executors & $.149$ & $5.7$ & $18\%$ \\
\bottomrule
\end{tabular}
\end{center}
\caption{\textbf{Held-out prediction of the strongest executor}, $1{,}270$ metrics. Each curve is refitted without its highest-$z$ executor and that executor's recovery is predicted; SE is the held-out point's own standard error ($.025$ at the median). Artifact \texttt{outputs/analyses/heldout\_baselines\_v1/}.}
\label{tab:regime-heldout}
\end{table}

\paragraph{Executor perturbations: a parametric bootstrap, and dropping each curve's most capable executor.} Every curve is fitted on the pooled panel: a median of $14$ executors ($950$ of the $1{,}270$ metrics have all $14$) from five families (Llama $4$, Qwen2.5 $5$, Gemma-2 $2$, Mistral $2$, Phi $1$); family enters only through the leave-one-executor-out, family-balanced consensus on the vertical axis. Two perturbations test the executor set. A parametric bootstrap redraws each executor's $y$ from its standard error, refits and re-applies the rule ($2{,}000$ draws). Group sizes hold: rising $849$--$890$ (printed $884$), reaches $287$--$326$ ($321$), bounded $54$--$74$ ($65$); a typical draw also sends about $30$ metrics ($21$--$39$) to noisy. Memberships hold less well (Table~\ref{tab:regime-membership}): the modal verdict equals the printed one for $94\%$ of metrics, but $397$ metrics have a modal probability below $.8$ and $163$ below $.6$, and a bounded criterion keeps its label in $65\%$ of draws on average against $90\%$ for a rising one. Dropping the highest-$z$ executor changes $12\%$ of verdicts, with reaches the most exposed: $106$ of the $321$ reaches curves become rising without their top executor, because that point is what made the curve bend. The group sizes then become $971$ rising, $224$ reaches and $62$ bounded, with $13$ noisy, so the bounded count holds and the split between rising and reaches does not.

\begin{table}[h]
\small
\begin{center}
\resizebox{\textwidth}{!}{%
\begin{tabular}{l c c c c c c c l}
\toprule
 & & \multicolumn{3}{c}{noise in recovery ($2{,}000$ draws)} & \multicolumn{3}{c}{noise in the capability index ($100$ draws)} & \\
\cmidrule(lr){3-5}\cmidrule(lr){6-8}
printed regime & $n$ & mean $P$(own) & modal $=$ printed & $P$(own) $<.6$ & mean $P$(own) & modal $=$ printed & $P$(own) $<.6$ & verdict changes without the top executor \\
\midrule
rising & $884$ & $.90$ & $99\%$ & $5\%$ & $.91$ & $98\%$ & $7\%$ & $3\%$ ($10$ to reaches, $4$ to bounded, $13$ to noisy) \\
reaches & $321$ & $.73$ & $84\%$ & $34\%$ & $.82$ & $95\%$ & $13\%$ & $36\%$ ($106$ to rising, $8$ to bounded) \\
bounded & $65$ & $.65$ & $78\%$ & $42\%$ & $.82$ & $92\%$ & $12\%$ & $23\%$ ($7$ to reaches, $8$ to rising) \\
\bottomrule
\end{tabular}}
\end{center}
\caption{\textbf{Membership uncertainty by regime.} $P$(own) is the share of draws in which a metric keeps its printed verdict. Columns 3--5: each executor's recovery redrawn from its standard error ($2{,}000$ draws). Columns 6--8: each executor's capability index redrawn from its split-half standard error, with the planted reference recomputed ($100$ draws). Last column: refit without the highest-$z$ executor. Artifacts \texttt{outputs/analyses/regime\_uncertainty\_v1/} and \texttt{osl\_pca\_v1/}.}
\label{tab:regime-membership}
\end{table}

\paragraph{Which curves are excluded as noisy, and the planted rules that are never fitted.} The bank has $1{,}347$ curves; $77$ with neither a preferred saturating fit nor a slope of at least $.35$ are excluded as noisy, which leaves the $1{,}270$ fitted metrics. The $40$ planted rules are never fitted; they set the planted reference (Appendix~\ref{app:osl-regimes}). An earlier count of ``$117$ excluded'' added the two groups together.

\paragraph{Regime composition by domain: bounded shares differ by domain, rising shares are large everywhere.} Figure~\ref{fig:osl-domains} (main body)
shows each domain's regime composition directly. The bounded share is domain-specific: humor
$15\%$ ($41$ tasks), peer review $11\%$ ($9$), news homepages $10\%$ ($13$), creative writing
$1\%$ ($2$), and zero in math, notice-and-comment, patents, and press releases. The
rising share is $65$--$86\%$ in every domain, so every domain has headroom. (Labels
are names-based; a near-duplicate pair of sourcing-transparency criteria that fall in different
regimes within news shows that names alone do not fully determine membership.) The composition was checked under the same perturbations (Table~\ref{tab:regime-domains}). Under both kinds of measurement noise and without the most capable executor, bounded criteria stay concentrated in humor, news homepages, peer review and, marginally, creative writing. The four domains printed with no bounded criteria are not always empty under noise: in the recovery-noise bootstrap, math has at least one bounded criterion in $73\%$ of draws, notice-and-comment in $64\%$ and press releases in $34\%$, never more than three; patents never has one. Under a differently built capability axis the composition does not hold: on the first component of the battery's sub-scores humor keeps $3$ bounded criteria, and on equally spaced ranks every domain but patents has bounded criteria.

\begin{table}[h]
\small
\begin{center}
\resizebox{\textwidth}{!}{%
\begin{tabular}{l c c c c c c c}
\toprule
 & & \multicolumn{6}{c}{number of bounded criteria} \\
\cmidrule(lr){3-8}
domain & criteria & printed & noise in recovery, $95\%$ of draws & noise in the index, expected & without the top executor & battery's first component & equally spaced ranks \\
\midrule
humor & $278$ & $41$ & $30$--$46$ & $44.0$ & $41$ & $3$ & $129$ \\
news homepages & $132$ & $13$ & $6$--$15$ & $11.0$ & $10$ & $4$ & $28$ \\
peer review & $84$ & $9$ & $6$--$12$ & $8.6$ & $9$ & $3$ & $29$ \\
creative writing & $367$ & $2$ & $2$--$8$ & $2.4$ & $2$ & $2$ & $67$ \\
math & $137$ & $0$ & $0$--$3$ & $0.4$ & $0$ & $0$ & $9$ \\
notice and comment & $66$ & $0$ & $0$--$3$ & $0.3$ & $0$ & $0$ & $6$ \\
press releases & $199$ & $0$ & $0$--$2$ & $0.1$ & $0$ & $0$ & $5$ \\
patents & $7$ & $0$ & $0$ & $0.0$ & $0$ & $0$ & $0$ \\
\bottomrule
\end{tabular}}
\end{center}
\caption{\textbf{Bounded criteria by domain under each check.} Noise in recovery: each executor's recovery redrawn from its standard error ($2{,}000$ draws). Noise in the index: each executor's capability index redrawn from its split-half standard error, expected count over $100$ draws. Without the top executor: each curve refit without its highest-$z$ executor. The last two columns refit every curve on a differently built capability axis (Table~\ref{tab:regime-axes}). Artifacts \texttt{outputs/analyses/regime\_uncertainty\_v1/} and \texttt{osl\_pca\_v1/}.}
\label{tab:regime-domains}
\end{table}

\paragraph{The five regime-by-type contrasts and their permutation test.} The main text's five regime-by-type contrasts each ask whether one kind of criterion is over-represented in one regime relative to the other two: reaches/local patterns, rising/whole-text structure, rising/verifiable features, bounded/identity and social group, and bounded/beyond-text features. They were tested by permuting regime labels within domain ($10{,}000$ draws) with Benjamini--Hochberg control; the regime-by-type table as a whole departs from independence ($\chi^2=90.0$ on $10$ degrees of freedom, Cram\'er's $V=.19$). Three contrasts remain significant and two do not (Table~\ref{tab:regime-contrasts}). Pooled across domains, ``rising is verifiable'' had $q=.013$; within domain it has $q=.31$, so that contrast was due to domain composition. Weighting each metric by its membership probabilities under recovery noise instead of its printed verdict gives the same three significant contrasts ($q=.0004$, $.0004$ and $.011$), with the bounded identity-and-social share falling from $23\%$ to $19\%$; so does weighting by the probabilities under capability-index noise ($q=.0004$, $.0004$ and $.008$). On the verdicts refit without each curve's most capable executor only reaches/local patterns remains significant ($34\%$ against $18\%$, $q=.001$); rising/whole-text structure ($27\%$ against $20\%$, $q=.09$) and bounded/identity and social group ($19\%$ against $6\%$, $q=.19$) keep their direction but lose significance. The same three also hold in a regression that adds, to domain effects and the type indicator, the label entropy $H(m)$ of the executor's per-item judgment ($10$ bins), the criterion's split-half reliability, document length and the share of items cut by the $4{,}000$-character truncation ($0\%$ in humor, peer review and notice-and-comment, $78\%$ in patents); length and truncation are properties of a domain's shared $300$-item corpus, so they enter pooled across the eight domains with a domain-clustered bootstrap. Coefficients move by at most about $16\%$ (Table~\ref{tab:regime-contrasts}; the rising/whole-text coefficient is $.43$ $[.22,.87]$ with length and truncation added). Of the added covariates, $H(m)$ has the largest effect: with reliability it raises the recovery regression's $R^2$ from $.35$ to $.74$, and the apparent recovery shortfall of beyond-text criteria at the top of the executor range ($-.053$ $[-.082,-.025]$) disappears once it enters ($.007$ $[-.011,.025]$); whole-text structure shows no shortfall before or after ($.0035$ $[-.020,.013]$; $.0044$ $[-.006,.015]$), so its ``harder'' reading follows from its classification as rising. Finally, the type signatures do not depend on the grain of the codebook: the rising/reaches inversion on whole-text against local patterns and the bounded social signature appear at $12$, $8$ and $6$ categories (Figure~\ref{fig:osl-granularity}), and the shift of bounded criteria beyond the text holds in the second labeling pass and within domain (Appendix~\ref{app:osl-regimes}).

\begin{table}[h]
\small
\begin{center}
\resizebox{\textwidth}{!}{%
\begin{tabular}{l c c c c c c c}
\toprule
contrast (regime / type) & share in regime vs.\ other regimes & permutation $q$ & $q$, recovery noise & $q$, capability-index noise & $q$, without the top executor & log-odds, domain effects only & $+H(m)$ and reliability \\
\midrule
reaches / local patterns & $34\%$ vs.\ $17\%$ & $.001$ & $.0004$ & $.0004$ & $.001$ & $.92$ $[.62,1.22]$ & $.77$ $[.46,1.09]$ \\
rising / whole-text structure & $28\%$ vs.\ $19\%$ & $.009$ & $.0004$ & $.0004$ & $.09$ & $.46$ $[.17,.78]$ & $.45$ $[.14,.76]$ \\
bounded / identity and social group & $23\%$ vs.\ $6\%$ & $.03$ & $.011$ & $.008$ & $.19$ & $.92$ $[.23,1.57]$ & $.80$ $[.08,1.46]$ \\
rising / verifiable features & $29\%$ vs.\ $22\%$ & $.31$ & $.24$ & $.13$ & $.50$ & $.18$ & $.16$ \\
bounded / beyond-text features & $15\%$ vs.\ $10\%$ & $.42$ & $.40$ & $.41$ & $.44$ & $.27$ & $.12$ \\
\bottomrule
\end{tabular}}
\end{center}
\caption{\textbf{The five regime-by-type contrasts under the checks.} $q$ from within-domain permutation with false-discovery control, on the printed verdicts, weighted by membership probabilities under recovery noise, weighted by membership probabilities under capability-index noise, and on the verdicts refit without each curve's most capable executor. Log-odds with $95\%$ domain-clustered bootstrap intervals from the nuisance regression (intervals omitted for the two contrasts that are null in every model). Artifacts \texttt{outputs/analyses/nuisance\_regression\_v1/}, \texttt{regime\_uncertainty\_v1/} and \texttt{osl\_pca\_v1/}.}
\label{tab:regime-contrasts}
\end{table}

\paragraph{One axis or several, and which one.} Two further checks follow \citet{ruan2024observational} (artifact \texttt{outputs/analyses/osl\_pca\_v1/}). First, we test whether recovery is low-rank across executors. On the $950$ metrics scored by all $14$ executors, with each metric centered, the first principal component of the executor-by-metric matrix explains $80\%$ of the variance (a permutation null gives $9\%$) and its executor scores correlate with $z$ at $r=.95$. The second explains $5\%$ and is the shape axis the regimes are built on: saturating criteria have positive weights on it (bounded $+.022$ $[+.013,+.030]$, reaches $+.016$ $[+.012,+.020]$) and rising ones negative weights ($-.008$ $[-.010,-.005]$), and no criterion type stands apart from its regime on it. A rank-one fit leaves a residual of $.082$ against a median measurement error of $.027$; the remaining fifth is real but spread over many small components, not a second capability dimension. Second, we test how much the verdicts depend on the axis. Under $z$'s own measurement error (each executor's index redrawn from the split-half standard error of its battery score, $.07$ for most executors and up to $.24$ for Llama-70B; $100$ draws; planted reference recomputed each time) the groups hold: bounded $46$--$100$; the mean probability that a metric keeps its printed regime is $.91$ (rising), $.82$ (reaches) and $.82$ (bounded); the modal verdict equals the printed one for $97\%$ of metrics; and, weighting each metric by these probabilities, the same three contrasts remain significant (reaches/local patterns $q=.0004$, rising/whole-text structure $q=.0004$, bounded/identity and social group $q=.008$). Under a differently constructed axis they do not (Table~\ref{tab:regime-axes}). The first component of the battery's five sub-scores orders the executors almost as $z$ does (rank correlation $.97$) but places the four strongest within $.13$ of each other where $z$ spreads them over $.42$; on it $30\%$ of verdicts change and the bounded group falls to $12$. Equally spaced ranks do the opposite and make $273$ curves bounded. Of the five contrasts, only reaches/local patterns remains significant on the battery's component and on ranks, none does on the circular axis built from the recovery values, and bounded/identity and social group does on $z$ and on ranks. The bounded verdict asks whether a fitted ceiling's upper limit falls below the planted reference at the top of the range, so it is decided by how far apart the strongest executors are on the axis: $z$ measures that spacing with the error propagated above, and the spacing is a property of the index, not of the criteria.

\begin{table}[h]
\small
\begin{center}
\resizebox{\textwidth}{!}{%
\begin{tabular}{l c c c c c l}
\toprule
capability axis & rank corr.\ with $z$ & rising & reaches & bounded & verdicts changed & contrasts surviving (of five) \\
\midrule
battery $z$ (the paper) & $1.00$ & $884$ & $321$ & $65$ & --- & $3$: reaches/local, rising/whole-text, bounded/identity \\
$z$ redrawn from its measurement error ($100$ draws) & --- & $674$--$1{,}013$ & $192$--$484$ & $46$--$100$ & $3\%$ (modal) & the same $3$ (probability-weighted) \\
first component of the battery's five sub-scores & $.97$ & $859$ & $373$ & $12$ & $30\%$ & $1$: reaches/local \\
first component of the recovery matrix (circular) & $.94$ & $1{,}012$ & $198$ & $47$ & $27\%$ & $0$ \\
executor ranks, equally spaced & $1.00$ & $269$ & $728$ & $273$ & $56\%$ & $3$: reaches/local, bounded/identity, bounded/beyond-text \\
\bottomrule
\end{tabular}}
\end{center}
\caption{\textbf{The regime verdicts under other capability axes.} Every curve is refitted with the shipped rule and the planted reference is recomputed on each axis; noisy verdicts are omitted from the counts ($0$; $2$--$28$; $26$; $13$; $0$). Contrasts are the five regime-by-type contrasts of Section~\ref{sec:scaling}, within-domain permutation with false-discovery control at $q<.05$. Artifact \texttt{outputs/analyses/osl\_pca\_v1/}.}
\label{tab:regime-axes}
\end{table}

\paragraph{What holds under every check, and what rests on the most capable executor.} The regime sizes, the domain composition of Figure~\ref{fig:osl-domains} and the three significant contrasts hold under the curve, exclusion and group checks and under measurement noise in both the recovery values and the capability index. Without each curve's most capable executor, the bounded count and the domain composition still hold, but a third of reaches curves become rising and only the reaches/local-patterns contrast stays significant. Three results are fragile: which particular metric is in which regime, most of all for bounded criteria, which keep their label in $65\%$ of recovery-noise draws and $82\%$ of index-noise draws on average; the reaches verdict of a curve whose saturation rests on the strongest executor; and, under a differently constructed capability axis, the bounded count, its domain composition and its identity-and-social signature (Tables~\ref{tab:regime-domains} and~\ref{tab:regime-axes}). The early saturation of local-pattern criteria is the one regularity that holds under every check except the circular axis built from the recovery values themselves.

\clearpage
\section{The two criterion codebooks: definitions, examples and how they relate}
\label{app:taxonomy}
The paper uses two codebooks: the six categories of Figure~\ref{fig:osl-types6}, merged from $12$, and the nine types behind the truth-condition axis of Figure~\ref{fig:osl-types}.

\subsection{The $12$-, $8$- and $6$-category codebook behind the main-text type figure}
\label{app:codebook-six}
\begin{figure}[h]
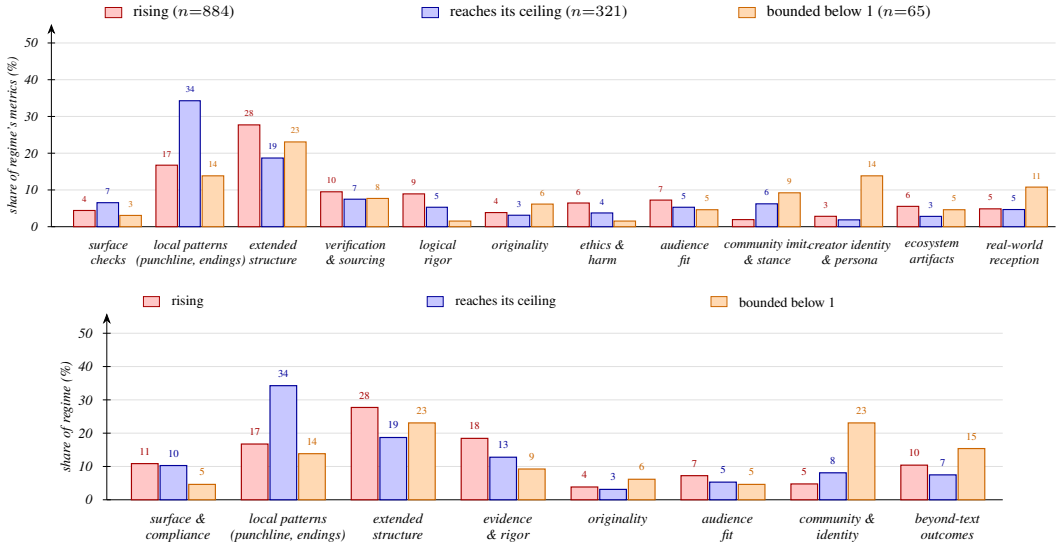

\centering
\resizebox{\textwidth}{!}{%
\begin{tikzpicture}[tlab/.style={font=\tiny\itshape, align=center}]
\input{figs/gen_osl_types12.tex}
\end{tikzpicture}}\\[4pt]
\resizebox{.9\textwidth}{!}{%
\begin{tikzpicture}[tlab/.style={font=\tiny\itshape, align=center}]
\input{figs/gen_osl_types8.tex}
\end{tikzpicture}}%
\caption{\textbf{OSL patterns for tasks, grouped into $k=12$ (top) and $k=8$ (bottom) categories}: the full codebook underlying the main text's $6$-category view (Fig.~\ref{fig:osl-types6}), which is a nested merge of these labels.}
\label{fig:osl-granularity}
\end{figure}

The six categories of Figure~\ref{fig:osl-types6} are deterministic merges of the $12$-category
codebook shown in Figure~\ref{fig:osl-granularity} (no re-labeling), which also draws an $8$-category intermediate: at $8$ categories, verification+rigor merge
into \emph{evidence \& rigor}, community+identity into one social category, and
ecosystem+reception into one beyond-text category; at $6$, compliance additionally folds into
the verifiable-features group and originality into audience fit. This appendix gives,
for each main-text category: its constituent $12$-category labels (with counts over the
$1{,}270$ fitted tasks), the truth-condition that defines it, and verbatim example criteria
from the banks. The organizing question throughout is \emph{what would make this criterion true},
not what a judge happens to inspect.

\paragraph{Local patterns within document} (\textsc{compact\_device}, $n{=}267$). A
self-contained craft device realized in a small span of text (a setup$\to$punchline pair, an
ending, a line-level economy rule), whose success or failure is visible at the sentence or
beat where it occurs. \emph{Examples:} ``Core joke structure: setup$\to$punch$\to$tags''
(humor); ``Chekhov's Gun: setup and payoff'' (creative writing); ``Active voice'' (press
releases); ``Abstract and introduction clarity'' (math).

\paragraph{Whole-text structural patterns} (\textsc{extended\_structure}, $n{=}320$).
Long-range compositional properties that exist only at the scale of the whole artifact:
arcs, escalation, and the coherence of a world or an argument maintained across the document.
\emph{Examples:} ``Worldbuilding coherence: rules and internal logic'' and ``Scene craft:
objective, conflict, change, escalation'' (creative writing); ``Accumulation and
format-subversion structures'' (humor); ``Choice and justification of tools'' (math).

\paragraph{Verifiable features} (merge of four: \textsc{surface\_check} $n{=}62$,
\textsc{verification\_sourcing} $n{=}113$, \textsc{logical\_rigor} $n{=}97$,
\textsc{ethics\_harm} $n{=}70$). Criteria whose truth conditions reduce to conformity with an
explicit, stateable standard: a format or convention rule (``Declare notation before use,''
math; ``Boilerplate inclusion and quality,'' press releases), factual accuracy and attribution
(``Accuracy, verification and transparency,'' news homepages; ``Attribution and sourcing
clarity,'' press releases), validity of reasoning (``Appropriate and economical proof
technique,'' math; ``Claim--evidence alignment and causal caution,'' peer review), or a
harm/representation policy (``Benign-violation calibration and cruelty check,'' humor; ``Avoid
discriminatory language, stereotypes, and framing,'' news homepages). The four merged labels share
checkability against a standard, in contrast to the two craft categories above (composition) and
the two social categories below (whose truth conditions leave the text).

\paragraph{Originality \& audience fit} (\textsc{originality\_novelty} $n{=}48$ $+$
\textsc{audience\_fit} $n{=}84$). Criteria judged relative to an expectation baseline: the
space of prior work (``Defamiliarization (estrangement),'' creative writing; ``News value and
angle strength,'' press releases) or a describable recipient class (``Audience and outlet
understanding, targeting, and tailoring,'' press releases; ``Accessibility and dependence on
shared context,'' humor). ``Audience'' here means tailoring to a describable recipient: the target can be
stated in a sentence, and fit is then judgeable from the text alone.

\paragraph{Identity \& social-group awareness} (\textsc{community\_stance} $n{=}43$ $+$
\textsc{identity\_persona} $n{=}40$). Properties of the text that can only be verified against
a specific social referent: participation in a community's practice through imitation, stance, and
culturally-local reference (``Parody craft and voice fidelity,'' ``Australian humor
conventions,'' humor; ``Fanfiction tropes, tagging, and canon alignment,'' creative writing),
or fidelity to the creator's own persona (``Distinct comic persona/voice and
authenticity,'' ``Personal history and lived experience as engine,'' humor). The contrast with
audience fit is depth of required knowledge: audience fit asks whether the text is tailored
\emph{to} a describable reader; these criteria ask whether the text successfully \emph{uses}
knowledge held \emph{within} a community or a person, which the evaluator must itself possess.
The two also behave differently in the data: in the nine-type labels, audience fit \emph{falls} toward the bounded regime
while community transformation rises monotonically (Appendix~\ref{app:osl-regimes}), which is evidence that the distinction is not only verbal.

\paragraph{Beyond-text features} (\textsc{ecosystem\_artifacts} $n{=}61$ $+$
\textsc{reception\_outcome} $n{=}65$). Criteria whose truth conditions are facts about the
world rather than about the text: companion artifacts and packaging (``Broadcast-usable B-roll
provided,'' press releases; ``Artifact availability, licensing, and FAIR discoverability,''
peer review) and achieved reception (``Memetic design, stickiness, and virality,'' humor;
``Agenda setting and attribute salience,'' news homepages; ``Catharsis of Pity and Fear,''
creative writing). This category differs from the previous one: identity \& social-group criteria
need external \emph{knowledge} to verify a property the text itself bears; beyond-text criteria
need external \emph{facts}, since no reading of the text, however informed, settles whether it
shipped with assets or actually went viral.

\subsection{The nine-type codebook and the truth-condition axis}
\label{app:codebook-nine}
The axis of Figure~\ref{fig:osl-types} comes from a separate codebook of nine types, applied to all $1{,}270$ fitted criteria with the regime held out of the labeling input: \emph{mechanical checks} (surface/compliance rules);
\emph{craft operations} (discrete inspectable compositional moves); \emph{evidence \& rigor}
(accuracy, verification, sourcing, logical correctness); \emph{holistic taste} (graded overall
aesthetic judgment); \emph{audience fit} (tailoring to an audience, venue, or genre);
\emph{community transformation} (community-specific imitation or stance: parody fidelity, satire
target, culturally-local reference); \emph{identity \& persona} (the creator's own voice, lived
experience, branding); \emph{external bundles} (artifacts and ecosystems beyond the text); and
\emph{reception outcomes} (achieved effects on real audiences or platforms: virality, salience,
resonance). These types coarsen onto a single axis, \emph{where the criterion's truth
conditions live}: \textbf{in the text} (mechanical, craft, evidence, taste), \textbf{at the
interface} (audience fit, community transformation), or \textbf{beyond the text} (identity,
bundles, reception).

\subsection{Boundary cases, reliability, and how the two codebooks relate}
\label{app:codebook-relation}
Two caveats apply to these definitions.
First, the least clear boundary in the codebook is between craft and reception: a criterion like ``Catharsis
of Pity and Fear (Aim and Conditions)'' can be read as conditions the text must set up (a craft
operation) or as an effect it must achieve (a reception outcome). The labeling convention
resolves it by truth-maker (catharsis is an achieved audience response, so it is reception),
but the two independent passes disagree most in this region (of the $65$
reception-labeled tasks, $32$ are reception in both nine-type passes and $10$ are craft in
both). The regime-level findings do not depend on these cases: the bounded externality shift
replicates in the second pass (Appendix~\ref{app:osl-regimes}). Second, this six-category merge and
the coarse truth-condition axis of Fig.~\ref{fig:osl-types} are \emph{different, non-nesting}
coarsenings of two codebooks: the six categories merge the $12$-category labels, which come from a single blinded pass, and the axis coarsens the nine-type labels, which come from two independent passes ($\kappa=.75$ on the nine types, $.73$ on the axis). The axis places community transformation \emph{at the interface} but
identity and persona \emph{beyond the text}, so the ``identity \& social-group awareness'' bar
deliberately groups by social anchoring and straddles the axis. Externality indices are computed
from the axis labels, never from this merge.

\section{Adding examples to the message, and what each process writes into the prompt}
\label{app:examples}
\subsection{Every version of the examples test: who picks the examples, which label they carry, and which models run them}
\label{app:examples-versions}

Section~\ref{sec:scaling} tests whether examples added to a definition raise recovery. This appendix documents every version of that test. All versions share one message format and one measure. An example-bearing message is the criterion's name and definition followed by a list headed ``Examples that satisfy this criterion:'' and a list headed ``Examples that do NOT satisfy it:''. The transmitter labels held-out items under that message; the receiver sees $30$ of those labeled items, balanced by label and drawn from a fixed subset of $60$, and writes a rule of at most $450$ tokens; the transmitter runs the rule on the remaining held-out items, and recovery is $I(m;m')$ in bits (Section~\ref{sec:reconstruction}). In every evaluation one model runs the messages and writes the rules. Every comparison is the change in recovery against the name and definition alone, on the same items and at the same model. The versions differ in who picks the examples, which label each example carries, and which models run the messages (Table~\ref{tab:examples-methods}). Across them, no way of choosing or writing examples beats arbitrary examples carrying the same kind of label.

\begin{table}[h]
\small
\begin{center}
\resizebox{\textwidth}{!}{%
\begin{tabular}{p{4.0cm} p{3.0cm} c p{2.1cm} p{4.0cm} p{3.4cm}}
\toprule
examples added & chosen by; label shown & criteria & models & change in recovery vs.\ definition (bits) & vs.\ arbitrary examples \\
\midrule
$0$--$5$ items that move the transmitter toward the labels two judges agree on & Llama-3.1-8B; label chosen with the item & $236$ & Llama-70B, Qwen-72B & pooled $+.002$; $3$ criteria above the noise bound at both judges (${\sim}0.6$ expected) & arbitrary sets set the bound \\
$3$ YES and $3$ NO items the definition's reconstruction labels most differently & each judge; its own label & $153$ / $146$ & Llama-70B, Qwen-72B & pooled $+.001$; $0$ above the bound at both judges & -- \\
$3$ near-identical pairs labeled oppositely & each judge; its own label & $125$ / $112$ & Llama-70B, Qwen-72B & pooled $+.004$; $2$ above the bound at both judges & -- \\
six items chosen for the receiver's decoding & Llama-3.1-8B with Qwen2.5-14B; label chosen with the item & $130$ & Llama-70B & bounded $-.001$ $[-.017,+.016]$; controls $+.002$ $[-.016,+.019]$ & $+.008$ $[-.006,+.021]$; $-.001$ $[-.022,+.019]$ \\
the same, with fixed labels & the same pair; transmitter's own label & $130$ & Llama-70B & bounded $+.002$ $[-.019,+.020]$; controls $+.019$ $[-.001,+.039]$ & $-.013$ $[-.029,+.003]$; $+.005$ $[-.009,+.021]$ \\
six items chosen at the judge & Llama-70B; its own label & $48$ & Llama-70B & bounded $+.028$ $[-.002,+.062]$; controls $+.028$ $[+.005,+.055]$ & $+.014$ $[-.016,+.048]$; $+.026$ $[-.007,+.059]$ \\
six pieces by an author and six by others & true authorship & $12$ authors & nine receivers, 1B--72B & author identification, not recovery (see text) & placebo sets at chance \\
four real items drawn at random, two the transmitter labels YES and two it labels NO & chance draw; transmitter's own label & $449$ & seven transmitters (3B--70B and Phi-4); the interval pools Qwen2.5-14B and Llama-70B & $+.006$ $[+.003,+.008]$ over Qwen2.5-14B and Llama-70B; negative at 7B and 8B (Table~\ref{tab:examples-self}) & these are the arbitrary examples \\
$2$ YES and $2$ NO written by the transmitter & the transmitter; as written & $449$ & seven transmitters (3B--70B and Phi-4) & Table~\ref{tab:examples-self} & at most $+.001$ above four random labeled items \\
four items searched for recovery & Llama-70B; its own label & $1$ & Llama-70B & two YES and two NO: $-.003$; any four: $+.098$ & random sets: $+.006$ to $+.027$; $+.021$ to $+.139$ \\
the receiver answers with $4$, $8$ or $30$ of the items it saw instead of a rule & the receiver; transmitter's own label & $449$ & Qwen2.5-14B and Llama-70B & $4$ chosen items vs.\ the rule: $-.014$ $[-.017,-.010]$ at Qwen2.5-14B, $-.013$ $[-.017,-.009]$ at Llama-70B; all $30$: $-.008$ and $-.007$; bounded $-.024$ and $-.029$ (Table~\ref{tab:examples-shown}) & chosen vs.\ random items $.000$ at both models \\
\bottomrule
\end{tabular}}
\end{center}
\caption{\textbf{Every version of the examples test.} Changes are means over criteria, with $95\%$ bootstrap intervals where they were computed. ``Bounded'' and ``controls'' are the $65$ bounded criteria with a bank definition and one matched reaches or rising criterion each. Where two counts are given, they are the criteria covered at Llama-70B and at Qwen-72B. The last column compares the chosen examples with arbitrary examples carrying the same kind of label.}
\label{tab:examples-methods}
\end{table}

\paragraph{Examples that move the transmitter toward agreed labels.} The first version, shown in a panel of Figure~\ref{fig:message-forms}, ran on $236$ criteria, drawn up to six per task and criterion type from the fitted bank. Llama-3.1-8B was the transmitter during selection. Candidates came from a design split of each task's items; held-out items were never candidates. Greedy rounds tried eight candidates at a time, each with a proposed label, and kept the best one only if it raised the transmitter's balanced agreement with the labels that Llama-70B and Qwen-72B both gave under the definition by at least $.01$ on one design split without lowering it on a second; sets were capped at $12$. That rule kept few examples: $131$ of the $236$ criteria received none (so their example-bearing message equals the definition), $58$ received one, $36$ two, and $11$ three to five. Evaluation used recovery alone, at Llama-70B and at Qwen-72B separately. Over all criteria and both judges, recovery is $.046$ bits for the definition and $.048$ with the examples. The $131$ criteria without examples show what reconstruction noise alone produces: their change has mean $+.0005$ and standard deviation $.025$ bits. The noise bound for one criterion is the $90$th percentile of the absolute change an arbitrarily chosen set produces, measured on $79$ criteria per judge ($.064$ bits at Llama-70B, $.057$ at Qwen-72B); about $5\%$ of arbitrary sets exceed it upward. Seventeen criteria exceed it at Llama-70B and $13$ at Qwen-72B, against about $12$ expected at each, and eight of the $17$ received no example. Three exceed it at both judges, against about $0.6$ expected if the two judges' noise is independent: ``delivery and timing craft'' ($+.096$ and $+.069$ bits, one example), ``incongruity and surprise mechanics'' ($+.071$ and $+.104$, one example) and ``conflict/tension as narrative and news value'' ($+.108$ and $+.081$, four examples). One criterion falls below the bound at both. Per-criterion changes correlate at $\rho=.15$ across the two judges. No task's mean change separates from zero, with humor highest at $+.011$ $[-.005,+.026]$ and peer review lowest at $-.005$; of the six criterion types, only local patterns is above zero, at $+.010$ $[+.002,+.019]$.

\paragraph{Items the definition's reconstruction gets wrong, and near-identical pairs.} Two further rules reused the same criteria, candidate pools, held-out items and evaluation, and labeled every example with the judge's own verdict under the definition, so each judge chose its own sets. The first took the three YES and three NO candidates on which the judge's verdict under the definition differs most from its verdict under the rule reconstructed from the definition-only message, that is, the items the definition fails to transmit. The second took the three most similar pairs of candidates, by character-trigram overlap, that the judge labels confidently and oppositely (probability of YES above $.6$ against below $.4$), since rival rules disagree on such pairs. A criterion whose candidates the judge labels all one way gets no set, which leaves $153$ criteria at Llama-70B and $146$ at Qwen-72B for the first rule, and $125$ and $112$ for the second. Pooled over both judges, recovery changes by $+.001$ and $+.004$ bits. No criterion exceeds the noise bound defined above at both judges under the first rule, and two do under the second (``premise and throughline clarity'' and ``cognitive surprise and audience processing fit''). No task's mean separates from zero, and humor turns negative under both rules ($-.015$ and $-.019$). Under the first rule, local-pattern criteria are again the one type above zero, at $+.022$ $[+.010,+.035]$ over $28$ criteria.

\paragraph{Examples chosen for the receiver's decoding.} The rules above choose examples by their effect on the transmitter's labels. This rule chooses them for the receiver. A candidate set of six is scored by the round trip on the design items alone: a Llama-3.1-8B transmitter runs the message, a Qwen2.5-14B receiver writes a rule from those labels, the transmitter runs that rule, and the score is the agreement of the two labelings on design items the receiver did not see. Sets are built by greedy forward selection over the same $24$-candidate pools, and a null set of six is drawn by hash. Each shown item carries the transmitter's own verdict under the definition, so the selector chooses items and never labels. The sample is every bounded criterion with a bank definition ($65$) and one matched control each, evaluated at Llama-70B on held-out items. Under the definition, the selected six and the null six, recovery is $.079$, $.081$ and $.093$ bits for bounded criteria, $.125$, $.132$ and $.138$ for reaches criteria, and $.037$, $.066$ and $.051$ for rising criteria. The selected six change recovery by $+.002$ $[-.019,+.020]$ for bounded criteria and $+.019$ $[-.001,+.039]$ for controls, and differ from the null six by $-.013$ $[-.029,+.003]$ and $+.005$ $[-.009,+.021]$. Six bounded and eight control criteria clear the $90$th percentile of the null sets' change ($.103$ bits), against $6.5$ expected each. The one interval above zero is for rising criteria, $+.029$ $[+.005,+.058]$. The judge's labels under the selected six agree with its definition-only labels at $.85$ balanced accuracy, and at $.84$ under the null six. A first run of this rule also let the selector choose each shown label: those labels matched the transmitter's own verdict on only $54\%$ of items, the gains on the design items did not transfer, and at the judge the set changed recovery by $-.001$ $[-.017,+.016]$ for bounded criteria and $+.002$ $[-.016,+.019]$ for controls. Definition-only recovery reproduces across the two runs at $r=.97$.

\paragraph{Examples chosen at the judge itself.} Running the same search with Llama-70B as both transmitter and receiver sets the ceiling for this rule. On a stratified subset of $48$ of the $130$ criteria ($24$ bounded, $24$ controls), bounded criteria recover $.083$, $.111$ and $.097$ bits under the definition, the selected six and the null six. The selected six raise recovery over the definition by $+.028$ $[-.002,+.062]$ for bounded criteria and $+.028$ $[+.005,+.055]$ for controls, and exceed the null six by $+.014$ $[-.016,+.048]$ and $+.026$ $[-.007,+.059]$. Five bounded and three control criteria clear the null sets' $90$th percentile ($.062$ bits), against $2.4$ expected each. Against selection at the Llama-3.1-8B and Qwen2.5-14B pair on the same criteria, selecting at the judge changes the gain by $+.019$ $[-.004,+.043]$ for bounded criteria and $-.008$ $[-.031,+.013]$ for controls, and definition-only recovery agrees across the runs at $r=.98$. The judge's labels under the selected six agree with its definition-only labels at $.88$ for bounded criteria and $.80$ for controls.

\paragraph{Six pieces identify an author no better than a description of the voice does.} Authorship is a criterion with ground truth, so here the measure is identification rather than recovery. For each of $12$ McSweeney's authors, a receiver answered YES or NO to whether each of $40$ held-out pieces, half by the author, is by the same author, and we scored balanced accuracy. The message was a description of the voice of at most $150$ words, written by gpt-oss-120b from the author's pieces; that description padded to the length of the examples; six pieces by the author and six by others; or a placebo in the same format with another author's pieces in place of this author's. The preregistered comparison is the examples against the padded description. Averaged over authors, examples minus padded description is $-.002$ at Llama-3.2-1B, $-.033$ at Llama-3.2-3B, $-.075$ at Llama-3.1-8B and $+.048$ at Llama-3.3-70B, and $+.002$, $-.031$, $+.050$, $+.069$ and $+.046$ at Qwen2.5-3B, 7B, 14B, 32B and 72B. The placebo stays at chance for every receiver ($.475$ to $.508$). The preregistered rule asked for a pooled gain above $.05$ over the two largest receivers of each family; the pooled gain is $+.022$.

\paragraph{Examples the transmitter writes itself.} The rules above choose among existing items; here the transmitter writes its own. Given the name and definition, it writes two short examples that satisfy the criterion and two that do not, each under $200$ characters (sampling temperature $.7$, fixed seeds, redrawn if the reply does not parse; $2$ of $2{,}245$ criterion--model pairs never parsed). The four examples then go under the definition, and the transmitter labels held-out items with that message. Two controls isolate what the examples contribute. In the first, the same four texts appear under exchanged headings, so only which texts are marked YES changes. In the second, the four examples are real design items that the transmitter itself labeled YES (two) and NO (two) under the definition alone, drawn at random with a fixed seed and shown in the same format and length, so only the texts change, from written prototypes to found items. We ran seven transmitters, each also its own receiver, on $449$ criteria: the $130$ criteria above (the $65$ bounded criteria with a bank definition and one matched reaches or rising control each) and $319$ reaches and rising criteria drawn per task, up to $60$ per task, from all eight tasks (Table~\ref{tab:examples-self}). The draw was fixed before any run. A criterion was eligible if it had a bank definition and a fitted regime. For each task, rising and reaches criteria were added alternately by regime, each regime's pool in a stable-hash order seeded by task and regime, until the task held $60$ criteria or its pool was exhausted. Humor keeps all $41$ bounded criteria and their $41$ controls and so has $82$; patents has $7$ eligible criteria; news homepages, peer review and creative writing reach $60$ with $34$, $42$ and $56$ drawn criteria on top of $26$, $18$ and $4$ from the base set; math, notice-and-comment and press releases have no bounded criteria and take $60$ drawn criteria each. No panel or consensus label enters the draw.

\begin{table}[h]
\small
\begin{center}
\resizebox{\textwidth}{!}{%
\begin{tabular}{l c c c c c c c}
\toprule
 & & \multicolumn{3}{c}{change against the definition} & & & \\
\cmidrule(lr){3-5}
transmitter & definition & own examples & headings swapped & random labeled items & own minus random & own minus swapped & labeled as written \\
\midrule
Qwen2.5-3B & $.001$ & $.000$ $[-.001,+.000]$ & $-.001$ $[-.001,-.000]$ & $-.001$ $[-.001,-.000]$ & $+.001$ $[+.000,+.001]$ & $.000$ $[-.000,+.001]$ & $.51$ \\
Llama-3.1-8B & $.008$ & $-.001$ $[-.002,-.001]$ & $-.002$ $[-.003,-.002]$ & $-.001$ $[-.002,-.001]$ & $.000$ $[-.001,+.001]$ & $+.001$ $[+.000,+.002]$ & $.79$ \\
Qwen2.5-7B & $.042$ & $-.005$ $[-.009,-.002]$ & $-.010$ $[-.013,-.007]$ & $-.005$ $[-.008,-.002]$ & $.000$ $[-.004,+.003]$ & $+.004$ $[+.002,+.007]$ & $.77$ \\
Phi-4 & $.035$ & $-.008$ $[-.012,-.005]$ & $-.012$ $[-.015,-.008]$ & $-.001$ $[-.005,+.002]$ & $-.007$ $[-.010,-.004]$ & $+.004$ $[+.001,+.006]$ & $.80$ \\
Qwen2.5-14B & $.036$ & $.000$ $[-.003,+.004]$ & $-.012$ $[-.015,-.008]$ & $+.005$ $[+.002,+.008]$ & $-.004$ $[-.008,-.001]$ & $+.012$ $[+.008,+.016]$ & $.74$ \\
Qwen2.5-32B & $.037$ & $+.003$ $[-.001,+.006]$ & $-.001$ $[-.005,+.002]$ & $+.007$ $[+.003,+.011]$ & $-.004$ $[-.009,-.000]$ & $+.004$ $[+.001,+.007]$ & $.83$ \\
Llama-3.3-70B & $.043$ & $+.001$ $[-.003,+.004]$ & $-.013$ $[-.018,-.009]$ & $+.007$ $[+.003,+.011]$ & $-.006$ $[-.011,-.002]$ & $+.014$ $[+.009,+.018]$ & $.82$ \\
\bottomrule
\end{tabular}}
\end{center}
\caption{\textbf{Examples the transmitter writes itself}, $449$ criteria from eight tasks; Phi-4 ran where the press-release items were unavailable, so its row covers the other seven tasks, $389$ criteria. Definition-only recovery in bits, then mean changes over criteria with $95\%$ bootstrap intervals. ``Random labeled items'' are four real design items labeled by the transmitter under the definition alone. The last column is the share of its own four examples that the transmitter, given only the definition, labels the way it wrote them. Each model is its own transmitter and receiver.}
\label{tab:examples-self}
\end{table}

Three results hold across the five transmitters of the first run (the Qwen2.5-32B and Phi-4 rows were added afterwards; Table~\ref{tab:examples-self}). First, the transmitter's own examples never beat real items carrying its own labels by more than $.001$ bits, and they fall below them at Qwen2.5-14B and Llama-70B. Second, the larger transmitters use the headings. Exchanging them costs $.012$ bits at Qwen2.5-14B and $.013$ at Llama-70B, against at most $.002$ at Qwen2.5-3B and Llama-3.1-8B, because swapped headings push the labels toward NO: at Llama-70B the number of criteria on which every held-out item is labeled NO rises from $121$ under the definition to $202$, and the information the labels contain about the items falls from $.421$ to $.295$ bits. At these two transmitters the same examples under their correct headings leave recovery at its definition-only level. Third, at the same two transmitters, real labeled items raise recovery by $.005$ and $.007$ bits; at Llama-70B they reduce the criteria labeled all NO to $92$ and raise the information in the labels to $.473$ bits. At Qwen2.5-7B and Llama-3.1-8B, any four examples lower recovery slightly, and Qwen2.5-3B labels every held-out item NO on $374$ criteria under the definition. For absurdism, Llama-70B, Qwen2.5-7B and Qwen2.5-14B each labeled all four of their own examples absurdist under the definition, including the two they wrote as examples that do not satisfy it.

\paragraph{Why real items beat written ones.} Real labeled items mostly keep the transmitter from answering NO to everything. Under the definition alone, Llama-70B labels every held-out item NO on $121$ of the $449$ criteria, and a labeling that never varies leaves nothing to recover. With four real labeled items, $39$ of those criteria get some YES labels, against $25$ with the transmitter's own written examples, and the gain from real items is largest there: $+.010$ bits on those $121$ criteria against $+.006$ on the other $328$, where written examples give $+.004$ and $-.001$. The two kinds of examples differ most in long-document tasks, where an example under $200$ characters is a sketch rather than an item (Table~\ref{tab:examples-real-vs-written}). The gain from real items does not depend on whether the transmitter labels its own written examples as written: $+.008$ $[+.002,+.014]$ on the $229$ criteria where it labels all four as written and $+.006$ $[+.000,+.012]$ on the other $220$. It is no smaller on the $109$ criteria where the transmitter labeled too few design items YES or NO to fill a set of two and two ($+.009$, against $+.006$ where the set was full). Three criteria that Section~\ref{sec:scaling} names are outside the $449$: we ran the same condition on them at both models, where delivery and timing craft gains $+.036$ and $+.196$ bits, incongruity and surprise mechanics $-.021$ and $+.079$, and novelty and significance of contribution $+.018$ and $+.062$. Conflict/tension as news value gains $+.051$ at Qwen2.5-14B ($+.037$ pooled over both models, the value Section~\ref{sec:scaling} reports).

\begin{table}[h]
\small
\begin{center}
\begin{tabular}{l c c}
\toprule
task & Llama-70B & Qwen2.5-14B \\
\midrule
math answers & $-.010$ $[-.018,-.002]$ & $-.008$ $[-.019,+.003]$ \\
public comments & $-.012$ $[-.025,-.001]$ & $-.007$ $[-.014,-.002]$ \\
creative writing & $-.003$ $[-.010,+.003]$ & $-.014$ $[-.023,-.005]$ \\
humor & $-.007$ $[-.022,+.007]$ & $+.005$ $[-.002,+.012]$ \\
\bottomrule
\end{tabular}
\end{center}
\caption{\textbf{Written examples minus real labeled items}, by task: mean change in recovery in bits, with $95\%$ bootstrap intervals over the criteria of each task ($60$, $60$, $60$ and $82$).}
\label{tab:examples-real-vs-written}
\end{table}

\paragraph{By regime, and criteria beyond two standard deviations of noise.} One pattern holds across the versions above: the transmitter's own written examples move bounded criteria more than any other regime, but the direction depends on the model. Bounded recovery changes by $+.011$ $[+.003,+.019]$ bits at Qwen2.5-14B and $+.007$ $[-.006,+.021]$ at Llama-70B, and by $-.023$ $[-.037,-.011]$ at Phi-4, $-.018$ $[-.028,-.007]$ at Qwen2.5-7B, $+.001$ at Qwen2.5-32B, $-.003$ at Llama-3.1-8B and $-.001$ at Qwen2.5-3B, while reaches and rising criteria stay within $\pm.008$ at every model; pooled over the three models with real headroom, bounded criteria change by $-.002$ $[-.008,+.005]$, so the gain at the two largest models does not hold in a third family. Four real items carrying the transmitter's own labels help every regime about equally at Qwen2.5-14B and Llama-70B ($+.007$, $+.005$ and $+.006$ bits) and lower bounded recovery at Phi-4 and Qwen2.5-7B ($-.017$ and $-.015$). Per criterion, reconstruction alone gives the change between two messages a standard deviation of about $.028$ bits at Llama-70B; there, random labeled items raise recovery by more than two standard deviations on $43$ of the $449$ criteria and lower it on $22$, and the transmitter's own examples on $32$ and $30$, against about $10$ each by chance. At Qwen2.5-14B real labeled items raise it on $42$, and $4$ criteria clear the bound at both models, about as many as two unrelated lists of those sizes would share; the per-criterion gains correlate at $r=-.08$ across the two models. By regime, real labeled items help $7$ of the $65$ bounded criteria at each of those two models, $12$ and $15$ of the $152$ reaches criteria and $23$ and $21$ of the $232$ rising criteria, against $1.5$, $3.5$ and $5.3$ by chance: about one criterion in ten in every regime. The transmitter's own written examples help $8$ and $8$ bounded criteria, $10$ and $9$ reaches criteria and $11$ and $15$ rising criteria, and three of the four criteria they help at both models are bounded: game identification and escalating beat development, memetic design and virality, and agenda setting and attribute salience.

\paragraph{The receiver answers with examples.} Every version above puts examples into the transmitter's message and still asks the receiver for a rule. The last version changes the receiver's output instead. For each of the $449$ criteria, the receiver (Qwen2.5-14B or Llama-70B, also the transmitter and executor) sees the same $30$ balanced labeled items as in the rule channel and returns either a rule, or two YES and two NO items of its choosing, or four and four, or all $30$; a random two-and-two draw and a rule plus the chosen four are the controls. The executor then labels the held-out items from the returned items alone, with no name, definition or rule, and recovery is $I(m;m')$ as everywhere else. Two floors are measured: the same items with their labels permuted (the permutation least correlated with the true labels, since a plain shuffle leaves a two-and-two set unchanged or inverted one time in three), and a nearest-neighbour labeler with no model (share of YES among the three nearest shown items by BAAI/bge-large-en-v1.5 cosine). The design was preregistered (\nolinkurl{notes/2026-10-08__exp-examples-shown-prereg.md}); the rule channel is the existing one with $R=2$ reconstructions per criterion, the example conditions have $R=8$. Table~\ref{tab:examples-shown} gives the result at Qwen2.5-14B and at Llama-70B; the noise standard deviation is $.052$ at Qwen2.5-14B and $.055$ at Llama-70B.

Showing recovers less than telling. Four chosen items recover $.022$ bits against $.036$ for the rule over all $449$ criteria ($-.014$ $[-.017,-.010]$), $.011$ against $.034$ on the $65$ bounded criteria ($-.024$ $[-.034,-.015]$), and the bounded-minus-controls interaction is $-.001$ $[-.014,+.013]$. No criterion gains more than two noise standard deviations ($.104$ bits, the standard deviation being $.052$) over its rule; $4\%$ lose that much. Recovery rises with the number of items and reaches the rule only when all $30$ are passed along, and only on the bounded criteria ($-.003$ $[-.016,+.009]$ on the $32$ bounded criteria with at least two items per class). The receiver's choice of items adds nothing over a random draw ($.000$ $[-.001,+.001]$), and a rule plus four items is worse than the better of the two alone ($-.010$ $[-.012,-.007]$). The nearest-neighbour floor is near zero ($.006$ at $30$ items), so the executor is not matching surface similarity; but the permuted-label floor keeps most of the recovery ($.018$ of $.022$ at four items, $.021$ of $.028$ at $30$), so the label-dependent part of showing is $.005$ $[+.003,+.006]$ bits at four items and $.007$ $[+.006,+.009]$ at $30$. On the $261$ criteria with at least two items per class those parts are $.008$ and $.012$; on the other $188$ the transmitter labels nearly every item one way, so no two-and-two set exists, and they are flagged in the artifact.

Llama-70B gives the same ordering with larger numbers. Four chosen items recover $.030$ bits against $.043$ for the rule over all $449$ criteria ($-.013$ $[-.017,-.009]$), $.047$ against $.076$ on the bounded criteria ($-.029$ $[-.042,-.017]$), and the bounded-minus-controls interaction is $-.006$ $[-.023,+.012]$. One criterion gains more than two noise standard deviations ($.110$ bits) over its rule and $3\%$ lose that much. Recovery rises with the number of items but does not reach the rule even with all $30$: $-.007$ $[-.011,-.003]$ over all criteria, $-.020$ $[-.032,-.010]$ on the bounded ones, and $-.030$ $[-.047,-.016]$ on the $43$ bounded criteria with at least two items per class. The receiver's choice of items again adds nothing ($.000$ $[-.002,+.002]$), and a rule plus four items is again below the better of the two ($-.007$ $[-.009,-.004]$). The nearest-neighbour floor is $.007$ at $30$ items. The permuted-label floor keeps $.025$ of $.030$ at four items and $.027$ of $.036$ at $30$, so the label-dependent part is $.005$ $[+.003,+.008]$ and $.009$ $[+.007,+.013]$ bits; on the bounded criteria it is $.002$ $[-.004,+.007]$ at four items and $.016$ $[+.010,+.022]$ at $30$. At Llama-70B $305$ criteria have at least two items per class and $144$ do not.

Letting both sides mix a rule with examples changes nothing. We re-ran the Qwen2.5-14B version with the transmitter's labels produced under its definition plus four of its own labeled items (two YES, two NO, drawn at random from the design split), so that the receiver reconstructs a message that already combines rule and examples, and scored every condition above against those labels. Every condition rises by $.002$ to $.005$ bits (the rule to $.040$ over all $449$ criteria and $.039$ on the bounded ones; thirty shown items to $.031$ and $.033$), no ordering changes, and the receiver's rule plus four items still recovers less than its rule alone, $.034$ against $.040$ over all criteria and $.025$ against $.039$ on the bounded ones (artifact \texttt{outputs/exp\_examples\_shown/summary\_qwen25-14b\_random\_own.json}).

\begin{table}[h]
\footnotesize
\begin{center}
\resizebox{\textwidth}{!}{%
\begin{tabular}{l c c c c c c}
\toprule
& \multicolumn{3}{c}{Qwen2.5-14B} & \multicolumn{3}{c}{Llama-70B} \\
\cmidrule(lr){2-4} \cmidrule(lr){5-7}
receiver output (executor sees only this) & bounded ($65$) & controls ($65$) & all ($449$) & bounded ($65$) & controls ($65$) & all ($449$) \\
\midrule
rule (existing channel, $R=2$) & $.034$ $[.023,.047]$ & $.044$ $[.029,.060]$ & $.036$ $[.031,.041]$ & $.076$ $[.053,.100]$ & $.078$ $[.053,.106]$ & $.043$ $[.037,.050]$ \\
$4$ items, random & $.013$ $[.009,.019]$ & $.021$ $[.014,.030]$ & $.022$ $[.019,.025]$ & $.048$ $[.033,.065]$ & $.049$ $[.034,.066]$ & $.030$ $[.026,.035]$ \\
$4$ items, receiver-chosen & $.011$ $[.007,.015]$ & $.021$ $[.014,.029]$ & $.022$ $[.019,.025]$ & $.047$ $[.032,.063]$ & $.055$ $[.038,.074]$ & $.030$ $[.026,.035]$ \\
$8$ items, receiver-chosen & $.024$ $[.015,.033]$ & $.031$ $[.020,.042]$ & $.026$ $[.023,.030]$ & $.055$ $[.037,.074]$ & $.060$ $[.040,.082]$ & $.035$ $[.030,.041]$ \\
all $30$ items & $.031$ $[.021,.043]$ & $.038$ $[.025,.053]$ & $.028$ $[.024,.032]$ & $.056$ $[.040,.074]$ & $.059$ $[.040,.080]$ & $.036$ $[.031,.042]$ \\
rule plus $4$ chosen items & $.024$ $[.017,.031]$ & $.025$ $[.017,.035]$ & $.031$ $[.027,.035]$ & $.072$ $[.051,.096]$ & $.067$ $[.046,.090]$ & $.042$ $[.036,.048]$ \\
\midrule
$4$ items, labels permuted (floor) & $.009$ $[.005,.013]$ & $.014$ $[.009,.021]$ & $.018$ $[.016,.021]$ & $.045$ $[.031,.061]$ & $.041$ $[.027,.057]$ & $.025$ $[.021,.029]$ \\
$30$ items, labels permuted (floor) & $.020$ $[.013,.029]$ & $.025$ $[.016,.035]$ & $.021$ $[.018,.024]$ & $.040$ $[.027,.056]$ & $.041$ $[.027,.057]$ & $.027$ $[.023,.031]$ \\
nearest neighbour on $30$ items, no model & $.002$ & $.004$ & $.006$ & $.004$ & $.003$ & $.007$ \\
\bottomrule
\end{tabular}}
\end{center}
\caption{\textbf{The receiver answers with examples instead of a rule} (each of Qwen2.5-14B and Llama-70B as transmitter, receiver and executor; $449$ criteria). Recovery $I(m;m')$ in bits, mean over criteria with $95\%$ bootstrap intervals. ``Bounded'' and ``matched controls'' are the $65$ bounded criteria with a bank definition and their matched reaches or rising criteria. Noise floor: two standard deviations of run-to-run reconstruction noise is $.104$ bits per criterion at Qwen2.5-14B and $.110$ at Llama-70B.}
\label{tab:examples-shown}
\end{table}

\paragraph{Examples searched for recovery on one criterion.} For ``absurdism and non sequitur technique'' we searched directly for examples that raise recovery, with Llama-70B as transmitter and receiver. Under the definition alone the model labels $130$ of the $148$ design jokes absurdist. We drew $40$ sets of two YES and two NO design items and $40$ sets of any four, each item shown with the model's own label, scored each set by the round trip on design items, re-scored the best three with more reconstructions, and tested the best set of each kind on held-out jokes against the definition and three random sets of the same kind. Held-out recovery is $.190$ bits under the definition. The chosen two-and-two set recovers $.187$ and the random ones $.217$, $.196$ and $.211$; the chosen set of any four recovers $.288$ and the random ones $.211$, $.329$ and $.242$. The chosen sets are ordinary reddit jokes, not recognizably absurdist ones.
\subsection{What each process writes into the prompt: concept content against measurement content}
\label{app:change-types}

\begin{figure}[h]
	\centering
\resizebox{.8\textwidth}{!}{%
\begin{tikzpicture}[tlab/.style={font=\tiny\itshape, align=center}]
\IfFileExists{figs/gen_change_types_simple.tex}{\input{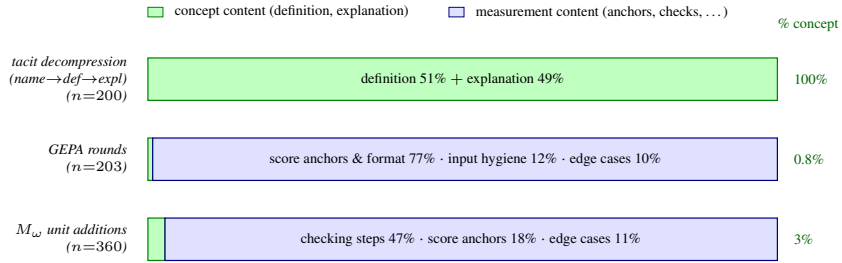}}{}
\end{tikzpicture}}%
	\caption{\textbf{What each process writes: concept vs.\ measurement content} (unsupervised
 tasks; two-pass labeling; the two-way collapse shown here has $\kappa=.93$, against
 $\kappa=.68$ for the full 8-way codebook, whose distribution this bar chart summarizes, with
 dominant subtypes named in each bar). Tacit decompression writes concept content
 exclusively; optimization, whether reflective or recombinative, writes measurement content
 almost exclusively. The divide is crossed only for esoteric concepts that the judge does not
 already hold.}
	\label{fig:change-types}
\end{figure}

Decompressing a criterion for a smaller executor and optimizing a prompt for score fidelity are
procedurally the same operation: text is added to a prompt to improve behavior. To locate where
they differ, we labeled the \emph{content} of every revision step, in one shared format, across
three corpora of unsupervised-task prompt changes: $200$ decompression-ladder steps
(name$\to$definition$\to$explanation, ten annotation domains; this decompression corpus spans ten domains rather than the eight of the scaling analyses), $203$ GEPA optimization rounds
(judge-criteria head-to-heads plus a $198$-rubric regulatory-comment bank), and the $360$
recombination units of the $M_\omega$ pools. Each step was tagged with the dominant kind of
material it adds: a \emph{definition} (what the quality is), an \emph{explanation} (why it
arises), \emph{checking steps}, \emph{edge cases and exclusions}, \emph{score anchors and output
format}, \emph{what to ignore in the input}, \emph{examples}, or \emph{rewording only}. The tags come from two
independent blinded passes with planted anchors (8-way $\kappa=.68$; collapsed to
concept-content vs.\ measurement-content, $\kappa=.93$, and every frequent disagreement lies
within the measurement family).

Figure~\ref{fig:change-types} shows the result. Decompression steps are $100\%$
concept-content: the name$\to$definition rung adds definitions, and the definition$\to$explanation
rung adds causal explanation. Neither rung touches scoring, edge cases, or input hygiene. (This half is partly
a manipulation check: the ladder texts were generated by instructions asking for definitions and
explanations.) The informative results concern optimization, which, although free to write anything,
essentially \emph{never} writes concept content. GEPA rounds are dominated by score anchors and
format ($75$--$77\%$); edge-case rules are the most frequent secondary content, followed closely by input hygiene
(``score only the visible comment text''). Low-fidelity rubrics receive \emph{more}
instrument discipline rather than re-teaching of the concept. $M_\omega$ unit additions are on the same
side of the divide, with a shift in subtype: they write checking steps for the task
executor ($51\%$) rather than score anchors for a judge. The
only optimization lineages that write definitions are the small number of esoteric concepts
(e.g.\ narratological story-vs-discourse distinctions) for which the judge cannot be assumed to hold
the concept. \textbf{Decompression writes the concept; optimization, whether reflective or
recombinative, writes the measurement}, except where the concept itself is missing; that is
the case the tacit-isomorphism experiment targets by construction.

\section{Acquisition and pretraining exposure: the corpus-frequency test}
\label{app:exposure}

The fact-learning literature finds that what a model knows tracks how often it was seen
\citep{kandpal2023large,carlini2023quantifying}. If criterion acquisition worked the same way, the order in which concepts become
name-indexed (Section~\ref{sec:tacit-iso}) should follow the corpus frequency of the concepts'
vocabulary. We tested this at two grains, with the estimator and decision rule frozen before
any query.

\begin{figure}[t]
\centering
\resizebox{\textwidth}{!}{%
\begin{tikzpicture}
\input{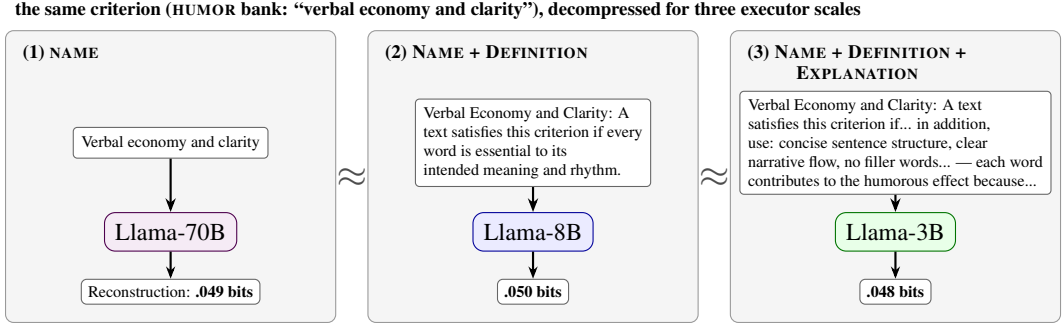}
\end{tikzpicture}}%
\caption{\textbf{Tacit-knowledge isomorphism} (humor bank, criterion ``Concise Wording and
clarity''). Recovery is $.049$/$.050$/$.048$ bits, with
overlapping bootstrap $95\%$ intervals ($[.027,.082]$/$[.026,.083]$/$[.011,.106]$), for a
\textsc{name} to Llama-70B, a name-plus-definition to Llama-8B, and a
name-plus-definition-plus-explanation to Llama-3B. The figure illustrates the pattern that Section~\ref{sec:tacit-iso} measures at scale; the three values are copied by hand from an earlier freeze. The 8B
condition's reference executor is self-consistency-advantaged.}
\label{fig:tacit-iso}
\end{figure}

\paragraph{Elicitable and acquired concepts: the $37$ metrics with a name-only crossing.} Figure~\ref{fig:tacit-iso} shows the pattern the main text calls internalization. ``Concise wording and clarity'' is recovered from its name alone by Llama-70B, needs the name and the definition at Llama-8B, and needs the full explanation at Llama-3B, at the same recovery each time. Such a concept is elicitable at every size we tested, since more text transmits it to the smaller models, and acquired at the larger ones, since the name suffices there. The $37$ metrics for which a name-only crossing $z^*$ can be fitted are the bank's cleanest cases of this kind, and they are the sample for the name-indexing test below (Grain 1). They are a selected class, not a special one: nothing we measure about them, exposure, form or family-relativity, differs from what holds of the $1{,}270$ fitted metrics, which is why the main text reports the bank-scale tests and this appendix keeps the crossing sample.

\paragraph{Exposure is the log corpus frequency of a criterion's name terms in Dolma.} For each criterion, content terms are the name's tokens minus
stopwords and punctuation; exposure is the mean of $\log_{10}(\mathrm{count}+1)$ over the
content unigrams and all adjacent content bigrams, with counts from the infini-gram index over
Dolma v1.7 (a declared proxy: the actual pretraining corpora of the evaluated families cannot
be queried, and Dolma stands in for open web-scale text). We also report two sensitivity
variants: the minimum instead of the mean, and terms drawn from the criterion's definition
text instead of its name. All $4{,}755$ queries resolved (zero missing terms).

\paragraph{Grain 1: name-indexing order (the crossing sample).} For the $37$ concepts with
fitted name-condition crossings, we compute the within-family Spearman $\rho(\text{exposure}, z^*_{\mathrm{name}})$,
handling censored crossings in both declared ways (excluded; imputed at the censor bound).
Across all $12$ combinations (family $\times$ estimator $\times$ censoring variant) no correlation is
negative with a confidence interval excluding zero; most point estimates are weakly positive
(e.g.\ Llama mean-estimator $+.36$ excluded / $+.02$ imputed; Qwen $+.13$ / $+.24$; all CIs
span zero). The preregistered support criterion ($\rho \le -.35$, CI $<0$, both families) is
not met in any combination.

\paragraph{Grain 2: acquisition onset at bank scale.} For all $1{,}270$ fitted metrics, onset
is the fitted logistic midpoint $z_0$ of the pooled recovery curve (defined for every metric).
This grain pools the panel rather than splitting it by family, so it is a generality probe
rather than a main result. The within-domain $\rho(\text{exposure}, z_0)$, Fisher-$z$ pooled across the eight
domains, is $\rho = +.033$, CI $[-.022, +.088]$ (min estimator $+.015$ $[-.040,+.075]$).
Per-domain estimates have mixed signs. The only intervals that exclude zero are creative writing
($+.12$) and humor ($+.13$), which are positive (i.e.\ the \emph{wrong} direction for the exposure
hypothesis), and press releases ($-.18$). Mean exposure is nearly identical across the three
scaling regimes (rising $6.11$ / reaches $5.99$ / bounded $5.94$ $\log_{10}$ units): exposure
separates none of the structure that the criterion taxonomy separates.

\paragraph{Residuals and concept form: a bank-scale test.} The rank residuals suggest
a reading: the criteria acquired far earlier than their exposure predicts look like
compact procedural concepts (``surprise, incongruity, and misdirection mechanics'';
``modularize longer proofs with explicit lemmas''), while maximal-exposure late-onset criteria
are common-vocabulary abstractions (``active voice,'' exposure $7.25$, onset $+2.33$; ``provide
context and continuity,'' onset $+3.79$). At bank scale, this reading does not hold.
We join the blinded nine-type taxonomy (Appendix~\ref{app:taxonomy}; regime results in Figure~\ref{fig:osl-types}) to the per-metric
exposures and onsets, standardize onset and exposure within each domain, and take the residual
of onset on exposure. Compact-checkable types (mechanical checks, craft operations, evidence
and rigor; $n=886$) and the remaining diffuse types ($n=384$) have indistinguishable mean
residuals ($+.006$ vs $-.013$ standardized; difference $+.02$, two-sided within-domain label
permutation $p=.75$, $10$k draws, seed $20260828$). The earliest-acquiring residual quartile is
$67\%$ compact-checkable against a $70\%$ base rate, so there is no enrichment. The named examples
above are members of the residual tails, but they do not reflect a pattern in the bank as a whole.
We record only the following type-level departures, as exploratory results: identity/persona and external-bundle
criteria fall early relative to their exposure (mean residual $-.25$, $n=38$ and $41$), while
mechanical checks and audience fit fall late ($+.15$, $n=61$; $+.13$, $n=187$). Artifact:
\texttt{outputs/analyses/exposure\_residual\_types\_v1/} (join recipe, seed, per-type table).

\paragraph{Limitations of the exposure test.} Name-term frequency is a coarse proxy that conflates common English
words with concept exposure. The two positive domains plausibly reflect this confound
(generic vocabulary names attach to vaguer concepts). A co-occurrence estimator is left to future
work. The crossing sample's fits predate the formulation-symmetric key policy, and the analysis
reruns automatically when refreshed fits become available. Artifacts: \texttt{outputs/exp\_exposure/}
(per-metric exposures and onsets, complete query cache, rerunnable pipeline).

\clearpage
\section{Simpler explanations for the bounded regime than tacitness, and what the tests say about each}
\label{app:plateau-checks}
A bounded criterion is one whose agreement with the panel consensus stops rising with capability below the planted reference (Appendix~\ref{app:osl-regimes}). The main text reads this as a limit of articulation. Several simpler explanations would produce the same plateau, and this appendix collects them with the evidence that bears on each. We state the verdict first and point to the test.

\begin{itemize}
\item \emph{The instrument makes it.} The bounded count depends on the capability axis and on the planted reference: on the battery's first principal component the bounded group falls from $65$ to $12$, on equally spaced executor ranks it rises to $273$, and the planted reference is highest in exactly the domains that hold bounded criteria (Appendix~\ref{app:regime-robustness}). The regime sizes and the domain composition survive measurement noise and executor perturbations, but which particular criterion is bounded does not, and the verdict is decided by how far apart the strongest executors sit on the axis. This explanation is not ruled out; it is bounded in size by the robustness tables, and the body should be read with it in view.
\item \emph{The transmitter's labels are one-sided, so there is little to recover.} Not supported: bounded criteria are no more one-sided than their matched controls, and at a single receiver they recover about as well (Appendix~\ref{app:onesided}). On the panel's own probes they are rarer-positive, which changes which criteria are bounded but not how many (Appendix~\ref{app:panel-nulls}).
\item \emph{Thirty labeled items are too few.} Not supported up to ninety. Recovery rises from four to thirty items and then still trails a written rule at Llama-70B, the receiver's choice of items adds nothing over a random draw, and letting both sides mix a rule with examples changes no ordering (Appendix~\ref{app:examples-versions}). With $60$ or $90$ labeled items in place of $30$, bounded recovery does not rise ($-.014$ $[-.023,-.006]$ bits from $30$ to $90$) and the controls fall by the same amount (Appendix~\ref{app:more-labels}). A sample-complexity account at hundreds of items is open.
\item \emph{The corpus never disambiguates the concept.} Thirty labels are consistent with several rules, and a receiver takes the simplest one that fits; if the corpus never contains the cases that separate the criterion from its nearest neighbour, every executor converges on the same stand-in. The input test supports this for a large part of the bounded humor set: on reddit jokes only $17$ of $41$ bounded humor criteria have a defined value, and on published humor pieces they gain $+.073$ $[+.035,+.115]$ bits over their controls (Appendix~\ref{app:input-test}). A direct test with contrastive items chosen to separate each criterion from its neighbour has not been run.
\item \emph{The label depends on a standpoint the message never states.} Identity and social-group criteria, which are over-represented among the bounded ones, depend on whose perspective the label takes. If the transmitter's labels encode one standpoint and the message does not say which, agreement cannot climb with capability, which is the signature we see. The human-agreement review is consistent with this reading, since the high values come from trained coders working with a closed codebook (Appendix~\ref{app:human-agreement}). A test that states the standpoint in the message has not been run.
\item \emph{The labels are unreliable, or the consensus is the ceiling.} Neither: form reliability is the same across regimes, and the strongest executors agree with each other least on bounded criteria (Appendix~\ref{app:panel-nulls}).
\item \emph{The panel stops too early.} Not supported, with a caveat. An executor far past the panel on the capability index (GPT-5.6, $z=4.8$ against the panel's best $2.6$) leaves $0$ of $65$ bounded criteria above the planted reference, and $1$ of $65$ controls. Its agreement with the panel consensus is, however, below the panel's own top executors on every stratum, planted rules included, where it is more accurate than they are against the code truth; past the panel, $y$ measures agreement with the panel's reading, not correctness (Appendix~\ref{app:frontier}).
\item \emph{Post-training, not the concept, sets the plateau.} Not supported: base models are bounded on the same criteria and post-training adds recovery only where recovery already moves (Appendix~\ref{app:posttrain}).
\end{itemize}
Tacitness adds to these only the claim that no finite message closes the gap, and nothing here tests that directly: the longest message any experiment sends is ninety labeled items, and the strongest executor any experiment uses is still read through the panel's consensus. The untested explanation above, contrastive items, is the experiment that would.

\subsection{Label reliability, the consensus, skew and probe halves: four checks on the shipped panel labels}
\label{app:panel-nulls}
Four of the explanations above can be tested on the panel labels that produced the regimes, without relabeling anything (artifact \texttt{outputs/analyses/bounded\_nulls\_v1/}). Table~\ref{tab:panel-nulls} gives the numbers by regime; the three strongest executors on the capability index are ``top three'' throughout.

\emph{Unreliable labels.} If bounded criteria were the ones an executor labels inconsistently across prompt forms, the plateau would be label noise. They are not: the split-half reliability across the four forms is $.95$ for bounded criteria, the same as for reaches and rising, and $.98$ at the top three executors. Not supported.

\emph{The consensus as the ceiling.} If the strongest executors agreed with each other on bounded criteria while disagreeing with a consensus dominated by mid-size models, the consensus would be what caps $y$. The opposite holds: pairwise agreement among the top three is lowest on bounded criteria ($.22$ against $.35$ for reaches and $.36$ for rising) and below each one's agreement with the consensus. Not supported.

\emph{Skew.} Bounded criteria are rarer-positive on the probes: the median YES rate at the top three executors is $.15$, against $.36$ for reaches and $.29$ for rising, and $56\%$ of their top-three points fall outside a YES rate of $.2$ to $.8$. A rare label gives a correlation less to work with. Refitting every curve with a chance-corrected agreement in place of Spearman (Cohen's $\kappa$ on labels thresholded at the median, divided by the same reliability) keeps the size of the bounded group, $72$ against $65$, but changes its membership: $31$ of the $65$ stay bounded and $41$ others join. Not supported as an explanation of the plateau: skew does not create the bounded regime, though it helps decide which criteria are in it.

\emph{Probe halves.} Fitting the regimes on the even-numbered probes alone and on the odd-numbered probes alone gives $92$ and $41$ bounded criteria, the two halves agree on $66\%$ of all verdicts, and $15$ of the $65$ shipped bounded criteria are bounded in both halves. Each half carries twice the noise of the full set, so some attrition is expected, but the number is lower than the $65\%$ single-draw retention of Appendix~\ref{app:regime-robustness} would predict for two independent draws. The bounded verdict of an individual criterion is therefore not stable at the probe level; the size of the group and its domain composition are what the robustness checks support.

\begin{table}[h]
\footnotesize
\begin{center}
\begin{tabular}{l c c c}
\toprule
 & bounded & reaches & rising \\
\midrule
form reliability, all executors (median) & $.95$ & $.95$ & $.94$ \\
form reliability, top three executors & $.98$ & $.97$ & $.96$ \\
pairwise Spearman among the top three & $.22$ & $.35$ & $.36$ \\
Spearman with the consensus, top three & $.40$ & $.63$ & $.52$ \\
median YES rate, top three & $.15$ & $.36$ & $.29$ \\
share with YES rate outside $[.2,.8]$, top three & $.56$ & $.32$ & $.51$ \\
Cohen's $\kappa$ with the consensus, top three & $.24$ & $.43$ & $.35$ \\
\midrule
regime count, shipped & $65$ & $321$ & $884$ \\
regime count, $\kappa$-based $y$ & $72$ & $264$ & $891$ \\
regime count, even probes only & $92$ & $373$ & $763$ \\
regime count, odd probes only & $41$ & $309$ & $888$ \\
\bottomrule
\end{tabular}
\end{center}
\caption{\textbf{Four checks on the shipped panel labels, by regime.} Medians over (executor, metric) points; ``top three'' are the three highest-$z$ executors. The refit rows recount the $1{,}270$ fitted metrics with the alternative $y$ (planted reference recomputed with the same $y$); metrics that no longer fit either curve form are counted as noisy and omitted here ($43$, $42$ and $32$ respectively).}
\label{tab:panel-nulls}
\end{table}

\subsection{One-sided labels do not separate bounded criteria from their controls}
\label{app:onesided}
If bounded criteria were the ones a transmitter labels almost always one way, there would be little information to transmit and the plateau would need no further explanation. The examples artifacts of Appendix~\ref{app:examples-versions} record, for each of the $449$ criteria, whether the transmitter labels every demonstration item one way (one-class) and whether fewer than two items fall on either side among the thirty the receiver sees (short-class), together with the rule channel's recovery. Table~\ref{tab:onesided} compares the $65$ bounded criteria with their $65$ matched controls and the $319$ others. Bounded criteria are no more one-sided than their controls at either transmitter, and on the criteria where both classes are present their rule recovery is comparable to the controls' and above the rest of the bank. Whatever bounds them is not a shortage of positive labels, and it is not visible as low recovery at a single receiver: the plateau is in how the panel's agreement moves with capability, not in how much one receiver recovers.

\begin{table}[h]
\footnotesize
\begin{center}
\begin{tabular}{l l c c c c}
\toprule
transmitter & criteria & one-class & short-class & rule recovery, all & rule recovery, two-class criteria \\
\midrule
Qwen2.5-14B & bounded ($65$) & $.40$ & $.69$ & $.005$ & $.055$ ($n{=}32$) \\
 & matched controls ($65$) & $.42$ & $.54$ & $.004$ & $.080$ ($n{=}36$) \\
 & other ($319$) & $.31$ & $.50$ & $.009$ & $.028$ ($n{=}193$) \\
\midrule
Llama-70B & bounded ($65$) & $.28$ & $.38$ & $.013$ & $.106$ ($n{=}43$) \\
 & matched controls ($65$) & $.29$ & $.35$ & $.022$ & $.088$ ($n{=}43$) \\
 & other ($319$) & $.24$ & $.39$ & $.011$ & $.023$ ($n{=}219$) \\
\bottomrule
\end{tabular}
\end{center}
\caption{\textbf{One-sided labels by stratum.} Share of criteria whose transmitter labels every demonstration item one way (one-class) or leaves fewer than two items on one side among the thirty shown (short-class), and the median rule-channel recovery in bits, over all criteria and over those with at least two items per class. Computed from the per-criterion rows of \texttt{outputs/exp\_examples\_shown/summary\_\{qwen25-14b,llama70b\}.json}.}
\label{tab:onesided}
\end{table}

\subsection{Ninety labeled items in place of thirty: recovery does not rise}
\label{app:more-labels}
The receiver in every rule channel writes its rule from $30$ balanced labeled items. If bounded criteria were simply the ones that need more items to pin down, recovery would rise with the number of items on them and not on the controls. We re-ran the rule channel at Qwen2.5-14B (transmitter, receiver and executor) with the receiver shown the first $30$, $60$ or $90$ finite items of a $90$-item demonstration pool, nested so that a larger set adds items and never swaps them, and recovery measured on the same held-out items at every size ($I(m;m')$ over at least $60$ items, four reconstructions per size, context window $24{,}576$ tokens). Of the $65$ bounded criteria and their $65$ matched controls, $112$ had enough held-out items after the pool and $18$ were skipped. The protocol, readouts and verdict rule were frozen first (\texttt{notes/2026-10-13\_\_exp-more-labels-prereg.md}; artifact \texttt{outputs/exp\_more\_labels/summary\_qwen25-14b.json}).

Table~\ref{tab:more-labels} gives the result. Recovery does not rise with more items on either stratum: it plateaus by $30$ items, and bounded criteria stay below their controls at every size. Bounded criteria recover $.042$ bits from $30$ items, $.026$ from $60$ and $.027$ from $90$, a change of $-.014$ $[-.023,-.006]$ from $30$ to $90$; the controls go from $.051$ to $.040$, a change of $-.011$ $[-.019,-.003]$, and the interaction is $-.004$ $[-.015,+.008]$. No criterion in either stratum gains more than two noise standard deviations ($.104$ bits) from the extra items. The small fall is the same for both strata, and we read it as the cost of a longer context (ninety items of up to $600$ characters) for the rule the receiver writes, not as information lost. By the frozen rule this is ``more labels do not lift bounded criteria''. The $30$-item condition is re-run here with a different held-out set from the shipped rule channel, so comparisons are within this table.

\begin{table}[h]
\footnotesize
\begin{center}
\begin{tabular}{l c c c c}
\toprule
stratum & $30$ items & $60$ items & $90$ items & change, $30$ to $90$ \\
\midrule
bounded ($56$) & $.042$ $[.028,.057]$ & $.026$ $[.017,.036]$ & $.027$ $[.018,.038]$ & $-.014$ $[-.023,-.006]$ \\
matched controls ($56$) & $.051$ $[.032,.070]$ & $.043$ $[.027,.061]$ & $.040$ $[.026,.058]$ & $-.011$ $[-.019,-.003]$ \\
\bottomrule
\end{tabular}
\end{center}
\caption{\textbf{Rule-channel recovery by the number of labeled items the receiver sees.} Mean $I(m;m')$ in bits with $95\%$ bootstrap intervals over criteria, Qwen2.5-14B throughout; the interaction (bounded change minus controls' change) is $-.004$ $[-.015,+.008]$, and no criterion gains more than two noise standard deviations.}
\label{tab:more-labels}
\end{table}

\subsection{Past the panel: a frontier executor on the same instrument}
\label{app:frontier}
The regimes are fitted on $14$ open executors from five model families, whose strongest member sits at $z=2.55$ on the capability index. If the bounded plateaus were only the panel stopping too early, an executor well past that point would lift them. We scored one such executor, GPT-5.6 through its subscription command-line interface, on the same instrument: the same rubric template, two of the four prompt forms, every other probe ($150$ of $300$ per metric), the $65$ bounded criteria, their $65$ matched controls and the $40$ planted rules, and the full $240$-item battery ($51{,}240$ prompts, ten records per call, all parsed). Its $y$ is agreement with the same consensus, which has no member of its family to leave out. A run of Claude Fable 5.1 through its command-line interface answered $1{,}419$ prompts before the subscription's session limit, which the lane did not recognize, and is not reported. Protocol and readouts were frozen first (\texttt{notes/2026-10-13\_\_exp-frontier-extension-prereg.md}; artifact \texttt{outputs/exp\_frontier\_extension/gpt56/summary.json}).

The executor functions, and well past the panel: battery AUC $.992$, $z=4.78$, and on the planted rules its AUC against the code truth is $.93$ against $.86$ for the panel's strongest executor (Qwen2.5-32B), so the frozen function check passes. It does not lift the bounded criteria: $0$ of $65$ sit above the planted reference, against $1$ of $65$ controls, and the median $y$ is $.31$ on both strata (Table~\ref{tab:frontier}). By the frozen rule this is ``the plateau holds''. The caveat is in the same table: the frontier executor's agreement with the consensus is below the panel's own top three on every stratum, by $.26$ on bounded criteria, $.32$ on controls and $.14$ on the planted rules, where it is the more accurate labeler. Past the panel, $y$ measures agreement with the panel's reading of a concept, and an executor that reads the rubric more exactly than the consensus does is scored lower for it. Refitting the $130$ curves with the frontier point added keeps $52$ of $65$ bounded criteria bounded and moves $13$ to reaches, while $17$ controls and others become bounded; because the frontier point enters the top three that set the planted reference and lowers it (creative writing $.77$ to $.72$, math $.62$ to $.55$), these transitions reflect the moved reference as much as the curves, and we report them for completeness only.

\begin{table}[h]
\footnotesize
\begin{center}
\begin{tabular}{l c c c c c}
\toprule
 & $n$ & $y$ at GPT-5.6 & $y$ at the panel's top three & $y$ at the panel's best & above the planted reference \\
\midrule
bounded & $65$ & $.31$ & $.52$ & $.61$ & $0/65$ \\
matched controls & $65$ & $.31$ & $.62$ & $.70$ & $1/65$ \\
planted rules & $40$ & $.58$ & $.67$ & $.76$ & -- \\
\bottomrule
\end{tabular}
\end{center}
\caption{\textbf{A frontier executor on the scaling instrument.} Medians over metrics with a finite value of the reliability-corrected agreement $y$ with the panel consensus: at GPT-5.6 ($z=4.78$), at the panel's three highest-$z$ executors (mean of the three), and at the single panel executor with the highest $y$ on that metric. The last column counts metrics whose frontier $y$ exceeds the planted reference of their domain. Planted-rule accuracy against the code truth on the same probes: $.93$ at GPT-5.6, $.86$ at the panel's strongest executor; battery AUC $.992$.}
\label{tab:frontier}
\end{table}

\subsection{The input test: bounded humor criteria scored on artifacts that contain what they name}
\label{app:input-test}
A bounded criterion may be inarticulable, or it may be articulable and starved of its input: relational tacit knowledge in Collins' sense, tacit only because the receiver was never shown what the criterion refers to. The bounded humor criteria were fitted on reddit jokes, a corpus of one-line and short-form posts from a single subreddit with no author identities, and many of them name things such a post cannot exhibit: a persona held across a body of work, recurring bits, a parody's target, a format being subverted, an ensemble. We therefore keep the criteria, the transmitter and the round trip fixed and change the artifact. All $41$ bounded humor criteria and $41$ matched controls (reaches and rising, same domain, drawn by a fixed seed) are scored on four item sets of $60$: the harness's own reddit-jokes subsample; published humor pieces from the McSweeney's archive ($250$--$1{,}500$ words, one piece per author, site boilerplate removed); Style Invitational contest entries, entry text only; and the same entries with their contest prompt prepended, which supplies the ``game'' the entry plays. Recovery is the free-mode round trip of Section~\ref{sec:reconstruction} ($R=8$, $60$ items, $30$ for the receiver), with no labels of any kind. The preregistered readouts are the interaction (the bounded criteria's gain on the richer artifact minus the controls' gain, since longer inputs may lift every criterion) and the prompt effect on the contest entries. A bounded criterion that recovers on the artifact that contains its subject matter was relational; one that stays flat there is not rescued by inputs. At Qwen2.5-14B (artifact \texttt{outputs/decoy\_control/*\_input\_richness.json}), the bounded criteria recover $.022$ bits on the jokes, where only $17$ of $41$ have a defined value, $.064$ on the McSweeney's pieces ($37$ of $41$), $.073$ on the contest entries and $.068$ with the prompt; the controls recover $.056$, $.045$, $.061$ and $.067$. Paired within criterion, the bounded gain on the pieces is $+.066$ $[+.040,+.093]$ ($15$ of $17$ criteria improve) against $-.007$ $[-.041,+.022]$ for the controls, a difference of $+.073$ $[+.035,+.115]$; on the entries $+.049$ $[+.008,+.100]$ against $+.002$ $[-.040,+.061]$; the contest prompt adds nothing to either group ($+.005$ and $+.001$). At Llama-3.3-70B the bounded criteria are already measurable on the jokes ($36$ of $41$), and the gain on the pieces is $+.070$ $[+.045,+.096]$ against $+.010$ $[-.046,+.061]$ for the controls, a difference of $+.061$ $[-.000,+.122]$; the prompt again adds nothing (artifact \texttt{Llama-3.3-70B-Instruct\_input\_richness.json}). The criteria that gain most name what a joke cannot show: satire's rhetorical impact, formal and format innovation, absurdism, parody craft. Those that stay flat need an audience or a field: memetic virality, high-concept uniqueness. On this test a large part of the bounded humor set behaves as relational tacit knowledge: articulable, but starved of its input on the corpus it was fitted on. The residue that no artifact supplies is the part Section~\ref{sec:scaling} reads as socially anchored; this experiment does not settle how large it is.

\clearpage
\subsection{Base models against their post-trained siblings on the same instrument}
\label{app:posttrain}

Every executor in the panel is a post-trained checkpoint. To see whether the bounded plateau depends on post-training, we ran base (pretrained-only) siblings of six panel executors, and a seventh pair whose post-trained model is not in the panel (Gemma-3-12B), through the same instrument: the metric's prompt (the metric's definition, rendered in four forms) on the $300$ probes, scored against the leave-one-out consensus of the panel's post-trained executors, with $z$ from the same battery. Base models cannot take a chat template, so every executor, base and post-trained alike, ran in completion mode behind one fixed scaffold: six worked records from three mechanical rules that no bank criterion, planted rule or battery item uses, each shown once satisfied and once not, then the probe. The post-trained sibling therefore has two readings, its original chat-mode panel labels and its scaffold labels, and the base model is compared with the latter so that prompt format is held fixed. The design was preregistered (\nolinkurl{notes/2026-10-08__exp-posttrain-regimes-prereg.md}) with four function gates for the base models, and amended before any ladder result was seen (addendum dated 2026-10-09): the gate on scaffold validity was moved from item-level label agreement to agreement of the measurement $y$ across metrics, because the scaffold changes a post-trained model's labels as much as switching to a neighbouring model does; the matched contrast below became the headline readout.

The gates say that base models function but not cleanly. Next-token mass on YES or NO is $.99$ or above for every base model and no probe is unreadable. On the five planted rules the base models reach $.61$ to $1.09$ of their siblings' AUC (Table~\ref{tab:posttrain-gates}), and the share of bank metrics on which a base model's labels are usable (both classes present, spread at least half the sibling's) runs from $.49$ (Gemma-3-12B) to $.94$ (Qwen2.5-32B); the dropped metrics are excluded below. The scaffold itself distorts the capability index: Qwen2.5-14B scores lower than Qwen2.5-7B on the battery under the scaffold although higher in chat mode, so the Qwen ladder fails the monotonicity gate and only the Llama pairs, and the single-point Gemma pair, enter the readouts that place base points on the fitted curves. The Gemma-3-12B post-trained model has no chat-mode panel reading, so its $y$ agreement gate is undefined. Only the Qwen2.5-32B pair passes the three function gates on the full metric set ($.87$, $.96$ and $.94$), and it fails the ladder gate with the rest of its family. The numbers are therefore descriptive and the preregistered thresholds are reported next to them.

\begin{table}[h]
\footnotesize
\begin{center}
\resizebox{\textwidth}{!}{%
\begin{tabular}{l c c c c c}
\toprule
pair (base vs.\ post-trained) & $y$ agreement (scaffold vs.\ chat) & min.\ planted-rule AUC ratio & usable metrics & $z$: base / post-trained scaffold / post-trained chat & family ladder \\
\midrule
Qwen2.5-3B & $.55$ & $.85$ & $.92$ & $1.14$ / $1.23$ / $1.47$ & \\
Qwen2.5-7B & $.80$ & $.86$ & $.78$ & $1.76$ / $2.03$ / $1.66$ & not monotone \\
Qwen2.5-14B & $.80$ & $.83$ & $.89$ & $1.57$ / $1.61$ / $2.00$ & \\
Qwen2.5-32B & $.87$ & $.96$ & $.94$ & $2.78$ / $3.00$ / $2.55$ & \\
Llama-3.2-3B & $.77$ & $.61$ & $.56$ & $0.41$ / $1.36$ / $1.17$ & \\
Llama-3.1-8B & $.86$ & $.76$ & $.68$ & $1.60$ / $1.72$ / $1.56$ & monotone \\
Gemma-3-12B & -- & $.72$ & $.49$ & $1.39$ / $1.41$ / -- & one point \\
\bottomrule
\end{tabular}}
\end{center}
\caption{\textbf{Function gates for the base models} on all $1{,}387$ metrics. Preregistered thresholds after the addendum: $y$ agreement $\ge .8$, planted-rule ratio $\ge .85$, usable share $\ge .75$, $z$ increasing with size within a family. The Qwen2.5-7B pair passed all four on the $131$-metric gate sample and misses the first by $.001$ on the full set. The Qwen2.5-32B pair passes the three function gates on the full set.}
\label{tab:posttrain-gates}
\end{table}

Post-training helps where transmission already moves and not where it is bounded. Table~\ref{tab:posttrain-r2} gives the matched contrast, the post-trained sibling's recovery under the scaffold minus the base model's, per metric, by regime. Pooled over the seven pairs the post-trained model is ahead by $+.036$ on reaches criteria and $+.034$ on rising ones (differences in $y$, the reliability-corrected agreement of the scaling instrument, not bits), but by only $+.008$ $[-.005,+.021]$ on bounded ones, and the bounded gap is smaller than the other two by $.026$ $[.013,.040]$. Every pair but Qwen2.5-3B shows the same ordering, including the Gemma-3-12B pair ($+.020$ bounded, $+.039$ reaches, $+.042$ rising). In the Qwen2.5-32B pair, the only one that passes every function gate, the base model is ahead in every regime under the scaffold, by $.011$ on rising, $.013$ on reaches and $.040$ on bounded criteria, so the bounded gap is the most negative there as well. So base models do not recover what their post-trained siblings cannot; post-training raises recovery where recovery already moves, barely touches the plateau, and at 32B is behind the base model under the scaffold. Two further readouts agree. Refitting the regimes with the base points added changes $319$ verdicts, against $430$ when the siblings' scaffold points are added instead, so base points move the regimes no more than a prompt-format change does; $13$ bounded criteria become reaches with base points, $12$ with scaffold points. The Qwen2.5-32B points lie above the panel's range ($z=2.8$ and $3.0$) and account for most of the change ($336$ and $504$ verdicts on their own). On the planted mechanical rules the base models are at or below their siblings on nearly every rule (the one exception is Qwen2.5-14B on the shorter-than-median rule, $.85$ against $.78$), so the post-training gap is visible even where the rule is mechanical. Placing base points on the siblings' own scaffold curves (the Llama pairs and the Gemma pair) gives residuals that point in opposite directions, positive at Llama-3.2-3B and negative at Llama-3.1-8B and Gemma-3-12B, and pool to about zero, which we do not interpret.

\begin{table}[h]
\footnotesize
\begin{center}
\resizebox{\textwidth}{!}{%
\begin{tabular}{l c c c c}
\toprule
pair & bounded & reaches & rising & bounded minus others \\
\midrule
Qwen2.5-3B & $-.009$ $[-.042,+.022]$ & $+.006$ $[-.011,+.024]$ & $-.019$ $[-.032,-.006]$ & $+.003$ $[-.032,+.037]$ \\
Qwen2.5-7B & $+.017$ $[-.003,+.039]$ & $+.040$ $[+.029,+.051]$ & $+.035$ $[+.028,+.044]$ & $-.020$ $[-.041,+.004]$ \\
Qwen2.5-14B & $+.024$ $[-.005,+.055]$ & $+.039$ $[+.026,+.052]$ & $+.051$ $[+.041,+.062]$ & $-.023$ $[-.053,+.008]$ \\
Qwen2.5-32B & $-.040$ $[-.069,-.014]$ & $-.013$ $[-.022,-.005]$ & $-.011$ $[-.017,-.004]$ & $-.028$ $[-.059,-.002]$ \\
Llama-3.2-3B & $+.015$ $[-.039,+.065]$ & $+.077$ $[+.054,+.098]$ & $+.052$ $[+.039,+.065]$ & $-.042$ $[-.099,+.009]$ \\
Llama-3.1-8B & $+.044$ $[+.020,+.067]$ & $+.108$ $[+.091,+.125]$ & $+.123$ $[+.113,+.133]$ & $-.076$ $[-.100,-.051]$ \\
Gemma-3-12B & $+.020$ $[+.002,+.040]$ & $+.039$ $[+.013,+.065]$ & $+.042$ $[+.025,+.060]$ & $-.022$ $[-.046,+.002]$ \\
\midrule
pooled & $+.008$ $[-.005,+.021]$ & $+.036$ $[+.027,+.044]$ & $+.034$ $[+.028,+.040]$ & $-.026$ $[-.040,-.013]$ \\
\bottomrule
\end{tabular}}
\end{center}
\caption{\textbf{Post-trained minus base recovery under one scaffold}, by regime: mean over metrics of $y_{\text{post-trained}} - y_{\text{base}}$ with $95\%$ bootstrap intervals, metrics with unusable base labels excluded. Positive means the post-trained sibling recovers more. Artifact: \texttt{outputs/exp\_posttrain\_regimes/ladder\_v3/summary\_7pairs.json}.}
\label{tab:posttrain-r2}
\end{table}
\clearpage

\subsection{Published human agreement on the bounded, reaches and rising criteria}
\label{app:human-agreement}

A concept that is bounded for the executors we study need not be inarticulable for people. To see which bounded criteria humans agree on, we searched for a published measurement of human agreement for each of the $65$. For each criterion we ran three to four web searches pairing the concept with inter-rater agreement terms (kappa, Krippendorff's alpha, intraclass correlation, percent agreement) and with its nearest named concept in the relevant literature (humor annotation, satire comprehension, content analysis of news, peer-review reliability). A language-model agent ran the searches; a second pass checked every citation it kept: the title and year were matched, the DOI or URL resolved, and each statistic was read from the paper's own text. Citations that could not be checked were dropped. We examined about $80$ papers and cite $53$. The match between the bank definition and the concept the study measured is recorded for every row as exact, close or loose, and the verdict uses the single-rater coefficient when the paper gives one, with high at $.80$ or above, moderate from $.40$ to $.79$ and low below $.40$.

Three limits apply to reading the tables. First, the high values come almost entirely from trained coders working with a closed codebook; untrained raters, or raters of different politics, agree far less on the same concepts (satire target and stance, sarcasm, roast tone). Second, some coefficients are pooled over many raters: the authenticity ratings of stand-up performers are intraclass correlations over up to $158$ viewers, and the paper does not say so, so single-viewer agreement there is much lower. Third, the agenda-setting figure is a correlation between a newspaper's issue emphasis and its readers' issue ranking, one number for a community of $100$ voters, which is transmission at the aggregate level and not agreement between raters.

Tables~\ref{tab:human-agreement-humor} and~\ref{tab:human-agreement-other} give, for each of the 65 bounded criteria, the closest published measurement of whether humans agree on it. ICC is the intraclass correlation coefficient, $\kappa$ is Cohen's or Fleiss' kappa, $\alpha$ is Krippendorff's alpha, $r$ is the Pearson correlation and $\rho$ the Spearman correlation; ``agg.'' marks a coefficient computed on ratings averaged over raters. We call agreement high at .80 or above, moderate from .40 to .79 and low below .40, using the single-rater coefficient when a study reports one; accuracy on a balanced two-way choice is first converted to $2\times\text{accuracy}-1$, and results without a chance-corrected coefficient are classed from the reported finding. The italic word after each verdict states how closely the measured construct matches the criterion (exact, close or loose). For criteria marked ``none found'' the table gives the nearest study we found. FAIR stands for findable, accessible, interoperable and reusable.

\begin{table}[p]
\centering
\caption{Human agreement on the 41 bounded humor criteria in the closest published study: 6 high, 15 moderate, 6 low and 14 none found (all 65 criteria: 8 high, 20 moderate, 9 low, 28 none found).}
\label{tab:human-agreement-humor}
\footnotesize
\renewcommand{\arraystretch}{1.0}
\resizebox{\textwidth}{!}{%
\rowcolors{2}{gray!12}{white}
\begin{tabular}{@{}>{\raggedright\arraybackslash}p{3.2cm}>{\raggedright\arraybackslash}p{3.6cm}>{\raggedright\arraybackslash}p{3.4cm}>{\raggedright\arraybackslash}p{1.2cm}>{\raggedright\arraybackslash}p{2.4cm}@{}}
\toprule
Criterion & What was measured & Agreement & Verdict & Citation\\
\midrule
Absurdism and non sequitur technique & Absurdity humor present in TV scenes & $\alpha$ = .82; 3 trained coders, 10\% of 107 series & high \textit{(close)} & \citep{vanderwal2020humor}\\
Accumulation and format subversion & Nearest: joke mechanism, e.g. exaggeration & $\kappa$ = .71; 3 annotators, 600 sentences & none found & \citep{zhang2019chinesejoke}\\
Balancing humor with pathos & Regular vs dark humor in short videos & $\alpha$ = .72; 3 annotators, 200 videos & moderate \textit{(loose)} & \citep{johns2026darkhumor}\\
Character and impersonation craft & Satirical vs regular news sentences & $\kappa$ = .40; 6 annotators, 13,873 sentences & moderate \textit{(loose)} & \citep{smadu2025selerosa}\\
Comedic narrative structure and closure & Which stand-up transcript lines are funny & 86.7\% raw agreement; 11 raters, 6 transcripts & moderate \textit{(close)} & \citep{romanowski2025punchlines}\\
Concrete imagery and vivid specificity & Word concreteness ratings (single words) & $r$ = .92 with older norms (agg.); 37,058 words & high \textit{(loose)} & \citep{brysbaert2014concreteness}\\
Delivery, timing, and commitment & Perceived sarcasm from synthetic prosody & ICC(2,1) = .15; ICC(2,$k$) = .92; 66 raters & low \textit{(loose)} & \citep{li2026sarcasmprosody}\\
Distinct comic persona and alignment & Comic style, self vs two close others & self-other $r$ = .44--.56; $N$ = 210 & moderate \textit{(close)} & \citep{ruch2018comicstyles}\\
Distinctive personal comic voice & Comic style, self vs two close others & self-other $r$ = .44--.56; $N$ = 210 & moderate \textit{(loose)} & \citep{ruch2018comicstyles}\\
Emotional honesty, vulnerability, resonance & Viewers rate authenticity of stand-up clips & ICC .86--.97 (likely averaged); 158 viewers & high \textit{(close)} & \citep{henderson2013authenticity}\\
Format innovation and risk-taking & Expert creativity ratings of student writing & pairwise judge $r$ = .66--.79; 13 judges & moderate \textit{(loose)} & \citep{baer2004extension}\\
Game identification and escalating beats & Crowd labels of improv ``yes, and'' turns & $\kappa$ = .74; 2 crowdworkers, 90 turn pairs & moderate \textit{(loose)} & \citep{cho2020improv}\\
High-concept hook uniqueness & Joke originality; screenplay scores & $\kappa$ = .19 (3 comedians, 240 jokes); script ICC .29--.37 & low \textit{(close)} & \citep{amir2026jokespace,leonard2026blacklist}\\
Host presence and utilization & Nearest: viewers rate stand-up authenticity & ICC .86--.97 (type unstated); no host-fit study & none found & \citep{henderson2013authenticity}\\
Improv ensemble support and agreement & Crowd labels of improv ``yes, and'' turns & $\kappa$ = .74; 2 crowdworkers, 90 turn pairs & moderate \textit{(close)} & \citep{cho2020improv}\\
In-the-moment responsiveness, ad-libs & Tell improvised from scripted acting & 69--83\% correct (chance 50\%); 216 viewers & moderate \textit{(close)} & \citep{hughes2026improvisation}\\
Memetic design, stickiness, virality & Guess which tweet wording gets more retweets & 61.3\% correct (chance 50\%); 106 raters & low \textit{(loose)} & \citep{tan2014wording}\\
News satire format and plausibility & Label headlines as satirical or real & 87.4\% correct (chance 50\%); 10 annotators & moderate \textit{(close)} & \citep{rogoz2021saroco}\\
Originality of premise and angle & Comedians rate joke originality & $\kappa$ = .19, mean $\rho$ = .24; 3 comedians, 240 jokes & low \textit{(close)} & \citep{amir2026jokespace}\\
Packaging and credits as comedic space & Nearest: humor ratings of short texts & $\alpha$ = .12; 20 raters per text, 10,000 texts & none found & \citep{meaney2021hahackathon}\\
Parody craft and voice fidelity & Which passage better imitates an author's style & $\kappa$ = .54--.58 trained; .10--.26 lay readers & moderate \textit{(close)} & \citep{chakrabarty2025readers}\\
Parody, homage, and pastiche & Parody humor present in TV scenes & $\alpha$ = .82; 3 trained coders & high \textit{(close)} & \citep{vanderwal2020humor}\\
Persona distinctness and consistency & Which passage better imitates an author's style & $\kappa$ = .54--.58 trained; .10--.26 lay readers & moderate \textit{(loose)} & \citep{chakrabarty2025readers}\\
Personal authenticity and vulnerability & Viewers rate authenticity of stand-up clips & ICC .86--.97 (likely averaged); 158 viewers & high \textit{(close)} & \citep{henderson2013authenticity}\\
Personal history, lived experience as engine & Post discloses author's own medical condition & $\kappa$ = .88 (mean pairwise); 3 annotators, 6,639 posts & high \textit{(close)} & \citep{valizadeh2021selfdisclosure}\\
Physical and visual comedy execution & Slapstick present in TV scenes & $\alpha$ = .74; 3 trained coders & moderate \textit{(close)} & \citep{vanderwal2020humor}\\
Roast/insult craft and tone & Teaser vs target ratings of the same tease & teaser-target $r$ = .55; 35 roommate pairs & moderate \textit{(close)} & \citep{kruger2006teasing}\\
Satire target selection and sharpness & Target category of dark-humor memes & $\kappa$ = .72; 3 annotators, 3,503 memes & moderate \textit{(close)} & \citep{kasu2025dhumor}\\
Satirical target clarity and intent & Whether viewers read Colbert's satire as sincere & reading depends on viewer ideology; $N$ = 332 & low \textit{(exact)} & \citep{lamarre2009satire}\\
Sociopolitical engagement and stance & Perceived political stance of a satirist & reading depends on viewer ideology; $N$ = 332 & low \textit{(close)} & \citep{lamarre2009satire}\\
Specificity and show-don't-tell & Nearest: word concreteness ratings & $r$ = .92 with older norms (agg.); 37,058 words & none found & \citep{brysbaert2014concreteness}\\
\bottomrule
\end{tabular}}
\par\smallskip
\parbox{\textwidth}{\scriptsize No human agreement study was found for: Guest/collaboration strategy; Show format and recurring structures. The nearest study reports no agreement statistic for: Cringe/awkwardness calibrated \citep{uysal2014vicarious,mayer2021cringe}; Musical branding: themes, recurring cues \citep{silas2026sonic}; Recurrence, motifs, running jokes, world-building; Recurring bits and identity building; Recurring world-building elements; Running gags and recurrence; Signature recurring bits/segments/characters \citep{topolinski2014runninggags}; Satire's rhetorical impact \citep{holbert2011juvenalian}.}
\end{table}

\begin{table}[p]
\centering
\caption{Human agreement on the 24 bounded creative-writing, news-homepage and peer-review criteria in the closest published study: 2 high, 5 moderate, 3 low and 14 none found (all 65 criteria: 8 high, 20 moderate, 9 low, 28 none found).}
\label{tab:human-agreement-other}
\footnotesize
\renewcommand{\arraystretch}{1.0}
\resizebox{\textwidth}{!}{%
\rowcolors{2}{gray!12}{white}
\begin{tabular}{@{}>{\raggedright\arraybackslash}p{3.2cm}>{\raggedright\arraybackslash}p{3.6cm}>{\raggedright\arraybackslash}p{3.4cm}>{\raggedright\arraybackslash}p{1.2cm}>{\raggedright\arraybackslash}p{2.4cm}@{}}
\toprule
Criterion & What was measured & Agreement & Verdict & Citation\\
\midrule
\multicolumn{5}{@{}l}{\textit{Creative writing}}\\
Epistolary form: structure, timing, authenticity & Nearest: crowd genre labels for web pages & Fleiss $\kappa$ = .874, 88.2\% agreement; 42 annotators & none found & \citep{asheghi2014genres}\\
\midrule
\multicolumn{5}{@{}l}{\textit{News homepages}}\\
AI use disclosure and fidelity & Nearest: spotting AI-generated news images & $\alpha$ = .22; 3 raters per item & none found & \citep{huang2024miragenews}\\
Accuracy, verification and transparency & Fact-checkers rate the same articles or claims & $\kappa$ = .38 (3 checkers, 207 articles); $\kappa$ = .52 (2 outlets) & moderate \textit{(close)} & \citep{allen2021crowds,lim2018factcheckers}\\
Agenda setting and attribute salience & Media issue emphasis vs voters' issue salience & $r$ = +.967 / +.979 (agg.); 100 voters, 1968 & high \textit{(close)} & \citep{mccombs1972agenda}\\
Clarity, simplicity, and precise language & Experts grade German text simplifications & $\alpha$ = .75; 3 experts, 360 pairs & moderate \textit{(close)} & \citep{korobeynikova2026detect}\\
Counter-misinformation reporting & Lay vs fact-checker trust in news sources & $r$ = .90 (agg., 60 sources); checkers' own $\kappa$ = .38 & high \textit{(loose)} & \citep{pennycook2019crowdsourced,allen2021crowds}\\
Originality, attribution, no fabrication/plagiarism & Label plagiarism type in generated text & unanimous among 3 raters: 52--87\% by type & moderate \textit{(close)} & \citep{lee2025plagbench}\\
Visual salience and promotability & Clickbait strength of news tweets & $\kappa$ = .21 (4 classes), .36 (binary); 5 raters, 38,517 tweets & low \textit{(close)} & \citep{potthast2018clickbait}\\
\midrule
\multicolumn{5}{@{}l}{\textit{Peer review}}\\
Accessibility and inclusive communication & Accessibility item of a learning-object rubric & 65\% agree on modal rating; ICC not computable & low \textit{(loose)} & \citep{vargo2003learning}\\
Audience-tailored communication and accessibility & Readability of scientific summaries (binarized) & $\kappa$ = .6 & moderate \textit{(loose)} & \citep{cachola2025evaluating}\\
Conciseness, length, and format & Nearest: reviewer agreement on manuscripts & mean ICC = .34, mean $\kappa$ = .17; 48 studies & none found & \citep{bornmann2010reliability}\\
Data availability statements & Nearest: coding data availability statements & $>$\,90\% agreement with gold standard; no $\kappa$ & none found & \citep{hardwicke2018data}\\
Educational materials: audience fit, exercises & Learning-object quality rubric items & single ICC .13--.55; 10-rater mean .60--.92 & low \textit{(close)} & \citep{vargo2003learning}\\
Professional, civil, constructive tone & Toxicity of peer-review sentences; talk-page attacks & $\kappa$ = .55 (3 annotators); attacks $\alpha$ = .45 & moderate \textit{(close)} & \citep{luo2025peertoxic,wulczyn2017exmachina}\\
Reproducibility artifacts and workflows & Nearest: human vs automated reproducibility checklist & $\kappa$ = .67, human vs tool only & none found & \citep{daros2026reproducibility}\\
\bottomrule
\end{tabular}}
\par\smallskip
\parbox{\textwidth}{\scriptsize No human agreement study was found for: Historical and intellectual contextualization; Live/developing coverage timeliness. The nearest study reports no agreement statistic for: Epistolary and mediated/documentary forms \citep{gius2019narrativelevels}; Engagement and ranking signals \citep{pennycook2019crowdsourced}; Explain responses, offer actionable insights \citep{lombard2002content}; Responsible statistical and polling reporting \citep{louwerse2022reporting}; Source transparency and citations \citep{zhang2018credibility}; Artifact availability, licensing, FAIR \citep{sun2022fair}; Independent validation and reproducibility evidence \citep{stodden2018empirical}.}
\end{table}

\paragraph{Human agreement on reaches and rising criteria is mostly low, the reverse of the bounded picture.} We ran the same search for the $321$ reaches and $884$ rising criteria, at the level of concept families rather than single criteria, since the sets are too large to search one by one (Tables~\ref{tab:human-agreement-reaches} to~\ref{tab:human-agreement-rising-b}). A study was found for $185$ of the reaches criteria and $494$ of the rising criteria. The picture is the reverse of the bounded one: of the $679$ with a measurement, $4\%$ show high agreement, $43\%$ moderate and $53\%$ low. Two cautions. One expert study, the fourteen pass-or-fail craft tests of \citet{chakrabarty2024ttcw} on $48$ stories ($\kappa=.27$ to $.39$), accounts for $131$ of the $357$ low verdicts, so the low share rests heavily on it. And most matches are loose, because the literature rates the whole (a manuscript, a joke, a story) rather than the specific criterion, so a low verdict often means that humans disagree about the whole, not about the part.

\begin{table}[p]
\centering
\caption{Human agreement on the 321 reaches criteria by family: a study was found for 185 (13 high, 98 moderate, 74 low), none for 136. Families of one task with the same study and verdict share a row.}
\label{tab:human-agreement-reaches}
\footnotesize
\renewcommand{\arraystretch}{1.0}
\setlength{\tabcolsep}{3pt}
\resizebox{\textwidth}{!}{%
\rowcolors{2}{gray!12}{white}
\begin{tabular}{@{}>{\raggedright\arraybackslash}p{3.7cm}>{\raggedleft\arraybackslash}p{0.5cm}>{\raggedright\arraybackslash}p{4.2cm}>{\raggedright\arraybackslash}p{4.9cm}>{\raggedright\arraybackslash}p{2.5cm}>{\raggedright\arraybackslash}p{3.8cm}@{}}
\toprule
Family & $n$ & What was measured & Agreement (statistic, raters, items) & Verdict (match) & Citation\\
\midrule
\multicolumn{6}{@{}l}{\textit{Creative writing}}\\
Figurative language; exposition; structure and pacing; setting & 22 & Expert pass/fail tests (TTCW) & $\kappa$: devices .37, exposition .27, structure .39, coherence .27, setting .33; 3 of 10 experts, 48 stories & low \textit{(close)} & \citep{chakrabarty2024ttcw}\\
Query, pitch, market fit & 5 & Reviewer agreement on manuscripts & mean $\kappa$ = .17, ICC = .34; 48 studies & low \textit{(loose)} & \citep{bornmann2010reliability}\\
Humor and satire & 3 & Crowd funniness ratings of tweets & $\alpha$ = .16; 27,282 tweets & low \textit{(loose)} & \citep{castro2018humor}\\
Translation fidelity & 1 & MT adequacy and fluency ratings & $\kappa$ = .23, .25; WMT 2007 judges & low \textit{(close)} & \citep{callisonburch2007meta}\\
Endings; theme; dialogue and voices; originality & 17 & Expert pass/fail tests (TTCW) & $\kappa$: ending .48, theme .66, voice .44, originality .40--.41; 48 stories & moderate \textit{(close)} & \citep{chakrabarty2024ttcw}\\
Prose style and diction & 11 & Readers pick the better excerpt & MFA readers $\kappa$ = .41--.67; lay .08--.26 & moderate \textit{(close)} & \citep{chakrabarty2025readers}\\
Economy and concision & 10 & Writers mark spans to cut or fix & span precision .57; 3 writers, 50 paragraphs & moderate \textit{(close)} & \citep{chakrabarty2025lamp}\\
Intertextuality and allusion & 2 & Experts identify biblical allusions & multi-rater $\kappa$ about .54--.76; 3 experts & moderate \textit{(loose)} & \citep{manjavacas2021intertextuality}\\
Mechanics and conventions & 2 & Raters judge preposition errors & $\kappa$ = .63; 2 raters, 1,800 contexts & moderate \textit{(close)} & \citep{tetreault2008native}\\
Narrative reliability & 1 & Experts classify unreliable narrators & $\kappa$ = .71--.75; 817 texts & moderate \textit{(close)} & \citep{brei2025tuna}\\
Rhythm, sound, musicality & 5 & Poets rate the creativity of poems & $\alpha$ = .83 (agg.); 10 poets, 205 poems & high \textit{(loose)} & \citep{kaufman2008comparison}\\
\midrule
\multicolumn{6}{@{}l}{\textit{Humor}}\\
Timing, pacing, delivery & 7 & Sarcasm perceived from prosody & ICC(2,1) = .15; 66 raters & low \textit{(loose)} & \citep{li2026sarcasmprosody}\\
Funniness and laugh yield & 5 & Crowd funniness ratings of jokes & $\alpha$ = .12; 20 raters per text, 10,000 texts & low \textit{(close)} & \citep{meaney2021hahackathon}\\
Originality & 4 & Comedians rate joke originality & $\kappa$ = .19; 3 comedians, 240 jokes & low \textit{(exact)} & \citep{amir2026jokespace}\\
Irony and play signals & 3 & Readers detect intended sarcasm & $\kappa$ = .36--.39; 3 readers per tweet & low \textit{(close)} & \citep{oprea2020isarcasm}\\
Political and topical satire & 3 & Viewers read Colbert's intent & split by ideology, no coefficient; $N$ = 332 & low \textit{(close)} & \citep{lamarre2009satire}\\
Self-deprecation and humor style & 2 & Coding of humor style & $\kappa$ = .27, .28; 3 annotators, 200 texts & low \textit{(close)} & \citep{kenneth2024humourstyle}\\
Joke structure, setup, premise & 9 & Locating punchlines in stand-up & 86.7\% raw agreement; 11 raters & moderate \textit{(loose)} & \citep{romanowski2025punchlines}\\
Offense, taboo, tonal calibration & 5 & Offensiveness ratings of jokes & $\alpha$ = .52; 20 raters per text, 10,000 texts & moderate \textit{(close)} & \citep{meaney2021hahackathon}\\
Audience fit, translation & 4 & Ratings of cultural translations & $\alpha$ = .45; 5 raters per language, 15 languages & moderate \textit{(loose)} & \citep{vandoren2026cheese}\\
Wordplay and puns & 4 & Pun sense annotation & $\alpha$ = .78; 2 judges, 1,652 puns & moderate \textit{(close)} & \citep{miller2015pun}\\
Improv collaboration & 3 & Crowd labels of ``yes, and'' turns & $\kappa$ = .74; 2 raters, 90 turn pairs & moderate \textit{(loose)} & \citep{cho2020improv}\\
Genre and style classification & 2 & Regular vs dark humor in videos & $\alpha$ = .72; 3 annotators, 200 videos & moderate \textit{(close)} & \citep{johns2026darkhumor}\\
Parody; absurdism, farce & 6 & Parody, absurdity in TV scenes & $\alpha$ = .82; 3 coders & high \textit{(close)} & \citep{vanderwal2020humor}\\
Persona authenticity & 2 & Viewers rate performer authenticity & ICC .86--.97 (agg.); 158 viewers & high \textit{(close)} & \citep{henderson2013authenticity}\\
\midrule
\multicolumn{6}{@{}l}{\textit{Mathematical exposition}}\\
Proof validity and rigor & 1 & Mathematicians judge one proof & split on validity, no coefficient; 109 raters & low \textit{(close)} & \citep{inglis2013standards}\\
Clarity and plain language & 6 & Readability ratings of passages & pairwise $\kappa$ median .50; 7 raters, 30 passages & moderate \textit{(loose)} & \citep{phani2014bengali}\\
\midrule
\multicolumn{6}{@{}l}{\textit{News homepages}}\\
Headlines and leads & 2 & Clickbait strength of headlines & $\kappa$ = .21; 5 raters, 38,517 tweets & low \textit{(close)} & \citep{potthast2018clickbait}\\
Emotional appeal & 1 & Emotional appeal of arguments & $\alpha$ = .26 (best pair); 304 texts & low \textit{(loose)} & \citep{wachsmuth2017argument}\\
Fact, opinion and satire labels & 1 & Adults sort fact from opinion & near chance, splits by party; 5,035 adults & low \textit{(close)} & \citep{mitchell2018pew}\\
Credibility and verifiability & 1 & Credibility ratings of articles & $\alpha$ = .44--.48 crowd, .75--.83 experts & moderate \textit{(loose)} & \citep{bhuiyan2020credibility}\\
Originality and plagiarism & 1 & Labels of plagiarism type & unanimous for 52--87\% of items; 3 raters & moderate \textit{(close)} & \citep{lee2025plagbench}\\
\midrule
\multicolumn{6}{@{}l}{\textit{Peer review}}\\
Soundness and generalizability & 3 & Reviewer agreement on manuscripts & mean $\kappa$ = .17, ICC = .34; 48 studies & low \textit{(loose)} & \citep{bornmann2010reliability}\\
Title and abstract & 1 & Peer ratings of abstracts & $\kappa$ = .21, .39; 11 reviewers, 87 abstracts & low \textit{(close)} & \citep{cohen2005abstracts}\\
Uncertainty and limitations & 1 & Certainty of reported findings & $\alpha$ = .67, .57; 2,200 findings & moderate \textit{(close)} & \citep{pei2021certainty}\\
\midrule
\multicolumn{6}{@{}l}{\textit{Press releases}}\\
Calls to action, audience framing & 9 & Argument effectiveness ratings & $\alpha$ $\le$ .22; 7 raters & low \textit{(loose)} & \citep{wachsmuth2017argument}\\
Headlines and leads & 1 & Clickbait strength of headlines & $\kappa$ = .21; 5 raters, 38,517 tweets & low \textit{(close)} & \citep{potthast2018clickbait}\\
Credibility and verifiability & 6 & Credibility ratings of articles & $\alpha$ = .44--.48 crowd, .75--.83 experts & moderate \textit{(close)} & \citep{bhuiyan2020credibility}\\
Clarity and plain language & 4 & Readability ratings of passages & pairwise $\kappa$ median .50; 7 raters, 30 passages & moderate \textit{(loose)} & \citep{phani2014bengali}\\
Uncertainty and limitations & 4 & Certainty of reported findings & $\alpha$ = .67, .57; 2,200 findings & moderate \textit{(close)} & \citep{pei2021certainty}\\
\midrule
\multicolumn{6}{@{}l}{\textit{Notice and comment}}\\
Regulatory comment content & 5 & Support type in rule comments & $\kappa$ = .73; 2 coders & moderate \textit{(loose)} & \citep{park2014support}\\
\bottomrule
\multicolumn{6}{@{}p{20.60cm}@{}}{\rule{0pt}{2.2ex}\scriptsize No study was found for 36 families (132 criteria, count per family in parentheses). Creative writing: artistic truth over literal fact (3), representation and memoir ethics (3); Humor: allusion and intertextuality (2), callbacks, running gags, escalation (5), incongruity, surprise, misdirection (14), show structure and coherence (3), visual comedy, cartoons, medium (6), concision and joke writing craft (7), crowd work, rapport, social function (4), memes, virality, platform (5), comic voice, brand and naming (3), observational and relatable humor (5); Mathematical exposition: assumptions, hypotheses, coverage, errata (6), reporting guidelines and artifacts (1), concision and focus (2), examples and counterexamples (5), elegance and economy (2), graphics and visual integrity (2), interdisciplinary connections (1), typesetting (1); News homepages: journalistic orientation (8), news values (4), sponsored-content labels (3), graphics and visual integrity (6); Peer review: rights and ai disclosure (2), reporting guidelines and artifacts (3), citation coverage (1), interdisciplinary connections (1), plain-language summaries (1); Press releases: press-release structure (7), quantitative specificity (4), promotional vs neutral tone (4), news values (4); Notice and comment: concision and focus (1), civil tone and privacy (2); Patents: patent terminology (1). A further 4 criteria inside families in the table also have none found.}\\
\end{tabular}}
\end{table}

\begin{table}[p]
\centering
\caption{Human agreement on the 884 rising criteria by family, part 1 (creative writing, humor, mathematical exposition). Over all rising criteria a study was found for 494 (15 high, 196 moderate, 283 low) and none for 390. Columns and verdict thresholds as in Table~\ref{tab:human-agreement-humor}.}
\label{tab:human-agreement-rising}
\footnotesize
\renewcommand{\arraystretch}{1.0}
\setlength{\tabcolsep}{3pt}
\resizebox{\textwidth}{!}{%
\rowcolors{2}{gray!12}{white}
\begin{tabular}{@{}>{\raggedright\arraybackslash}p{3.9cm}>{\raggedleft\arraybackslash}p{0.5cm}>{\raggedright\arraybackslash}p{4.9cm}>{\raggedright\arraybackslash}p{5.9cm}>{\raggedright\arraybackslash}p{2.5cm}>{\raggedright\arraybackslash}p{4.0cm}@{}}
\toprule
Family & $n$ & What was measured & Agreement (statistic, raters, items) & Verdict (match) & Citation\\
\midrule
\multicolumn{6}{@{}l}{\textit{Creative writing}}\\
Plot structure; causality; twists; pacing; character; exposition; sensory detail; figurative language; setting & 101 & Expert pass/fail tests (TTCW) & $\kappa$: structure .39, coherence .27, pacing .39, character .31, exposition .27, setting .33, devices .37; 3 of 10 experts, 48 stories & low \textit{(close)} & \citep{chakrabarty2024ttcw}\\
Query, pitch, market fit & 14 & Reviewer agreement on manuscripts & mean $\kappa$ = .17, ICC = .34; 48 studies & low \textit{(loose)} & \citep{bornmann2010reliability}\\
Motivation, conflict, stakes & 13 & Crowd labels of character motives & $\alpha$ = .24--.31; crowd, ROCStories & low \textit{(close)} & \citep{rashkin2018psychology}\\
Emotional impact & 9 & Sentence emotion labels in tales & $\alpha$ = .39; 3 annotators & low \textit{(close)} & \citep{christ2022emotion}\\
Argument and exposition & 4 & Expert argument-quality ratings & $\alpha$ $\le$ .22 (7 raters); best pair .35--.39 & low \textit{(close)} & \citep{wachsmuth2017argument}\\
Humor and irony & 1 & Crowd funniness ratings & $\alpha$ = .12; 10,000 texts & low \textit{(close)} & \citep{meaney2021hahackathon}\\
Voice; point of view; theme; originality & 33 & Expert pass/fail tests (TTCW) & $\kappa$: voice .44, theme .66, originality .40--.41; 48 stories & moderate \textit{(loose)} & \citep{chakrabarty2024ttcw}\\
Diction, style, register & 11 & Readers pick the better excerpt & MFA readers $\kappa$ = .41--.67; lay .08--.26 & moderate \textit{(close)} & \citep{chakrabarty2025readers}\\
Suspense and narrative drive & 5 & Readers rate sentence suspense & $\alpha$ = .52--.57; 5 raters, 200 stories & moderate \textit{(close)} & \citep{wilmot2020suspense}\\
Unreliable narration & 3 & Experts classify unreliable narrators & $\kappa$ = .71--.75; 817 texts & moderate \textit{(close)} & \citep{brei2025tuna}\\
Factual accuracy & 1 & Annotators verify claims & $\kappa$ = .68; 5 annotators, 7,506 claims & moderate \textit{(close)} & \citep{thorne2018fever}\\
Poetic form, meter, sound & 4 & Poets rate the creativity of poems & $\alpha$ = .83 (agg.); 10 poets, 205 poems & high \textit{(loose)} & \citep{kaufman2008comparison}\\
\midrule
\multicolumn{6}{@{}l}{\textit{Humor}}\\
Titles, packaging, shareability & 10 & Clickbait strength of headlines & $\kappa$ = .21; 5 raters, 38,517 tweets & low \textit{(loose)} & \citep{potthast2018clickbait}\\
Narrative structure and realism & 8 & Expert story-craft tests (TTCW) & $\kappa$ = .27--.48; 48 stories & low \textit{(loose)} & \citep{chakrabarty2024ttcw}\\
Standards and backlash risk & 8 & Should-it-be-banned ratings & $\alpha$ = .18--.29; 56 raters, 20 tweets & low \textit{(loose)} & \citep{ross2016hate}\\
Timing and delivery & 8 & Sarcasm perceived from prosody & ICC(2,1) = .15; 66 raters & low \textit{(loose)} & \citep{li2026sarcasmprosody}\\
Satire, irony, play frame & 3 & Readers detect intended sarcasm & $\kappa$ = .36--.39; 3 readers per tweet & low \textit{(close)} & \citep{oprea2020isarcasm}\\
Originality & 2 & Comedians rate joke originality & $\kappa$ = .19; 3 comedians, 240 jokes & low \textit{(exact)} & \citep{amir2026jokespace}\\
Emotional resonance & 1 & Story empathy ratings & ICC2k = .34 (agg.); 1,056 stories & low \textit{(loose)} & \citep{chhun2022human}\\
Self-deprecation & 1 & Coding of humor style & $\kappa$ = .27, .28; 3 annotators & low \textit{(close)} & \citep{kenneth2024humourstyle}\\
Offense; target ethics & 18 & Offensiveness ratings of jokes & $\alpha$ = .52; 20 raters per text & moderate \textit{(close)} & \citep{meaney2021hahackathon}\\
Audience fit, translation & 7 & Ratings of cultural translations & $\alpha$ = .45; 5 raters per language & moderate \textit{(loose)} & \citep{vandoren2026cheese}\\
Clarity and plain prose & 6 & Readability ratings of passages & pairwise $\kappa$ median .50; 7 raters & moderate \textit{(close)} & \citep{phani2014bengali}\\
Wordplay and one-liners & 5 & Joke writers rate funniness & $\alpha$ = .40; 7 raters, 80 texts & moderate \textit{(close)} & \citep{toplyn2026jest}\\
Factual accuracy & 4 & Fact-checkers rate articles, claims & $\kappa$ = .38 (3 checkers), .52 (2 outlets) & moderate \textit{(close)} & \citep{allen2021crowds,lim2018factcheckers}\\
Improvisation & 3 & Crowd labels of ``yes, and'' turns & $\kappa$ = .74; 2 raters, 90 turn pairs & moderate \textit{(close)} & \citep{cho2020improv}\\
Tonal balance & 1 & Regular vs dark humor in videos & $\alpha$ = .72; 3 annotators & moderate \textit{(loose)} & \citep{johns2026darkhumor}\\
Persona and authenticity & 5 & Viewers rate performer authenticity & ICC .86--.97 (agg.); 158 viewers & high \textit{(close)} & \citep{henderson2013authenticity}\\
\midrule
\multicolumn{6}{@{}l}{\textit{Mathematical exposition}}\\
Significance, novelty, venue fit & 14 & Reviewer agreement on manuscripts & mean $\kappa$ = .17, ICC = .34; 48 studies & low \textit{(loose)} & \citep{bornmann2010reliability}\\
Abstracts, introductions, titles & 7 & Peer ratings of abstracts & $\kappa$ = .21, .39; 11 reviewers, 87 abstracts & low \textit{(loose)} & \citep{cohen2005abstracts}\\
Proof validity and rigor & 6 & Mathematicians judge one proof & split on validity, no coefficient; 109 raters & low \textit{(close)} & \citep{inglis2013standards}\\
Attribution ethics & 2 & Professors judge plagiarism & moderate disagreement, no coefficient & low \textit{(close)} & \citep{roig2001plagiarism}\\
Prose; audience fit & 11 & Readability ratings of passages & pairwise $\kappa$ median .50; 7 raters, 30 passages & moderate \textit{(loose)} & \citep{phani2014bengali}\\
Elegance and beauty & 5 & Paired comparisons of beauty & split-half $r$ = .70--.72 (agg.) & moderate \textit{(close)} & \citep{sa2023mathematicians}\\
Explanation quality & 4 & Paired comparisons of explanations & split-half $r$ = .78--.80 (agg.) & moderate \textit{(close)} & \citep{evans2022explanation}\\
Citation practice & 3 & Labels of citation function & $\kappa$ = .72; 3 annotators, 548 citations & moderate \textit{(loose)} & \citep{teufel2006citation}\\
\end{tabular}}
\end{table}

\begin{table}[p]
\centering
\caption{Human agreement on the rising criteria by family, part 2 (news homepages, peer review, press releases, notice and comment). Counts as in part 1.}
\label{tab:human-agreement-rising-b}
\footnotesize
\renewcommand{\arraystretch}{1.0}
\setlength{\tabcolsep}{3pt}
\resizebox{\textwidth}{!}{%
\rowcolors{2}{gray!12}{white}
\begin{tabular}{@{}>{\raggedright\arraybackslash}p{3.9cm}>{\raggedleft\arraybackslash}p{0.5cm}>{\raggedright\arraybackslash}p{4.9cm}>{\raggedright\arraybackslash}p{5.9cm}>{\raggedright\arraybackslash}p{2.5cm}>{\raggedright\arraybackslash}p{4.0cm}@{}}
\toprule
Family & $n$ & What was measured & Agreement (statistic, raters, items) & Verdict (match) & Citation\\
\midrule
\multicolumn{6}{@{}l}{\textit{News homepages}}\\
Inclusive language; harm & 11 & Hate speech, offensiveness ratings & $\alpha$ = .18--.29; 56 raters, 20 tweets & low \textit{(close)} & \citep{ross2016hate}\\
News vs opinion; argument & 5 & Adults sort fact from opinion & near chance, splits by party; 5,035 adults & low \textit{(close)} & \citep{mitchell2018pew}\\
Sensationalism and clickbait & 3 & Clickbait strength of headlines & $\kappa$ = .21; 5 raters, 38,517 tweets & low \textit{(close)} & \citep{potthast2018clickbait}\\
Storytelling; surprise & 3 & Story surprise and empathy ratings & ICC2k = .28, .34 (agg.); 1,056 stories & low \textit{(loose)} & \citep{chhun2022human}\\
Framing and context & 2 & Annotators mark policy frames & $\alpha_U$ = .08--.23 by issue & low \textit{(close)} & \citep{card2015frames}\\
Visual manipulation & 2 & Viewers spot manipulated photos & 66\% correct (chance 50\%); 707 viewers & low \textit{(loose)} & \citep{nightingale2017photos}\\
Balance and bias & 1 & Crowd labels of biased sentences & $\kappa$ = .21; 10 raters per sentence & low \textit{(close)} & \citep{spinde2021mbic}\\
Clarity and readability & 7 & Readability ratings of passages & pairwise $\kappa$ median .50; 7 raters, 30 passages & moderate \textit{(loose)} & \citep{phani2014bengali}\\
Verification and corrections & 7 & Fact-checkers rate articles, claims & $\kappa$ = .38 (3 checkers), .52 (2 outlets) & moderate \textit{(close)} & \citep{allen2021crowds,lim2018factcheckers}\\
Fabrication and plagiarism & 3 & Labels of plagiarism type & unanimous for 52--87\%; 3 raters & moderate \textit{(close)} & \citep{lee2025plagbench}\\
Uncertainty signaling & 2 & Certainty of reported findings & $\alpha$ = .67, .57; 2,200 findings & moderate \textit{(close)} & \citep{pei2021certainty}\\
\midrule
\multicolumn{6}{@{}l}{\textit{Peer review}}\\
Methodological rigor & 8 & Risk-of-bias ratings of studies & $\kappa$ = .24--.37; 2 reviewers, 154 trials & low \textit{(close)} & \citep{hartling2013rob}\\
Questions; feasibility; venue & 8 & Reviewer agreement on manuscripts & reject vs not $\kappa$ = .11; 5,881 reviews & low \textit{(loose)} & \citep{kravitz2010reviewers}\\
Claims and conclusions & 3 & Reviewers of a planted-error paper & 68\% of 203 missed unsupported conclusions & low \textit{(close)} & \citep{baxt1998reviewers}\\
Ethics approval & 2 & IRBs review one multisite protocol & 15 sites differ, no coefficient & low \textit{(close)} & \citep{dziak2005irb}\\
Clarity and organization & 5 & Ratings of summary quality & $\alpha$ = .49 crowd, .41 experts; 100 articles & moderate \textit{(close)} & \citep{fabbri2021summeval}\\
Novelty & 3 & Experts rate novelty of ideas & $\kappa$ = .64; 2 experts, 51 ideas & moderate \textit{(close)} & \citep{shahid2025novelty}\\
Citation practice & 2 & Labels of citation function & $\kappa$ = .72; 3 annotators, 548 citations & moderate \textit{(loose)} & \citep{teufel2006citation}\\
Language and grammar & 2 & Raters judge preposition errors & $\kappa$ = .63; 2 raters, 1,800 contexts & moderate \textit{(close)} & \citep{tetreault2008native}\\
Research integrity & 2 & Labels of plagiarism type & unanimous for 52--87\%; 3 raters & moderate \textit{(close)} & \citep{lee2025plagbench}\\
Personal attacks & 1 & Toxicity of review sentences & $\kappa$ = .55; 3 annotators & moderate \textit{(close)} & \citep{luo2025peertoxic}\\
\midrule
\multicolumn{6}{@{}l}{\textit{Press releases}}\\
Headlines and subject lines & 9 & Clickbait strength of headlines & $\kappa$ = .21; 5 raters, 38,517 tweets & low \textit{(close)} & \citep{potthast2018clickbait}\\
Scannable layout & 4 & Design items of a rubric & single-rater ICC = .18--.22; 8 objects & low \textit{(loose)} & \citep{vargo2003learning}\\
Sensitivity and tone-deafness & 3 & Toxicity ratings by rater identity & shift with politics and race, no coefficient & low \textit{(close)} & \citep{sap2022annotators}\\
Evidence-backed claims & 8 & Fact-checkers rate articles, claims & $\kappa$ = .38 (3 checkers), .52 (2 outlets) & moderate \textit{(close)} & \citep{allen2021crowds,lim2018factcheckers}\\
Limitations; disclosure & 8 & Coders mark limitations, funding & $\kappa$ = .53; 5 coders, 507 releases & moderate \textit{(close)} & \citep{winters2019pressrelease}\\
Third-party credibility & 6 & Credibility ratings of articles & $\alpha$ = .44--.48 crowd, .75--.83 experts & moderate \textit{(close)} & \citep{bhuiyan2020credibility}\\
Clarity and plain language & 4 & Readability ratings of passages & pairwise $\kappa$ median .50; 7 raters & moderate \textit{(close)} & \citep{phani2014bengali}\\
Empathy and humble tone & 2 & Crowd labels of empathy & pairwise $\kappa$ = .69 & moderate \textit{(close)} & \citep{sharma2020empathy}\\
Overstatement and causality & 3 & Coders mark exaggeration & $\kappa$ = .88; 27\% of 462 releases & high \textit{(close)} & \citep{sumner2014exaggeration}\\
Risk communication & 2 & Actionability of health materials & AC1 = .92--.93; 2 raters, 110 items & high \textit{(close)} & \citep{vishnevetsky2018pemat}\\
\midrule
\multicolumn{6}{@{}l}{\textit{Notice and comment}}\\
Argument and evidence & 6 & Argument quality of comments & weighted $\kappa$ .14--.20; 84 comments & low \textit{(close)} & \citep{falk2022discourse}\\
Authorship authenticity & 1 & Raters spot bot-written comments & 49.6\% correct (chance 50\%); 108 raters & low \textit{(close)} & \citep{weiss2019deepfake}\\
Engagement; impact evidence & 10 & Support type in rule comments & $\kappa$ = .73; 2 coders & moderate \textit{(loose)} & \citep{park2014support}\\
Clarity and organization & 4 & Readability ratings of passages & pairwise $\kappa$ median .50; 7 raters & moderate \textit{(loose)} & \citep{phani2014bengali}\\
Clear stance & 1 & Annotators mark major claims & $\kappa$ = .88; 3 annotators, 80 essays & high \textit{(close)} & \citep{stab2017parsing}\\
\bottomrule
\multicolumn{6}{@{}p{22.70cm}@{}}{\rule{0pt}{2.2ex}\scriptsize No study was found for 60 families (386 criteria, count per family in parentheses). Creative writing: tagging and titles (4), cross-medium adaptation (11), audience and young readers (4), representation and authenticity (9), visual and layout design (10), genre conventions (9), defamiliarization (3), subplots and ensembles (7), aristotelian categories (5), historical authenticity (5), openings and hooks (7), compliance (4); Humor: crowd work, hosting, casting, chemistry (13), topicality (7), visual, cartoon, production craft (11), relatability (1); Mathematical exposition: completeness, fallacies, case splits (4), definitions and notation (10), unification and strategy (11), figures and representations (9), proof architecture and signposting (17), mathematics pedagogy (4), reproducibility (3); News homepages: attribution and sourcing (7), solutions and service journalism (11), local relevance and curation (8), independence and press freedom (4), photographic craft (6), news values and timeliness (5), investigative reporting (5); Peer review: title and abstract accuracy (3), assumptions and limitations (4), societal impact and fairness analysis (2), reporting quality and figures (5), data and software documentation (7), positioning against prior work (2), theory and formal methods (2); Press releases: multimedia assets (9), quotes and attribution (7), journalist targeting (10), brand and message (11), announcement genres (4), lede and structure (11), seo and distribution (7), stakeholder engagement (8), crisis communication (10), ai and pr origin disclosure (2), newsworthiness (9), media-relations measurement (3), press kits and formats (11), investor communications (4); Notice and comment: participation process (5), off-domain norms (5), actionable alternatives (2), policy-domain content (7), regulatory analysis quality (9), legal precision (4), originality of comments (3); Patents: patent drafting clarity (3), examiner judgments (3). A further 4 criteria inside families in the table also have none found.}\\
\end{tabular}}
\end{table}

\clearpage

\section{Preregistered protocols and scoring-session fingerprints}
\label{app:prereg}
Every ``frozen before the result'' claim points to a dated protocol file whose SHA-256 prefix is recorded before any result exists, and every Table~\ref{tab:main-results} cell to the server fingerprint of the scoring session that produced its row. The files are in \texttt{notes/}; the campaign ledger (\texttt{notes/2026-07-19\_\_gepa-consolidation-and-momega-upgrade-plan.md}, HB\,120--235) records each entry and its outcome.

\begin{center}
{\small
\resizebox{\textwidth}{!}{%
\begin{tabular}{llll}
\toprule
protocol & frozen & hash / fingerprint & result \\
\midrule
Observational-scaling instrument spec (battery, freeze) & 2026-07-07 & \texttt{ac4c4ed2}, \texttt{8c62d3cb} & Section~\ref{sec:scaling} \\
70B name-sufficiency (tacit acquisition) & 2026-07 & \texttt{62e4b3f0} & Section~\ref{sec:tacit-iso} \\
EXP-EXAMPLES-BANK-1 (examples at bank scale) & 2026-08-17 & see note file & Appendix~\ref{app:examples}, $3/236$ \\
EXP-EXPOSURE-1 (corpus-frequency test) & 2026-08-17 & \texttt{d57d8cff7e2f9efe} & Appendix~\ref{app:exposure} \\
EXP-EXAMPLES-BOUNDED-1 (queued) & 2026-08-17 & \texttt{ba48f771b6e192e3} & not run \\
EXP-EXAMPLES-SELECT-2 (selection robustness) & 2026-08-28 & \texttt{d2f607c64cd33318} & null, Appendix~\ref{app:examples} \\
EXP-EXAMPLES-SELF-1 (the transmitter writes its own examples) & 2026-09-14 & see note file & Appendix~\ref{app:examples} \\
EXP-EXAMPLES-SELF-2 (same test, eight tasks, five transmitters) & 2026-09-15 & \texttt{68d2a3b3a841baf1} & Section~\ref{sec:scaling}, Appendix~\ref{app:examples} \\
EXP-VOICE-NOUS-1 / -2 (real authorial voices) & 2026-08-15 / 08-24 & \texttt{85981514891fd900} / \texttt{b1d73d3bb607a6a1} & not supported, Appendix~\ref{app:examples} \\
HotpotQA equal-call decision rule (HB\,193) & 2026-07-29 & fp \texttt{2026-07-29T21:39:06Z} & $+.069$ at $16{,}700$ calls \\
LiveBench re-mint (HB\,209) & 2026-08-19 & fp \texttt{2026-08-20T01:32:58Z} & Table~\ref{tab:main-results} \\
HoVer re-mint (HB\,211) & 2026-08-20 & fp \texttt{2026-08-20T07:01:05Z} & Table~\ref{tab:main-results} \\
IFBench re-mint (HB\,214) & 2026-08-25 & fp \texttt{2026-08-25T07:48:51Z} & Table~\ref{tab:main-results} \\
PUPA re-mint (HB\,216) & 2026-08-26 & fp \texttt{2026-08-27T13:48:44Z} & Table~\ref{tab:main-results} \\
AIME re-mint (HB\,218) & 2026-09-11 & fp \texttt{2026-09-11T09:54:14Z} & Table~\ref{tab:main-results} \\
HotpotQA 7-candidate re-mint (HB\,220) & 2026-09-11 & fp \texttt{2026-09-12T00:19:48Z} & $+.077$ at $16{,}700$ calls, Table~\ref{tab:main-results} \\
No-GEPA-unit ablation and selection replicates (HB\,221) & 2026-09-11 & ledger HB\,221 & HotpotQA $-.013$ n.s., HoVer $-.037$ $[-.058,-.016]$; replicates $.539$/$.541$; \S\ref{app:budgets} \\
Decoy-rule control, identification readout (HB\,222/223) & 2026-09-11 & ledger HB\,222 & readout fooled; Appendix~\ref{app:recon-alternatives} \\
Screened $M_\omega$ selection, HoVer (HB\,225) & 2026-09-12 & ledger HB\,225 & $.520$, $-.052$ $[-.077,-.027]$ vs row of record; \S\ref{app:budgets} \\
Same-session re-mints for HB\,221/225 (HB\,226) & 2026-09-12 & ledger HB\,226 & \S\ref{app:budgets} \\
Complementary-pair probe, HoVer (HB\,227) & 2026-09-12 & ledger HB\,227 & $0/120$ rejected pairs clear the band; Appendix~\ref{app:unit-gate} \\
Decoy-rule control, relabeling channel (HB\,228) & 2026-09-12 & ledger HB\,228 & decoy $.013$ vs criterion $.044$ bits; Appendix~\ref{app:recon-instrument} \\
GEPA warm start from $M_\omega$'s starting prompt (HB\,229) & 2026-09-12 & ledger HB\,229 & $+.001$ / $-.015$ n.s.; \S\ref{app:budgets} \\
Receiver executes its own reconstruction (HB\,231) & 2026-09-12 & ledger HB\,231 & $-.013$ $[-.023,-.003]$; Appendix~\ref{app:recon-instrument} \\
Context supplementation, peer review (HB\,230) & 2026-09-13 & ledger HB\,230 & run; not reported in this draft \\
Capability-axis and low-rank sensitivity (HB\,234) & 2026-09-13 & ledger HB\,234 & Appendix~\ref{app:regime-robustness} \\
Input-richness test, bounded humor criteria (HB\,235) & 2026-09-13 & ledger HB\,235 & Appendix~\ref{app:input-test} \\
EXP-EXAMPLES-SHOWN-1 (the receiver answers with examples) & 2026-10-08 & see note file & showing $<$ telling; Appendix~\ref{app:examples-versions} \\
EXP-POSTTRAIN-REGIMES-1 (base against post-trained siblings) & 2026-10-08 & see note file & Appendix~\ref{app:posttrain} \\
EXP-MORE-LABELS-1 ($30$, $60$, $90$ labeled items) & 2026-10-13 & see note file & no rise; Appendix~\ref{app:more-labels} \\
EXP-FRONTIER-EXTENSION-1 (GPT-5.6 on the instrument) & 2026-10-13 & see note file & plateau holds, with caveat; Appendix~\ref{app:frontier} \\
EXP-PRETRAIN-DYNAMICS-1 (OLMo-2 7B stage-1 checkpoints) & 2026-10-12 & see note file & checkpoints fail the function check (planted-rule AUC $.49$ to $.54$); released base running \\
\bottomrule
\end{tabular}}}
\end{center}

\section{Reproducibility and data statement}
\label{app:reproducibility}
\textit{Code and artifacts.} The supervised harness (\texttt{paperexact\_arms.py}, \texttt{rescore\_k3.py}, the lane scripts) and the reconstruction channel (\texttt{recon\_channel.py}, \texttt{vllm\_backend.py}) are in the project repository, together with every analysis script named in the appendices under \texttt{outputs/analyses/}. Appendix~\ref{app:prereg} lists every preregistered protocol with its hash and every Table~\ref{tab:main-results} cell with its scoring-session fingerprint (host, vLLM version, served model, context length); every supervised number in the paper has one. All executor runs used vLLM $0.25.1$ with the sampling settings recorded in each session fingerprint (Appendix~\ref{app:prereg}); reconstruction runs name the executor and reconstructor models per experiment.

\textit{Benchmarks.} HotpotQA, HoVer, IFBench, PUPA, AIME and LiveBench are the public GEPA evaluation suite \citep{agrawal2025gepa}, used with the published splits and metrics; we add nothing to them.

\textit{The criterion bank.} The ${\sim}5{,}000$ preference metrics are the release of \citet{spangher}. They were mined from public rubrics, style guides and community guidelines for each annotation task and normalized by an LLM into a name and a description; we use them unchanged and added no human annotation for this paper. Item-level labels are of two kinds: labels produced by the models under study, and, for validation only, mention labels derived from human comments on the items (whether a commenter mentioned the criterion), with an arbiter-relabeled subset (Appendix~\ref{app:mi-silver}). No human subjects were recruited. Licensing and access follow the bank's release [authors: license and access URL to be filled in]. The nine datasets, and which analyses use which, are listed in the footnote of Section~\ref{sec:results}.

\end{document}